%% file: main.tex
\documentclass[10pt, conference]{IEEEtran}

\usepackage{amsmath,amssymb,amsfonts}
\usepackage{graphicx}
\usepackage{pifont}
\usepackage{tikz}

\usepackage{amsthm}
\usepackage{xcolor}
\usepackage[figuresright]{rotating}

\usepackage{graphicx}
\graphicspath{{/}{fig/}}

\usepackage{array}
\usepackage{textcomp}
\usepackage{xcolor}
\usepackage{multirow}
\usepackage{booktabs}

\usepackage{mathtools}
\usepackage{breqn}
\usepackage{float}

\usepackage{fancyhdr}
\usepackage{orcidlink}
\usepackage{diagbox}

\usepackage{pgfplots}
\pgfplotsset{compat=1.7}
\usepgfplotslibrary{groupplots}

\usepackage{silence}
\usepackage{multirow,tabularx}
\usepackage{hyperref}
\usepackage{flushend}
\usepackage{algorithmic}
\usepackage[vlined, ruled, shortend]{algorithm2e}

\usepackage{subcaption}
\usepackage{enumitem}

\newlength\figureheight
\newlength\figurewidth
\SetAlCapNameFnt{\footnotesize}
\SetAlCapFnt{\footnotesize}

\title{
    LiDAR-Generated Images Derived Keypoints Assisted Point
Cloud Registration Scheme in Odometry Estimation
}

\author{
    \IEEEauthorblockN{
        \vspace{1em}
        Haizhou Zhang\IEEEauthorrefmark{1}\IEEEauthorrefmark{2}\IEEEauthorrefmark{3}\orcidlink{0009-0005-1321-8687},
        Xianjia Yu\IEEEauthorrefmark{1}\IEEEauthorrefmark{3}\orcidlink{0000-0002-9042-3730},
        Sier Ha\IEEEauthorrefmark{3}\orcidlink{0009-0000-3617-107X},
        Tomi Westerlund\IEEEauthorrefmark{3}\orcidlink{0000-0002-1793-2694}
    }
    \IEEEauthorblockA{
        \normalsize
        \IEEEauthorrefmark{1}These authors contributed equally to this work.\\
        \IEEEauthorrefmark{2}School of Information Science and Technology, Fudan Universtiy, China\\
        \IEEEauthorrefmark{3}\href{https://tiers.utu.fi}{Turku Intelligent Embedded and Robotic Systems (TIERS) Lab, University of Turku, Finland}.\\
        Emails: \textsuperscript{1}\{haizhouzhang21\}@m.fudan.edu.cn, \textsuperscript{2}\{xianjia.yu, sierha, tovewe\}@utu.fi\\[+6pt]
    }
}

\begin{document}

\maketitle
\thispagestyle{empty}
\pagestyle{empty}

\input{sections/00-abstract.tex}
\IEEEpeerreviewmaketitle

\input{sections/01-introduction}

\input{sections/02-related}

\input{sections/03-methodology}

\input{sections/04-experiment_result}

\input{sections/05-conclusion}


\section*{Acknowledgment}

This research was supported by the Research Council of Finland’s AeroPolis project (Grant 442
No. 348480) and RoboMesh project (Grant No. 336061). 

\bibliographystyle{unsrt}
\bibliography{bibliography}

\end{document}

%% file: sections/00-abstract.tex


\begin{abstract}%
    \label{sec:abstract}%
    Keypoint detection and description play a pivotal role in various robotics and autonomous applications including visual odometry (VO), visual navigation, and Simultaneous localization and mapping (SLAM). While a myriad of keypoint detectors and descriptors have been extensively studied in conventional camera images, the effectiveness of these techniques in the context of LiDAR-generated images, i.e. reflectivity and ranges images, has not been assessed.  These images have gained attention due to their resilience in adverse conditions such as rain or fog.  Additionally, they contain significant textural information that supplements the geometric information provided by LiDAR point clouds in the point cloud registration phase, especially when reliant solely on LiDAR sensors. This addresses the challenge of drift encountered in LiDAR Odometry (LO) within geometrically identical scenarios or where not all the raw point cloud is informative and may even be misleading. This paper aims to analyze the applicability of conventional image key point extractors and descriptors on LiDAR-generated images via a comprehensive quantitative investigation. 
Moreover, we propose a novel approach to enhance the robustness and reliability of LO. 
After extracting key points, we proceed to downsample the point cloud, subsequently integrating it into the point cloud registration phase for the purpose of odometry estimation.
Our experiment demonstrates that the proposed approach has comparable accuracy but reduced computational overhead, higher odometry publishing rate, and even superior performance in scenarios prone to drift by using the raw point cloud. This, in turn, lays a foundation for subsequent investigations into the integration of LiDAR-generated images with LO. Our code is available in the github:~\href{https://github.com/TIERS/ws-lidar-as-camera-odom}{https://github.com/TIERS/ws-lidar-as-camera-odom}.
\end{abstract}

\begin{IEEEkeywords}
    LiDAR;
    LiDAR-generated images;
    keypoint detector and descriptor;
    Point cloud registration;
    LiDAR Odometry;
    LO;
    Lidar-as-a-camera;
\end{IEEEkeywords}

%% file: sections/01-introduction.tex
\section{Introduction}
\label{section: Introduction}
\begin{figure}[t]
    \centering
    \begin{subfigure}{0.45\textwidth}
        \centering
        \includegraphics[width=0.8\textwidth]{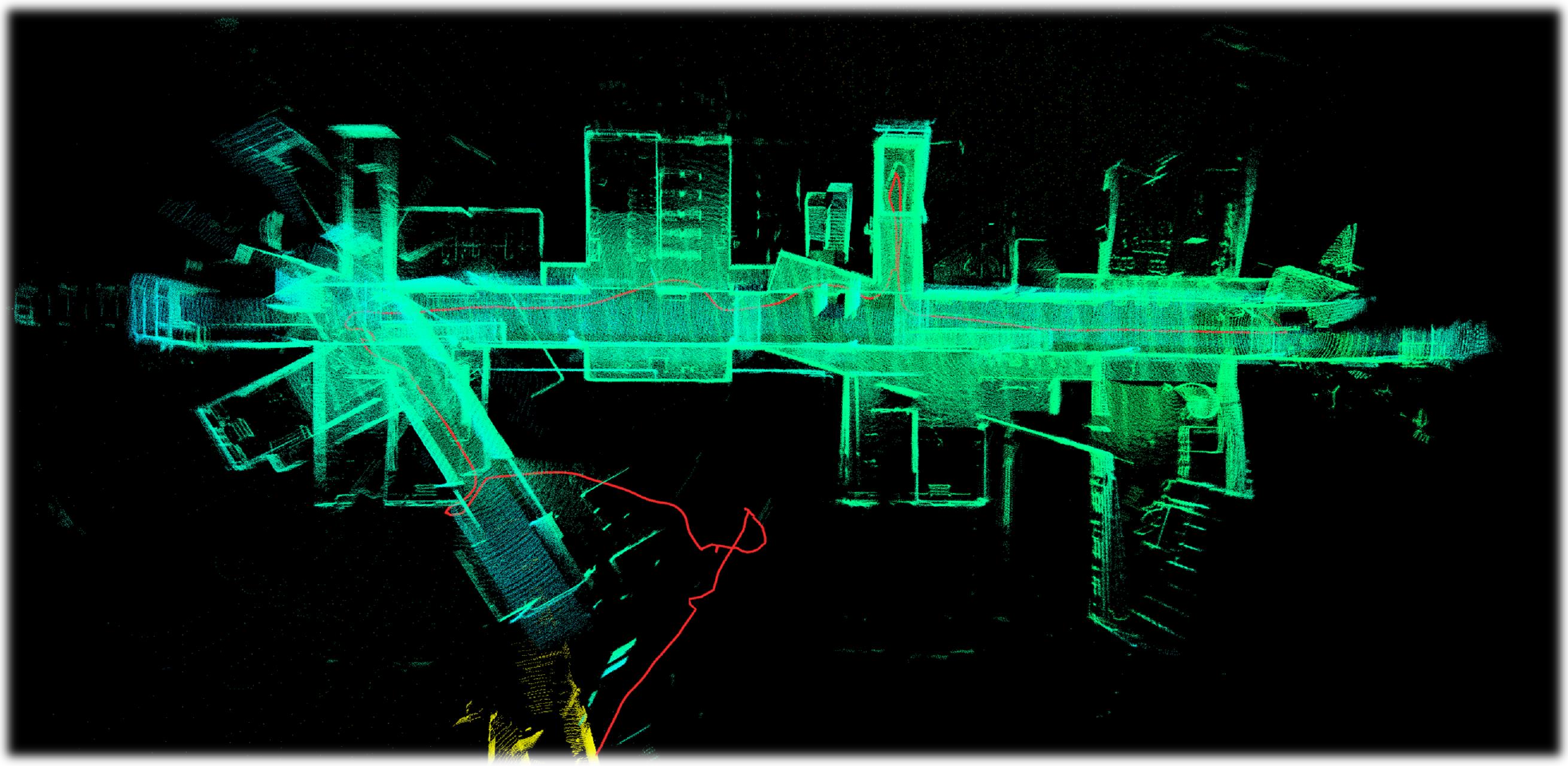}
        \caption{Raw point cloud with point cloud matching approach (KISS-ICP)}
        \label{fig:kiss-icp-bad}
    \end{subfigure}\\[\baselineskip]

    \begin{subfigure}{0.45\textwidth}
        \centering
        \includegraphics[width=0.8\textwidth]{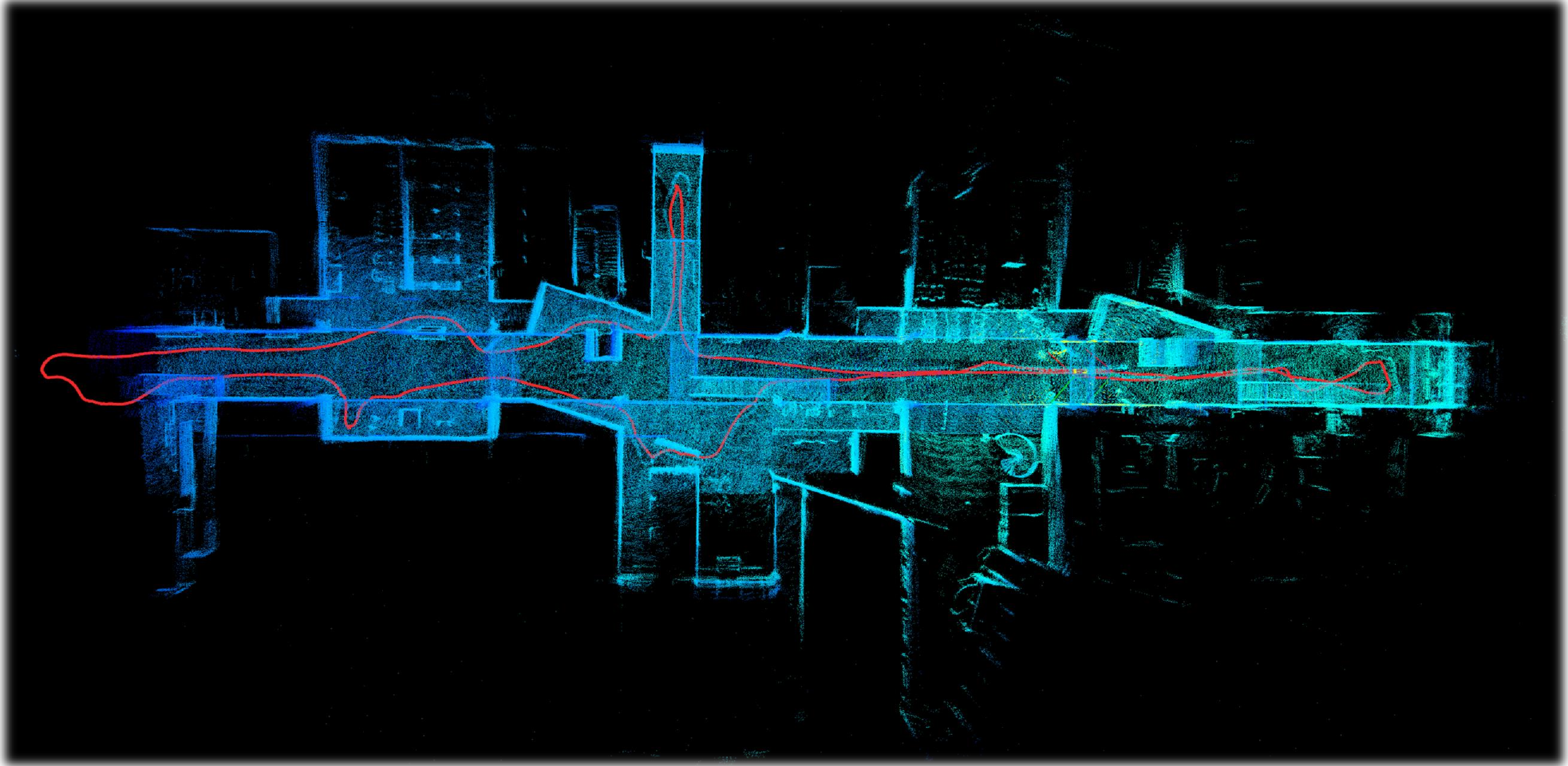}
        \caption{Our proposed LiDAR-generated keypoint extraction-based approach}
        \label{fig:kiss-icp-keypoint}
    \end{subfigure}
    \caption{Samples of LiDAR odometry results run in our experiment}
    \label{fig:results-sample}
\end{figure}

LiDAR technology has become a primary sensor for facilitating advanced situational awareness in the domains of robotics and autonomous systems ranging from LiDAR Odometry(LO), Simultaneous Localization and Mapping (SLAM), object detection and tracking, as well as navigation.
Among these applications, LO, as a fundamental component in robotics has significantly drawn our attention. 
Extensive research efforts have focused on the integration of diverse sensors, including Inertial Measurement Units (IMUs), to bolster LO performance. However, in scenarios where LiDAR data lacks geometric distinctness or even contains misleading information, 
the process of point cloud registration continues to present challenges in achieving precise estimations and even causing drift in certain cases(Fig~\ref{fig:kiss-icp-bad}).

Notably, recent years have witnessed substantial progress in LiDAR technology, marked by the emergence of numerous high-resolution spinning and solid-state LiDAR devices offering various modalities of sensor data~\cite{9981078, sier2023benchmark}. The increased density of the point cloud brings a  challenge for point cloud registration with a significant computation overhead, especially for devices with limited computational resources.

Within the aforementioned modalities, LiDAR-generated images, including reflectivity images, range images, and near-infrared images, have introduced the potential to apply conventional camera image processing techniques to LiDAR-generated images. These images are low-resolution but possibly panoramic and exhibit heightened resilience and robustness in challenging environments, such as those characterized by fog and rain, compared to conventional camera images. Additionally, these images can potentially provide crucial information for point cloud registration when there is a deficiency of geometric data or the raw point cloud lacks useful information so as to avoid drift (Fig.~\ref{fig:kiss-icp-keypoint}).

Keypoint detectors and descriptors have found extensive utility across diverse domains within visual tasks such as place recognition, scene reconstruction, Visual Odometry (VO), Visual Simultaneous Localization and Mapping (VSLAM), and Visual Inertial Odometry (VIO). Nevertheless, there remains a lack of investigation into the performance of extant keypoint detectors and descriptors when applied to LiDAR-generated imagery.

Contemporary methodologies for Visual Odometry (VO) or Visual Inertial Odometry (VIO) rely significantly on the operability of visual sensors, necessitating knowledge of camera intrinsics to facilitate Structure from Motion (SfM) – a requisite not met by LiDAR-generated images. This poses the difficulty of extracting key points from LiDAR-generated images in a certain way to further apply in the odometry estimation.

Therefore, to address the above issues, in this study,  
\begin{enumerate}[label=\roman*)]
    \item We investigate the efficacy of the existing keypoint detectors and descriptors on LiDAR-generated images with multiple specialized metrics providing a quantitative evaluation.
    \item  We conduct an extensive study of the optimal resolution and interpolation approaches for enhancing the low-resolution LiDAR-generated data to extract key points more effectively.  
    \item we propose a novel approach by leverages the detected key points and their neighbors to extract a reliable point cloud (downsampling) for the purpose of point cloud registration with reduced computational overhead and fewer deficiencies in valuable point acquisition.
\end{enumerate}

The structure of this paper is as follows. In Section~\ref{sec:related-work}, we survey the recent progress on keypoint detectors and descriptors including approaches and metrics, point cloud matching, and the status of LO. Section~\ref{sec:meth} provides an overview of the quantitative evaluation of the existing keypoint detectors and descriptors, the proposed keypoint-assisted point cloud registration, and others. Section~\ref{sec:experi-results} demonstrates the experimental results in detail. In the end, we conclude the work and sketch out some future research directions in Section~\ref{sec:conclusion-future}.

%% file: sections/02-related.tex
\section{Related Work}
\label{sec:related-work}

In this section, we commence by presenting a comprehensive review of the prevailing detector and descriptor algorithms documented in the literature. Subsequently, a brief summary of the current advancements in the domain of LiDAR-imaged techniques is offered. We conclude with a concise analysis of the leading algorithms for point cloud registration in LO.

\subsection{Keypoint Detector and Descriptor} 
\label{subsec:keypoint}
In recent years, there have been multiple widely applied detectors and descriptors in the field of computer vision. As illustrated in Table~\ref{tab:detectors-and-descriptors}, we've captured the essential characteristics of different detectors and descriptors. 

Harris detector~\cite{harris1988combined} can be seen as an enhanced version of Moravec's corner detector~\cite{Moravec-1980-15096}~\cite{dey2012comparative}. It's used to identify corners in an image, which are the regions with large intensity variations in multiple directions. The Shi-Tomasi Corner Detector\cite{323794}, is an improvement upon the Harris Detector with a slight modification in the corner response function that makes it more robust and reliable in certain scenarios. The Features from Accelerated Segment Test (FAST)\cite{rosten2006machine} algorithm operates by examining a circle of pixels surrounding a candidate pixel and testing for a contiguous segment of pixels that are either significantly brighter or darker than the central pixel. 

For descriptor-only algorithms, Binary Robust Independent Elementary Features (BRIEF)~\cite{10.1007/978-3-642-15561-1_56} utilizes a set of binary tests on pairs of pixels within a patch surrounding one key point. Fast Retina Keypoint (FREAK)~\cite{6247715} is inspired by the human visual system, which constructs a retinal sampling pattern that is more densely sampled towards the center and sparser towards the periphery. Then it compares pairs of pixels within this pattern to generate a robust binary descriptor.

With respect to the combined detector-descriptor algorithms, the Scale-Invariant Feature Transform (SIFT)~\cite{Lowe2004DistinctiveIF}~\cite{790410} detects key points by identifying local extrema in the Difference of Gaussian scale-space pyramid, then computes a gradient-based descriptor for each keypoint. Speeded-Up Robust Features (SURF)~\cite{10.1007/11744023_32} is designed to address the computational complexity of SIFT while maintaining robustness to various transformations. Binary Robust Invariant Scalable Keypoints (BRISK)~\cite{6126542} uses a scale-space FAST~\cite{rosten2006machine} detector to identify key points and computes binary descriptors based on a sampling pattern of concentric circles. Oriented FAST and Rotated BRIEF (ORB)~\cite{rublee2011orb} extends the FAST detector with a multi-scale pyramid and computes a rotation-invariant version of the BRIEF~\cite{10.1007/978-3-642-15561-1_56} descriptor, aiming to provide a fast and robust alternative to SIFT and SURF. Accelerated-KAZE (AKAZE)~\cite{Alcantarilla2013FastED} employs a Fast Explicit Diffusion scheme to accelerate the detection process and computes a Modified Local Difference Binary (M-LDB) descriptor~\cite{LocalDifferenceBinary2014} for robust matching.

The emergence of deep learning (DL) techniques, particularly convolutional neural networks (CNN) \cite{hassaballah2020deep}\cite{girshick2015fast}, has revolutionized computer vision, over the last decade. SuperPoint\cite{detone2018superpoint} detector employs a fully CNN to predict a set of keypoint heatmaps, where each heatmap corresponds to an interest point's probability at a given pixel location. Then, the descriptor part generates a dense descriptor map for the input image by predicting a descriptor vector at each pixel location. 

To sum up, while numerous detector and descriptor algorithms have gained popularity, it is imperative to note that they have primarily been designed for traditional camera images, not LiDAR-based images. Consequently, it's of paramount importance for this study to identify the algorithms that maintain efficacy for LiDAR-based images.

\input{tbs/key-des-relatedwork}

\subsection{LiDAR-Generated Images in Robotics}


Within the realm of robotics, some studies over the years have delved into the utilization of LiDAR-based images.
But before exploring specific applications, it is vital to know the process by which range images, signal images are generated from point cloud, as detailed in~\cite{wu2021detailed},~\cite{PointCloudLibrary}.
And it's also essential to  understand the effectiveness of LiDAR-based images, through an extensive evaluation in the article~\cite{s23052845}, showing that LiDAR-based images have remarkable resilience to seasonal and environmental variations.


Perception emerges as the indisputable first step in the use of LiDAR within robotics. In~\cite{angus2018LiDAR}, Ouster introduced their work, to explain the possibility of using LiDAR as a camera. They demonstrate the effectiveness of car and road segmentation by putting the LiDAR-based image into a pretrained DL model.  In the work~\cite{9213910}, Tsiourva et al. proposed a saliency detection model based on LiDAR-generated images. In the model, the attributes of reflectivity, intensity, range, and ambient images are carefully contrasted and analyzed. After several advanced image processing steps, multiple conspicuity maps are created. These maps help make a unified saliency map, which identifies and emphasizes the most distinct objects in the image. 
In the research\cite{sier2023uav}, Sier et al. explored using LiDAR-as-a-camera sensors to track Unmanned Aerial Vehicles (UAVs) in GNSS-denied environments, fusing LiDAR-generated images and point clouds for real-time accuracy. The work~\cite{yu2023general} explores the potential of general-purpose deep learning perception algorithms, specifically detection and segmentation neural networks, based on LiDAR-generated images. The study provides both a qualitative and quantitative analysis of the performance of a variety of neural network architectures, proving that the DL models built for visual camera images also offer significant advantages when applied to LiDAR-generated images.

Delving deeper into subsequent applications, for example, localization, research in~\cite{9561335}, explores the problem of localizing mobile robots and autonomous vehicles within a large-scale outdoor environment map, by leveraging range images produced by 3D LiDAR. 

\subsection{Evaluation Metrics for Keypoint Detectors and Descriptors}
\label{subsec:metrics}
The efficacy of detector and descriptor algorithms is typically assessed through some specific evaluation metrics. As illustrated in Table~\ref{tab:metrics}, the first three metrics, Number of Keypoints, Computational Efficiency, and Robustness of Detector are
straightforward to comprehend and implement, and also widely adopted in numerous studies~\cite{Heinly2012ComparativeEO}~\cite{10.1007/11744023_32}~\cite{AComparisonofAffine}. For instance, the Robustness of the Detector~\cite{8346440} is implemented by contrasting key points before and after the transformations like \textit{scaling}, \textit{rotation}, and \textit{Gaussian noise interference}.

When assessing the precision of the entire algorithmic procedure, which is prioritized by the majority of tasks, the prevalent metrics often necessitate benchmark datasets, such as KITTI~\cite{Geiger2013IJRR}, HPatches~\cite{hpatches2017cvpr}. These datasets either provide the transformation matrix between images or directly contain the key point ground truth. For example, in Mukherjee et al.'s study~\cite{Acomparativestudyofimagefeature}, one crucial metric: ``Precision'', is defined as \textit{correct matches/all detected matches}, where correct matches are ascertained through the geometric verification based on a known camera position provided by dataset~\cite{4587706}. Similarly, in this recent work~\cite{OntheComparisonofClassicandDeep}, the evaluation tasks including ``keypoint verification'', ``image matching'', and ``keypoint retrieval'', all rely on the homography matrix between images in the benchmark dataset~\cite{hpatches2017cvpr}.

\input{tbs/metrics}

Nevertheless, given that research predicated on LiDAR images is at a nascent stage, there exists no benchmark dataset in the field of LiDAR-based images. And the effort required for data labeling~\cite{surveyofimagelabelling}~\cite{8634750} to produce such a dataset is considerable and challenging. To bridge this gap, we select multiple key evaluation metrics: Match ratio, Match score, and Distinctiveness, as shown in Table~\ref{tab:metrics} from previous studies. Match Ratio~\cite{Acomparativestudyofimagefeature} is quantitatively defined as \textit{number of matches/number of key points}. A high Match Ratio can suggest that the algorithm is adept at identifying and correlating distinct features; While the exact homography matrix between images remains unknown when lacking banchmark datasets, it can be approximated using mathematical methodologies from two point sets. This computed homography can subsequently be utilized to find correct matches. And \textit{number of estimated correct matches/number of matches} is denoted as Match Score in our work; And Distinctiveness is computed as follows: For every image, the k-nearest neighbors algorithm, with k=2, is employed to identify the two best matches~\cite{Lowe2004DistinctiveIF}. If the descriptor distance of the primary match is notably lower than that of the secondary match, it demonstrates the algorithm's competence in recognizing and describing highly distinctive key points. Consequently, this defines the metric: Distinctiveness.

\subsection{3D Point Cloud Downsampling}
Point cloud downsampling is crucial in operating LO or SLAM within a computation-constrained device. Nowadays, there is a substantial of work focusing on the employment of DL networks, for example, a lightweight transformer~\cite{wang2022lightn}. Other approaches utilized various filters in order to achieve not only point cloud downsampling but also denoising~\cite{zou2020point}.

\subsection{3D Point Cloud Matching in LO}
LO has been widely studied yet challenging due to the complexity of the environment in the robotic field. Contemporary research endeavors have witnessed a notable surge in efforts integrating supplementary sensors, such as Inertial Measurement Units (IMUs), aimed at augmenting the precision and resilience of LO. However, as we focus on the point cloud registration phase of LO, this is out of the scope of the related work of this part. We primarily discuss the solely LiDAR-based LO. Among these solely LiDAR-based approaches, LOAM~\cite{zhang2014loam} as a popular matching-based SLAM and LO approach has encouraged a great amount of other LO approaches including Lego-LOAM~\cite{shan2018lego} and F-LOAM~\cite{wang2021f}.

Point cloud matching or registration constitutes the key component in LO. Since its inception approximately three decades ago, the Iterative Closest Point (ICP) algorithm, as introduced by Besl and McKay~\cite{besl1992method}, has spawned numerous variants. These include notable adaptations such as Voxelized Generalized ICP (GICP)\cite{koide2021voxelized}, CT-ICP\cite{dellenbach2022ct}, and KISS-ICP~\cite{vizzo2023kiss}. Among these ICP iterations, KISS-ICP, denoting ``keep it small and simple'', distinguishes itself by providing a point-to-point ICP approach characterized by robustness and accuracy in pose estimation. Furthermore, the Normal Distributions Transform (NDT)~\cite{biber2003normal} represents another prominent point cloud registration technique frequently employed in LO research. As the latest ICP approach, KISS-ICP is the designated methodology for the point cloud registration we adopted in this study.

%% file: tbs/key-des-relatedwork.tex
\begin{table}[H]
\centering
\caption{Keypoint detectors and descriptors}
\label{tab:detectors-and-descriptors}
\begin{tabular}{m{0.1\linewidth}m{0.1\linewidth}m{0.1\linewidth}m{0.4\linewidth}}
\toprule
\textbf{Method} & \textbf{Detector} & \textbf{Descriptor} & \textbf{Description} \\
\midrule
Harris & $\checkmark$ & & Corner detection method focusing on local image variations. \\ [2.0ex]
Shi-Tomasi & $\checkmark$ & & Variation of Harris with modification in the response function to be more robust. \\[2.0ex]
FAST & $\checkmark$ & & Efficient corner detection for real-time applications. \\[2.0ex]
FREAK & & $\checkmark$ & Robust to transformations, based on human retina's structure. \\[2.0ex]
BRIEF & & $\checkmark$ & Efficient short binary descriptor for key points. \\[2.0ex]
SIFT & $\checkmark$ & $\checkmark$ & Invariant to scale, orientation, and partial illumination changes. \\[2.0ex]
SURF & $\checkmark$ & $\checkmark$ & Addresses the computational complexity of SIFT while maintaining robustness. \\[2.0ex]
BRISK & $\checkmark$ & $\checkmark$ & Faster binary descriptor method, efficient compared to SIFT/SURF. \\[2.0ex]
ORB & $\checkmark$ & $\checkmark$ & Combines FAST detection and BRIEF descriptor, commonly used now. \\[2.0ex]
AKAZE & $\checkmark$ & $\checkmark$ & Builds on KAZE but faster, good for wide baseline stereo correspondence. \\[2.0ex]
Superpoint & $\checkmark$ & $\checkmark$ & A state-of-the-art AI approach that exhibits superior performance when applied to traditional camera images. \\
\bottomrule
\end{tabular}
\end{table}

%% file: tbs/metrics.tex
\begin{table}[h]
\small
\centering
\caption{Metrics for evaluating keypoint detectors and descriptors}
\label{tab:metrics}
\begin{tabular}{m{0.4\linewidth}m{0.4\linewidth}}  
\toprule
\textbf{Metrics} & \textbf{Description} \\
\midrule
Number of Keypoints & A high number of key points can always lead to more detailed image analysis and better performance in subsequent tasks like object recognition. \\ [6.0ex]
Computational Efficiency & Computational efficiency remains paramount in any computer vision algorithms. We gauge this efficiency by timing the complete detection, description, and matching process.\\ [6.0ex]
Robustness of Detector & An efficacious detector should recognize identical key points under varying conditions such as scale, rotation, and Gaussian noise interference. \\ [6.0ex]
Match Ratio & The ratio of successfully matched points to the total number of detected points, offers insights into the algorithm's capability in identifying and relating unique keypoints.\\ [6.0ex]
Match Score & A homography matrix is estimated from two point sets, to distinguish spurious matches, then the algorithm precision is quantified by the inlier ratio.\\ [6.0ex]
Distinctiveness & Distinctiveness entails that the key points isolated by a detection algorithm should exhibit sufficient uniqueness for differentiation among various key points. \\
\bottomrule
\end{tabular}
\end{table}

%% file: sections/03-methodology.tex
\section{Methodology}
\label{sec:meth}

In this section, we first introduce the dataset we used. Then, we describe our experimental procedure in detail.
\subsection{Dataset}
\label{subsec:dataset}
For the evaluation of keypoint detectors and descriptors and our proposed approach, we utilized the published open-source dataset for multi-modal LiDAR sensing~\cite{9981078}. The dataset consists of various LiDARs and among them, Ouster LiDAR provides not only point cloud but also its generated images. The Ouster LiDAR applied in the dataset is OS0-128 with its detailed specifications shown in Table~\ref{tab:ouster-table}.

\input{tbs/hardware-Ouster}

\begin{figure}
    \centering
    \includegraphics[width=0.48\textwidth]{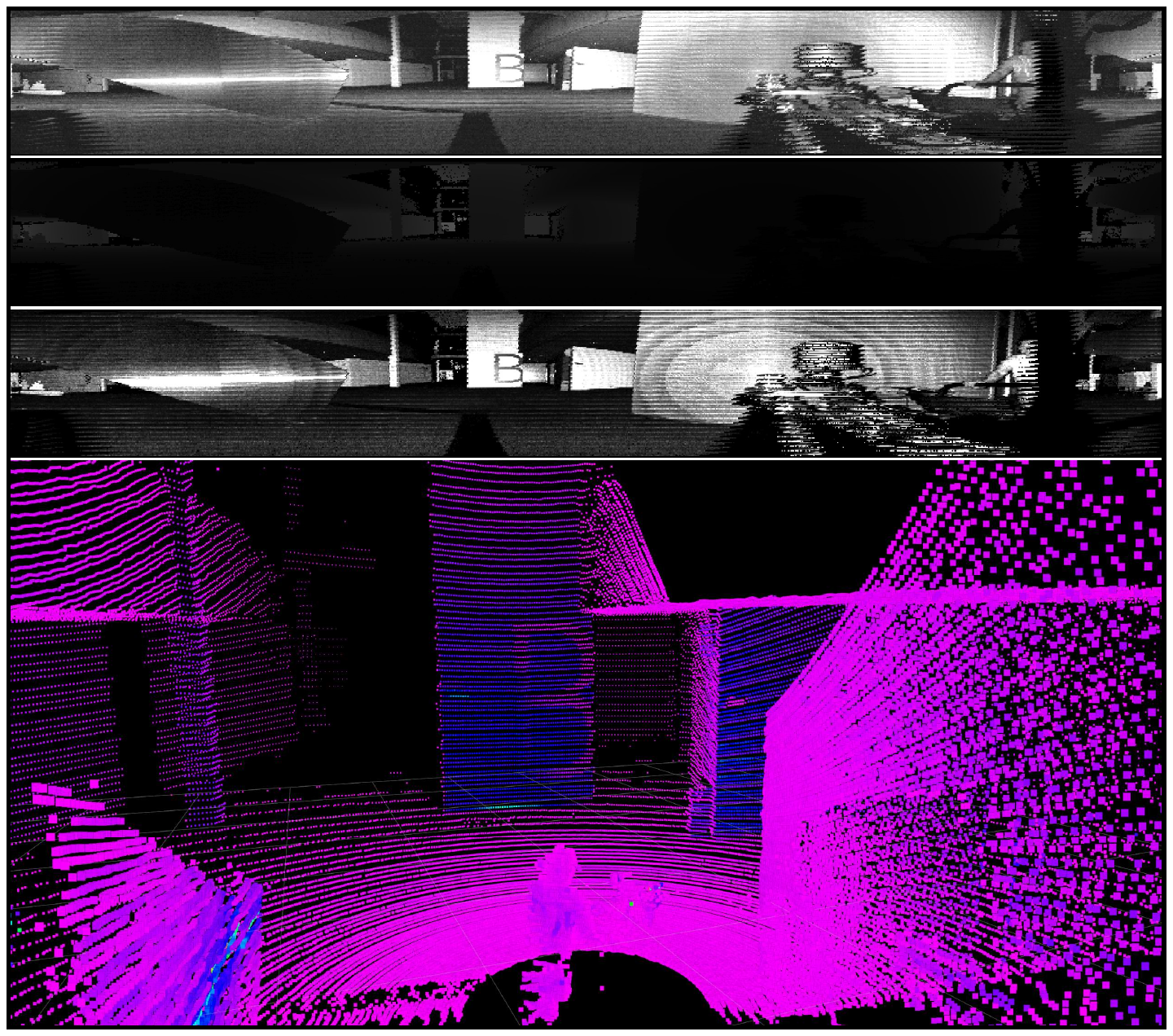}
    \caption{Samples of LiDAR-Generated Images, from above to bottom are signal image, range image, reflectivity image, and point cloud.}
    \label{fig:image-samples}
\end{figure}

The images generated by OS0-128 shown in Fig.~\ref{fig:image-samples} include signal images, reflectivity images, near-infrared images, and range images with its expansive $360^\circ \times 90^\circ$ field of view. Signal images are representations of the signal strength of the light returned to the sensor for a given point, which depends on various factors, such as the angle of incidence, the distance from the sensor, and the material properties of the object. In near-infrared images, each pixel's intensity is represented by the amount of detected photons that are not emitted by the sensor's own laser pulse but may come from sources such as sunlight or moonlight. And every pixel in a reflectivity image represents the calculated calibrated reflectivity. Then, range images demonstrate the distance from the sensor to objects in the environment. 

As indicated by the findings of our previous research, signal images have exhibited superior performance in the execution of conventional DL tasks within the domain of computer vision~\cite{yu2023general}. In light of this, for the first two parts of our experiment, we opt to employ signal images from the "indoor\_01\_square" scene provided by the dataset, which is a scene that spans 114 seconds and comprises 1146 image messages.

\subsection{Optimal Preprocessing Configuration Searching for LiDAR-Generated Images}
LiDAR-generated images at hand are typically panoramic but low-resolution. Moreover, these images often exhibit a substantial degree of noise. This prompts a concern of utilizing the original images for facilitating the functionality evaluation of the keypoint detector and descriptor algorithms.  
And our preliminary experiments have evinced unsatisfactory performance across an array of detectors and descriptors when employing the unaltered original LiDAR-generated images. 
To identify the optimal resolution and interpolation methodology for augmenting image resolution, an extensive comparative experiment was conducted.


\input{algs/alg-optimal-para}

In this part, we implement an array of interpolation techniques on the original images, employing an extensive spectrum of image resolution combinations. The interpolation methodologies encompass bicubic interpolation (CUBIC), Lanczos interpolation over 8x8 neighborhood (LANCZOS4), resampling using pixel area relation (AREA), nearest neighbor interpolation (NEAREST), and bilinear interpolation (LINEAR). The primary procedure of the preprocessing is elucidated in Algorithm~\ref{alg:optimal}.

More specifically, we iterate a range of image dimensions and interpolation methods in conjunction with the suite of detector and descriptor algorithms designated for evaluation. Each iteration involves a rigorous evaluation of a comprehensive metrics set detailed in Table~\ref{tab:metrics}. Following a quantitative analysis, we compute mean values for these metrics. This extensive assessment aims to identify the optimal preprocessing configuration that offers a balanced performance for different keypoint detectors and descriptors.


\subsection{Keypoint Detectors and Descriptors for LiDAR-Generated Images}
\label{subsec:key-det-des}
The evaluation workflow of detector-descriptor algorithms typically comprises three stages including feature extraction, keypoint description, and keypoint matching between successive image frames. In this section, the specific procedures for executing these stages in our experimental setup will be elaborated upon. 

\subsubsection{Designated Keypoint Detector and Descriptor}
An extensive array of keypoint detectors and descriptors, as detailed in Table~\ref{tab:detectors-and-descriptors} from Section~\ref{subsec:keypoint}, were investigated. The employed keypoint detectors include SHITOMASI, HARRIS, FAST, BRISK, SIFT, SURF, AKAZE, and ORB. Additionally, we integrated Superpoint, a DL-based keypoint detector, into our methodology.
The keypoint descriptors implemented in our experiment are  BRISK, SIFT, SURF, BRIEF, FREAK, AKAZE, ORB.
 

\subsubsection{Key Points Matching between Images}
\label{subsubsection:feature-matching}

Keypoint matching, the final stage of the detector-descriptor workflow, focuses on correlating key points between two images, which is essential for establishing spatial relationships and forming a coherent scene understanding. The smaller the distance of the descriptors between two points, the more likely it is that they are the same point or object between two images. In our implementation, we employ a technique termed ``brute-force match with cross check'', which means for a given descriptor \(\mathcal{D}_A\) in image \(A\) and another descriptor \(\mathcal{D}_B\) in image \(B\), a valid correspondence requires that both descriptors recognize each other as their closest descriptors.

\subsubsection{Selected Evaluation Metrics}
As explained in Section~\ref{subsec:metrics}, we have opted not to rely on ground truth-based evaluation methodologies due to the lack of benchmark datasets and the substantial labor involved in data labeling. Instead, we combined some specially-designed metrics that are independent of ground truth, together with several intuitive metrics, to form the complete indicators listed in Table~\ref{tab:metrics}. To our best understanding, this represents the most extensive set of evaluation metrics currently available in the absence of a benchmark dataset.

\subsubsection{Evaluation Process}

The flowchart shown in Algorithm~\ref{alg:workflow} below provides an outline of the steps carried out by the program. Two nested loops are employed to iterate over different detector-descriptor pairs. For each image, the algorithm detects and describes its keypoints. If more than one image has been processed, keypoints from the current image are matched to the previous one. And metrics are placed in corresponding positions to assess the algorithm's performance.

\input{algs/alg-keypoint-eval}

\subsection{LiDAR-Generated Image Keypoints Assisted Point Cloud Registration}
\subsubsection{Selected Data}
The selected data for the evaluation from the dataset mentioned in Section~\ref{subsec:dataset} includes indoor and outdoor environments. The outdoor environment are from the normal road, denoted as ``Open road'', and a forest, denoted as ``Forest''. The indoor data include a hall in a building, denoted as ``Hall (large)'', and two rooms, denoted as ``Lab space (hard)'', and ``Lab space (easy)''. 

\subsubsection{Point Cloud Matching Approach}
\label{subsec:pc_matching}
In this part, we applied KISS-ICP~\footnote{\url{https://github.com/PRBonn/kiss-icp.git}} as our point cloud matching approach. It provides also the odometry information, affording us the means to assess the efficacy of our point cloud downsampling approach through an examination of a positioning error, namely translation error and rotation error.
To generalize our proposed approach, we tested an NDT-based simple SLAM program~\footnote{\url{https://github.com/Kin-Zhang/simple_ndt_slam.git}} as well.

\subsubsection{Proposed Method for Point Cloud Downsampling}
\begin{figure}
    \centering
    \includegraphics[width=0.45\textwidth]{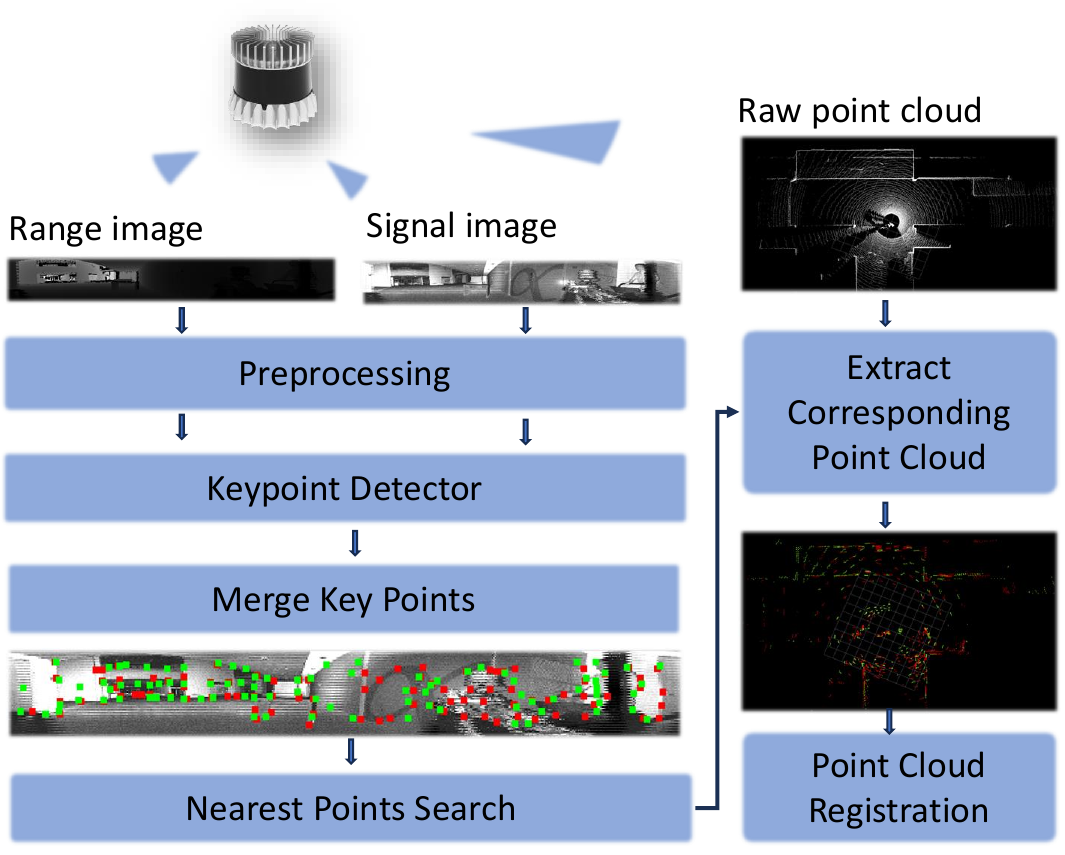}
    \caption{The process of the proposed LiDAR-generated images assisted point cloud registration}
    \label{fig:keypoint-lo}
\end{figure}

Following the preprocessing of LiDAR-generated images outlined in Section~\ref{subsec:key-det-des}, we derive optimal configurations for the keypoint detectors and descriptors. Utilizing these configurations as a foundation, we establish the workflow of our proposed methodology, illustrated in Fig.~\ref{fig:keypoint-lo}.
Within this process, we conduct distinct preprocessing procedures for both the range and signal images, employing them individually for keypoint detection and descriptor extraction. Subsequently, we combine the key points obtained from both images and search the \(K\) nearest points to each of these key points. We systematically varied \(K\) within the range of 3 to 7, adhering to a maximum threshold of 7 to align with our primary objective of downsampling the point cloud. Consequently, we find the corresponding point cloud of the key points and their neighbors within the raw point cloud, thereby constituting the downsampled point cloud.

In our analysis, we examined not only the positional error but also the rotational error, computational resource utilization, downsampling-induced alterations in point cloud density, and the publishing rate of LO.
\subsection{Hardware and Software Information}

Our experiments are run on the ROS Noetic on the Ubuntu 20.04 system. The platform is equipped with an i7 8-core 1.6\,GHz CPU and an Nvidia GeForce MX150 graphics card. Primarily, we used libraries like OpenCV and PCL. Note, that we have used some non-free copyright-protected algorithms from OpenCV, such as SURF, just for research.

The assessment of keypoint-based point cloud downsampling was conducted on a Lenovo Legion notebook equipped with the following specifications: 16\,GB RAM, a 6-core Intel i5-9300H processor (2.40\,GHz), and an Nvidia GTX 1660Ti graphics card (boasting 1536 CUDA cores and 6\,GB VRAM). Within this study, our primary focus was on the evaluation of the two open-source algorithms delineated in subsection~\ref{subsec:pc_matching}, namely, KISS-ICP and Simple-NDT-SLAM. It is imperative to highlight that a consistent voxel size of 0.2\,m was employed for both algorithms.
Our project is primarily written in C++ (including the DL approach, Superpoint), publicly available in GitHub~\footnote{\href{https://github.com/TIERS/ws-lidar-as-camera-odom}{https://github.com/TIERS/ws-lidar-as-camera-odom}}.

%% file: tbs/hardware-Ouster.tex
\begin{table}[htb]
\centering
\caption{Specifications of Ouster OS0-128.}
\label{tab:ouster-table}
\resizebox{0.48\textwidth}{!}{%
\begin{tabular}{@{}lccccccccc@{}}
\toprule
 &
  \textbf{IMU} &
  \textbf{Type} &
  \textbf{Channels} &
  \textbf{Image Resolution} &
  \textbf{FoV} &
  \textbf{Angular Resolution} &
  \textbf{Range} &
  \textbf{Freq} &
  \textbf{Points} \\ \midrule
\textbf{Ouster OS0-128} &
  ICM-20948 &
  spinning &
  128 &
  $ 2048 \times 128 $ &
  $360^\circ \times 90^\circ$ &
  $V:0.7^\circ, H:0.18^\circ$ &
  50m &
  10Hz &
  2,621,440 pts/s \\ \bottomrule
\end{tabular}%
}
\end{table}

%% file: algs/alg-optimal-para.tex
\begin{algorithm}[h]
\caption{Preprocessing configuration evaluation}
\label{alg:optimal}
\KwIn{\\
\(N\) number of signal images: \(\{S_i\}, i \sim N\);\\


Interpolation methods: \(IA = \{CUBIC, LANCZOS4, AREA,\) \\\(NEAREST, LINEAR\}\);\\

Targeted Width: \(TW = \{ min:512; max:4096; step:128 \}\);\\

Targeted Height: \(TH = \{ min:32; max:256; step:32 \}\);\\
}

Detectors and descriptors: \\
\(DET = \{SURF, SIFT, SHITOMASI, HARRIS, BRISK,\) \\\(FAST, AKAZE, ORB\}\);\\
\(DES = \{FREAK, SIFT, BRISK, SURF, BRIEF, AKAZE,\) \\\(ORB\}\);\\

\KwOut{Metrics}
\BlankLine

\ForEach{interplation approach in \(IA\)}{
    \ForEach{\(width\) in \(TW\)}{
        \ForEach{\(height\) in \(TH\)}{
            \ForEach{\(det\) in \(DET\) and \(des\) in \(DES\)}{
                \ForEach{\(S_i\)}{
                Calculate the value of aforementioned metrics; \\
                Save the calculated value;    
                }
            }
        }
    }
}
Analyze the metric values.

\end{algorithm}

%% file: algs/alg-keypoint-eval.tex
\begin{algorithm}
\caption{Overall evaluation pipeline of keypoint detectors and descriptors}
\label{alg:workflow}

\KwIn{\\
\(N\) number of signal images: \(\{S_i\}, i \sim N\);\\
\(DET = \{SURF, SIFT, SHITOMASI, HARRIS, BRISK,\) \\\(FAST, AKAZE, ORB\}\);\\
\(DES = \{FREAK, SIFT, BRISK, SURF, BRIEF, AKAZE,\) \\\(ORB\}\);\\
}
\KwOut{Metrics}
\BlankLine
\ForEach{Detector in \(DET\)}{
    \ForEach{Descriptor in \(DES\)}{
        \ForEach{\(S_i\)}{
            Preprocess the image \(S_i\) by the method in Algorithm~\ref{alg:optimal};\\
            Detect keypoints;\\
            Measure the number of keypoints: \(N_{kp}\);\\
            Apply different transformations to image \(S_i\), then calculate robustness: \(R_{rot}\), \(R_{scale}\), \(R_{blur}\);\\
            \If{\(i>1\)}{
                Match keypoints between successive frames;\\
                Record the algorithm running time \(RT\) as Computational Efficiency;\\
                Calculate Match Ratio, Match Score, Distinctiveness;\\
            }
        }
    }
}

Analyze the metric values.
\end{algorithm}

%% file: sections/04-experiment_result.tex
\section{Experiment Result}
\label{sec:experi-results}

Through this section, we first cover the final results of our exploration of the preprocessing workflow of LiDAR-based images. Subsequently, an in-depth analysis of keypoint detectors and descriptors for LiDAR-based images is conducted. Then, a detailed quantitative assessment of the performance of LO facilitated by LiDAR-generated image keypoints is presented.

\subsection{Results of preprocessing methods for LiDAR-generated image}

\input{tbs/res-inter-result}

As elucidated in Section~\ref{subsec:metrics}, Distinctiveness and Match Score are considered as paramount measures for the overall accuracy of the entire algorithm pipeline. Consequently, in scenarios where different sizes and interpolation methods show peak performance on different metrics, these two metrics are our primary concern. Based on such an criteria, the size \textbf{1024 x 64} demonstrated better performance across all dectectors and descriptors methods. Then in Table~\ref{tab:res-inter-result}, our evaluation also revealed that the \textbf{linear} interpolation method yielded the most optimal results among the various interpolation techniques.  

\input{tbs/res-resolution-result}

The findings in Table~\ref{tab:res-resolution-result}, also suggest that there is a clear advantage in properly reducing the size of an image as opposed to enlarging it. Additionally, in the process of image downscaling, one pixel often corresponds to several pixels in original image. So overly downscaled images might lead to substantial deviations in the detected key points when re-projected to their original poistions, suggesting that extreme image size reductions should be avoided. 

\begin{figure}[h]
\centering
    \begin{subfigure}{0.45\textwidth}
        \centering
        \includegraphics[width=\textwidth, height=0.15\textwidth]{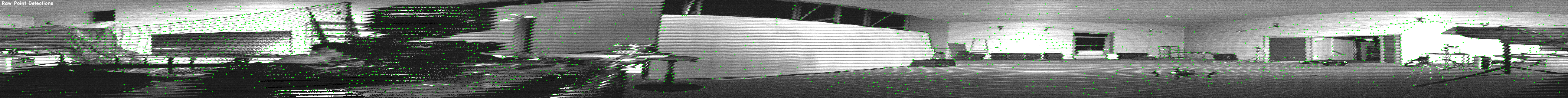}
        \caption{Detect key points in an enlarged image.}
        \label{fig:detect-keypoints-in-a-enlarged-image}
    \end{subfigure}\\[\baselineskip]

    \begin{subfigure}{0.45\textwidth}
       \centering
       \includegraphics[width=\textwidth, height=0.15\textwidth]{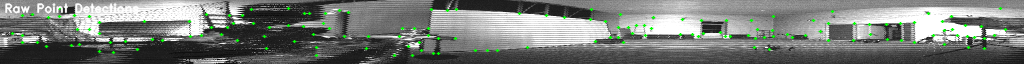}
       \caption{Detect key points in a downscaled image.}
       \label{fig:detect-keypoints-in-a-downscaled-image} 
    \end{subfigure}
    \caption{Keypoint detected in the resized signal images}
\end{figure}

And here is a more intuitive result to show that how reducing the size of a image is far better than enlarging it. In Fig.~\ref{fig:detect-keypoints-in-a-enlarged-image} and Fig.~\ref{fig:detect-keypoints-in-a-downscaled-image}, Superpoint detectors identify keypoints as green dots. The enlarged image Fig.~\ref{fig:detect-keypoints-in-a-enlarged-image} displays many disorganized points. Conversely, the downscaled image Fig.~\ref{fig:detect-keypoints-in-a-enlarged-image}, reveals distinct keypoints, such as room corners and the points where various planes of objects meet. Note that we resized the two images for paper readability, originally, their sizes varied.

\subsection{Results of Keypoint Detectors and Descriptors For LiDAR Image}

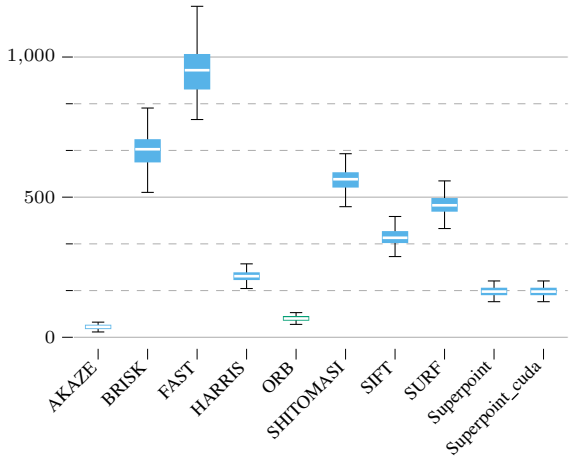
\begin{figure}[H]
    \centering
    \setlength{\figurewidth}{0.45\textwidth}
    \setlength{\figureheight}{0.35\textwidth}
    \scriptsize{\input{tex/1_Keypoints_number}}
    \caption{Number of key points}
    \label{fig:Keypoints_number}
\end{figure}

In Fig.~\ref{fig:Keypoints_number}, which is a  metric that only related to detectors, FAST and BRISK algorithms detected the highest number of keypoints, but there were significant fluctuations in the counts. Comparatively, AKAZE, ORB, and Superpoint identified a reduced number of keypoints, but the consistency was notable. 

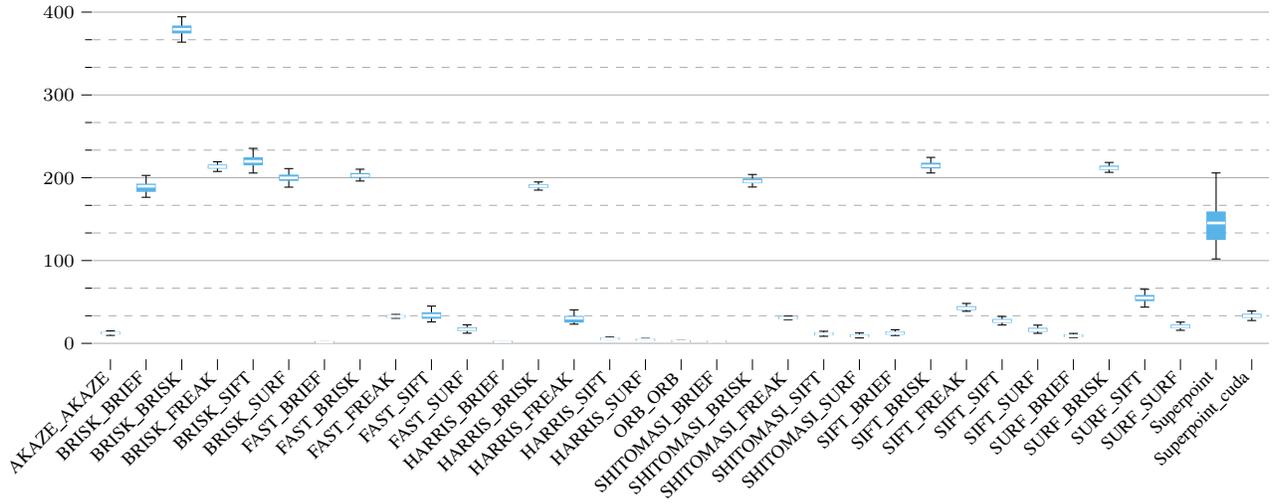
\begin{figure*}
    \centering
    \setlength{\figurewidth}{0.95\textwidth}
    \setlength{\figureheight}{0.35\textwidth}
    \scriptsize{\input{tex/3_Computational_efficiency}}
    \caption{Computational efficiency}
    \label{fig:Computational_efficiency}
\end{figure*}

Fig.~\ref{fig:Computational_efficiency} depicts the Computational Efficiency, where the majority of the algorithms operate in less than 50 ms. After CUDA enabled, SuperPoint runs significantly faster with minimal variance. Among all algorithms, BRISK is the most time-consuming, just using BRISK solely as a descriptor with other detectors will hinder the overall efficiency. 

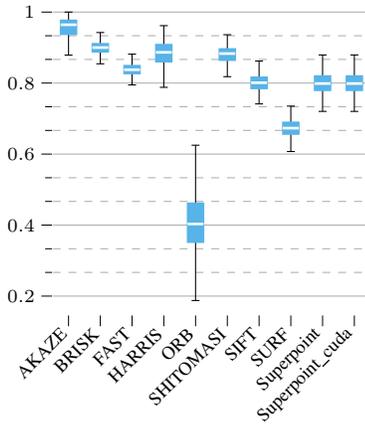
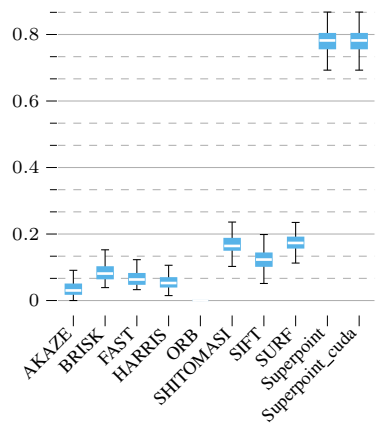
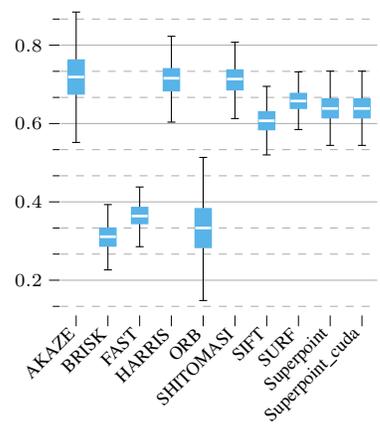
\begin{figure*}[htbp]
    \begin{subfigure}{.32\textwidth}
        \centering
        \setlength{\figurewidth}{\textwidth}
        \setlength{\figureheight}{\textwidth}
        \scriptsize{\input{tex/2_1_Robustness_rotated}}
        \caption{\scriptsize{Rotation}}
        \label{fig:Robustness_rotated}
    \end{subfigure}
    \hfill
    \begin{subfigure}{.32\textwidth}
        \centering
        \setlength{\figurewidth}{\textwidth}
        \setlength{\figureheight}{\textwidth}
        \scriptsize{\input{tex/2_2_Robustness_scaling}}
        \caption{\scriptsize{Scaling}}
        \label{fig:Robustness_scaling}     
    \end{subfigure}
    \hfill
    \begin{subfigure}{.32\textwidth}
        \centering
        \setlength{\figurewidth}{\textwidth}
        \setlength{\figureheight}{\textwidth}
        \scriptsize{\input{tex/2_3_Robustness_blurred}}
        \caption{\scriptsize{Noise Interference}}
        \label{fig:Robustness_blurred}     
    \end{subfigure}
    \caption{Robustness of the detector.}
    \label{fig:Robustness} 
\end{figure*}

Fig.~\ref{fig:Robustness} shows the Robustness of Detector. Superpoint consistently demonstrates robust performance across various transformations. Among conventional detectors, AKAZE has proven effective, especially in handling rotated transformations and noise interference. And most detectors, exhibit marked poor performance under scale invariance. The horizontal textures inherent in LiDAR-based images might explain such weakness: when the images are enlarged, these textures can be erroneously detected as keypoints. 

As emphasized in Section~\ref{subsec:metrics}, a multitude of keypoint detections and rapid matches could be useless if their accuracy is not guaranteed. Therefore, Match Ratio, Match Score, and Dinctiveness, which pertain to algorithmic accuracy, can be regarded as the most pivotal indicators across various application perspectives. Fig.~\ref{fig:Match_Ratio}, Fig.~\ref{fig:Match_Score}, and Fig.~\ref{fig:Distinctiveness} present the results of these three metrics, indicating that Superpoint, when augmented with CUDA, is the most effective solution. Moreover, among traditional algorithms, AKAZE demonstrates top-tier performance across the majority of evaluated metrics, making it a commendable choice.

\begin{figure*}[H]
    \centering
    \setlength{\figurewidth}{0.9\textwidth}
    \setlength{\figureheight}{0.45\textwidth}
    \scriptsize{\input{tex/4_Match_Ratio}}
    \caption{Match ratio}
    \label{fig:Match_Ratio}
\end{figure*}

\begin{figure*}[H]
    \centering
    \setlength{\figurewidth}{0.9\textwidth}
    \setlength{\figureheight}{0.45\textwidth}
    \scriptsize{\input{tex/5_Match_Score}}
    \caption{Match score}
    \label{fig:Match_Score}
\end{figure*}

\begin{figure*}[H]
    \centering
    \setlength{\figurewidth}{0.9\textwidth}
    \setlength{\figureheight}{0.45\textwidth}
    \scriptsize{\input{tex/6_Distinctiveness}}
    \caption{Distinctiveness}
    \label{fig:Distinctiveness}
\end{figure*}




\subsection{Results of LiDAR-generated Image Keypoints Assisted LO}

\subsubsection{Downsampled Point Cloud}
\begin{figure*}[H]
    \centering
    \begin{subfigure}{0.95\textwidth}
        \centering
        \includegraphics[width=\textwidth]{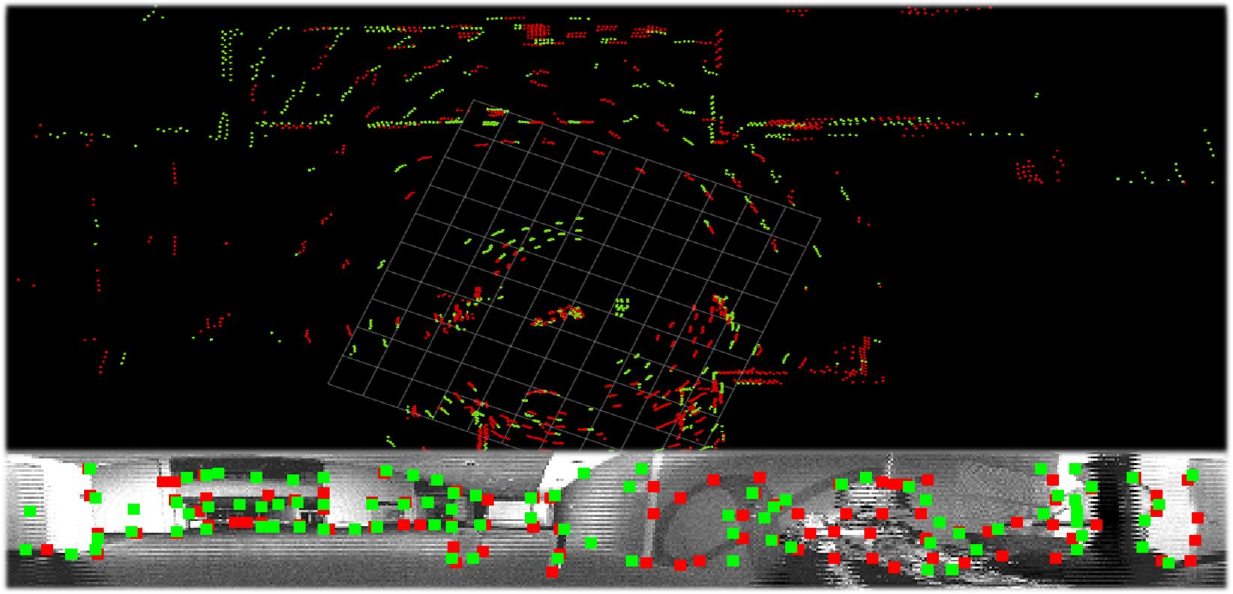}
        \caption{The upper part presents the downsampled point cloud from our lidar-based method, while the bottom illustrates key point distribution in the signal image: red from the signal image and green from the distance image.}
        \label{fig:downsampled-pointcloud}
    \end{subfigure} \\ [\baselineskip]

    \begin{subfigure}{0.95\textwidth}
        \includegraphics[width=\textwidth]{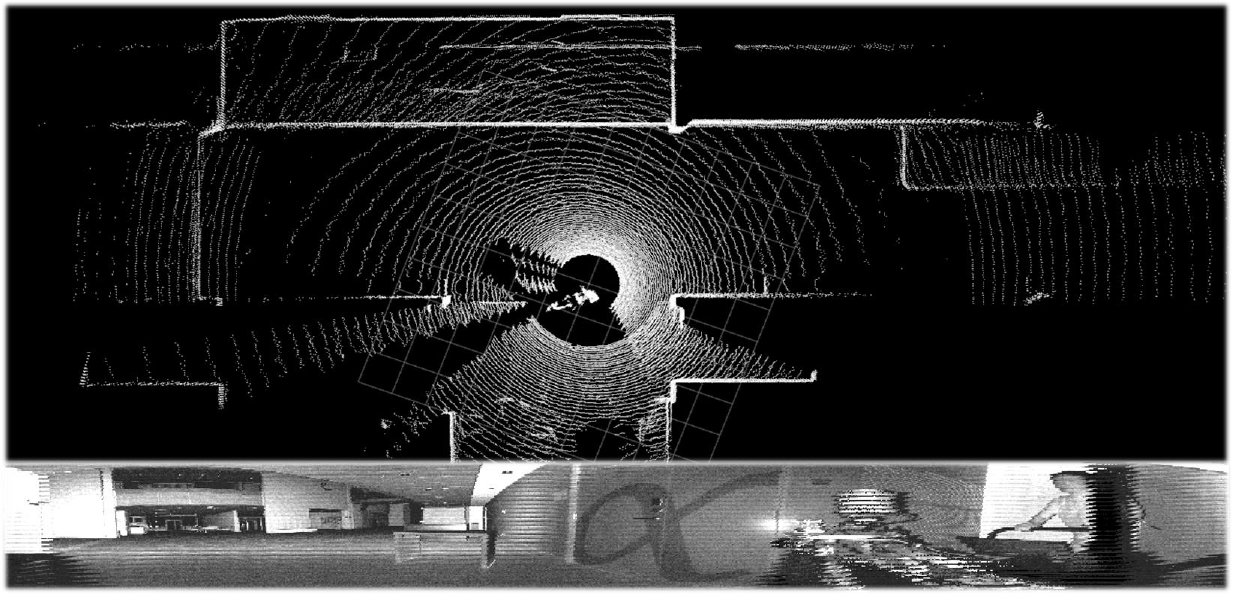}
        \caption{Raw point cloud and signal image.}
        \label{fig:raw-pointcloud}
    \end{subfigure}
     \caption{Samples of point cloud data}
    \label{fig:sample-pointcloud}
\end{figure*}

In Fig.~\ref{fig:sample-pointcloud}, we demonstrate the sample result of the downsampled point cloud in Fig.~\ref{fig:downsampled-pointcloud}  compared with the raw point cloud in Fig.~\ref{fig:raw-pointcloud}.  Notably, in the downsampled point cloud in Fig.~\ref{fig:downsampled-pointcloud}, the red points are extracted based on signal images and the green ones are from range images. 
We draw the key points from both images to the signal image shown in the lower part of Fig.~\ref{fig:downsampled-pointcloud}.
The disparity between the points extracted based on these two types of images shows the significance of different LiDAR-generated images. Additionally, in the preliminary evaluation of LO, we found the accuracy of LO is lower if we only integrated the signal images instead of both modalities. This encourages us to utilize both signal and range images in the latter part.

\subsubsection{LO based Evaluation}
In our experiment, various numbers of neighbor points are utilized, ranging from 3 to 7 for each type of LiDAR-generated image. We selected part of them to show the result here based on the principle that more accurate but less amount points. As we found in the previous section, the Superpoint has reliable key points detected, so we utilize this DL method to extract key points in our proposed approach while KISS-ICP is the point cloud registration and LO method. Table~\ref{tab:kissicp-size-evaluate} shows the performance of LO based on different sizes of neighbor point sizes in both indoor (Lab space, Hall) and outdoor (Open road and Forest) environments.

\input{tbs/kissicp-neighbor-evaluate}

As shown in Table~\ref{tab:kissicp-size-evaluate}, in the scenarios of Open road, Lab space (hard), and Hall(Large), the LO from KISS-ICP applying raw point cloud can not work properly with large drift which the error can not be calculated. Meanwhile, our proposed approach works all the time. Additionally, even when applying raw point cloud to KISS-ICP works, our approach can achieve comparable translation state estimation while more robust in the rotation state estimation across most of the situations. 


In outdoor settings, a neighbor size \(4\_7\) (\(4\times4\) for signal images and \(7\times7\) for range images) exhibits notable efficacy in both translation and rotation state estimation. Conversely, in indoor environments, a neighbor size \(5\_5\) (\(5\times5\) for signal images and \(5\times5\) for range images) demonstrates commendable performance in the estimation of translation and rotation states, in addition to exhibiting efficient downsampling capabilities, as delineated in Table~\ref{tab:kissicp-size-evaluate}.

\input{tbs/kiss-points}

Based on the above result, we apply the neighbor size \(4\_7\) for outdoor settings and the neighbor size \(5\_5\) for indoor settings to further extend the performance evaluation by including the conventional keypoint detector approach and another point cloud matching approach, NDT. It is worth noting that the purpose of applying NDT here is not to compare with KISS-ICP but to show the generalization of our proposed approach among other point cloud registration methods. 

The result in Table.~\ref{tab:kiss-icp-perf} and Table.~\ref{tab:kiss-ndt-evaluate} proves that the conventional keypoint extractor can achieve comparable LO translation estimation and more accurate rotation estimation to Superpoint. This performance is obtained with much less CPU and memory utilization, and fewer cloud points, but higher odometry publishing rates.  Similar results are achieved by NDT based approach which validates the above result in a certain way. 
Notably, the memory consumption using KISS-ICP with raw point cloud in Table.~\ref{tab:kiss-icp-perf} is lower than others. As our observation indicates, the primary reason behind this is the drift, resulting in few points for the point cloud registration.

\input{tbs/kiss-icp-perf}

\input{tbs/kiss-ndt-evaluate}

%% file: tbs/res-inter-result.tex
\begin{table}[!ht]
    \centering
    \caption{Evaluation metrics under different interpolation approaches.}
    \label{tab:res-inter-result}
\resizebox{0.48\textwidth}{!}{%
    \begin{tabular}{cccc}
    \toprule
        \textbf{Interpolation}  & \textbf{Robustness of (rotation, scaling, noise)} & \textbf{Distinctiveness} & \textbf{Matching Score} \\ 
        \midrule
        AREA & (0.81,0.106,0.574) & 0.309 & 0.415 \\ 
        CUBIC & (\textbf{0.82},0.121,0.569) & 0.292 & 0.408 \\ 
        LANCZOS4 & (0.819,\textbf{0.127},0.559) & 0.286 & 0.405 \\ 
        NEAREST & (0.818,0.128,0.573) & 0.275 & 0.401 \\ 
        LINEAR & (0.815,0.1,\textbf{0.583}) & \textbf{0.314} & \textbf{0.415} \\
        \bottomrule
    \end{tabular}
    }
\end{table}

%% file: tbs/res-resolution-result.tex
\begin{table}[!ht]
    \centering
    \caption{Evaluation metrics under different resized resolutions.}
    \label{tab:res-resolution-result}
\resizebox{0.48\textwidth}{!}{%
    \begin{tabular}{cccc}
         \toprule
         \textbf{Size} & \textbf{Robustness of (rotation, scaling, noise)} & \textbf{Distinctiveness} & \textbf{Matching Score} \\ 
         \midrule
        512\texttimes32 & (\textbf{0.856},\textbf{0.157},0.551) & 0.267 & 0.366 \\ 
        896\texttimes128 & (0.827,0.156,0.591) & 0.37 & 0.515 \\ 
        896\texttimes256 & (0.843,0.146,\textbf{0.663}) & 0.309 & 0.486 \\ 
        1024\texttimes64 & (0.809,0.124,0.54) & \textbf{0.427} & \textbf{0.53} \\ 
        1024\texttimes128 & (0.832,0.147,0.584) & 0.372 & 0.504 \\ 
        1024\texttimes256 & (0.851,0.134,0.659) & 0.32 & 0.483 \\ 
        1280\texttimes64 & (0.798,0.116,0.527) & 0.41 & 0.5 \\ 
        1280\texttimes128 & (0.823,0.138,0.575) & 0.353 & 0.479 \\ 
        1280\texttimes256 & (0.849,0.127,0.652) & 0.301 & 0.464 \\ 
        1920\texttimes128 & (0.808,0.124,0.553) & 0.321 & 0.436 \\ 
        1920\texttimes256 & (0.844,0.114,0.644) & 0.274 & 0.43 \\ 
        2048\texttimes128 & (0.799,0.129,0.544) & 0.309 & 0.425 \\ 
        2048\texttimes256 & (0.837,0.119,0.633) & 0.263 & 0.421 \\ 
        2560\texttimes128 & (0.802,0.111,0.547) & 0.294 & 0.4 \\ 
        2560\texttimes256 & (0.842,0.106,0.644) & 0.249 & 0.401 \\ 
        4096\texttimes128 & (0.799,0.087,0.557) & 0.257 & 0.339 \\
        \bottomrule
    \end{tabular}
    }
\end{table}

%% file: tex/1_Keypoints_number.tex
\begin{tikzpicture}

\definecolor{darkgoldenrod}{RGB}{184,134,11}
\definecolor{darkgray176}{RGB}{176,176,176}
\definecolor{firebrick}{RGB}{178,34,34}
\definecolor{steelblue31119180}{RGB}{31,119,180}

\definecolor{color0}{rgb}{0.90, 0.62, 0.00}
\definecolor{color1}{rgb}{0.34, 0.70, 0.91}
\definecolor{color2}{rgb}{0.00, 0.62, 0.45}
\definecolor{color3}{rgb}{0.94, 0.89, 0.27}
\definecolor{color4}{rgb}{0.00, 0.45, 0.69}
\definecolor{color5}{rgb}{0.83, 0.37, 0.00}

\begin{axis}[
width=\figurewidth,
height=\figureheight,
axis line style={white},
tick align=outside,
tick pos=left,
x grid style={darkgray176},
xmin=0.5, xmax=10.5,
xtick style={color=black},
xtick={1,2,3,4,5,6,7,8,9,10},
xticklabel style={rotate=45.0,anchor=east},
xticklabels={AKAZE,BRISK,FAST,HARRIS,ORB,SHITOMASI,SIFT,SURF,Superpoint,Superpoint\_cuda},
y grid style={darkgray176},
ymin=-39.1, ymax=1239.1,
ytick style={color=black},
ymajorgrids,
ymajorticks=true,
minor y tick num = 2,
minor y grid style={dashed},
yminorgrids,
]
\addplot [black]
table {%
1 32
1 19
};
\addplot [black]
table {%
1 42
1 54
};
\addplot [black]
table {%
0.875 19
1.125 19
};
\addplot [black]
table {%
0.875 54
1.125 54
};
\addplot [black]
table {%
2 626.75
2 517
};
\addplot [black]
table {%
2 704
2 818
};
\addplot [black]
table {%
1.875 517
2.125 517
};
\addplot [black]
table {%
1.875 818
2.125 818
};
\addplot [black]
table {%
3 887
3 777
};
\addplot [black]
table {%
3 1008
3 1181
};
\addplot [black]
table {%
2.875 777
3.125 777
};
\addplot [black]
table {%
2.875 1181
3.125 1181
};
\addplot [black]
table {%
4 207
4 174
};
\addplot [black]
table {%
4 229
4 262
};
\addplot [black]
table {%
3.875 174
4.125 174
};
\addplot [black]
table {%
3.875 262
4.125 262
};
\addplot [black]
table {%
5 62
5 46
};
\addplot [black]
table {%
5 73
5 88
};
\addplot [black]
table {%
4.875 46
5.125 46
};
\addplot [black]
table {%
4.875 88
5.125 88
};
\addplot [black]
table {%
6 537
6 466
};
\addplot [black]
table {%
6 586
6 655
};
\addplot [black]
table {%
5.875 466
6.125 466
};
\addplot [black]
table {%
5.875 655
6.125 655
};
\addplot [black]
table {%
7 339
7 288
};
\addplot [black]
table {%
7 376
7 431
};
\addplot [black]
table {%
6.875 288
7.125 288
};
\addplot [black]
table {%
6.875 431
7.125 431
};
\addplot [black]
table {%
8 451
8 388
};
\addplot [black]
table {%
8 494
8 558
};
\addplot [black]
table {%
7.875 388
8.125 388
};
\addplot [black]
table {%
7.875 558
8.125 558
};
\addplot [black]
table {%
9 153
9 127
};
\addplot [black]
table {%
9 174
9 201
};
\addplot [black]
table {%
8.875 127
9.125 127
};
\addplot [black]
table {%
8.875 201
9.125 201
};
\addplot [black]
table {%
10 153
10 127
};
\addplot [black]
table {%
10 174
10 201
};
\addplot [black]
table {%
9.875 127
10.125 127
};
\addplot [black]
table {%
9.875 201
10.125 201
};
\path [draw=color1, fill=color1, semithick]
(axis cs:0.75,32)
--(axis cs:1.25,32)
--(axis cs:1.25,42)
--(axis cs:0.75,42)
--(axis cs:0.75,32)
--cycle;
\path [draw=color1, fill=color1, semithick]
(axis cs:1.75,626.75)
--(axis cs:2.25,626.75)
--(axis cs:2.25,704)
--(axis cs:1.75,704)
--(axis cs:1.75,626.75)
--cycle;
\path [draw=color1, fill=color1, semithick]
(axis cs:2.75,887)
--(axis cs:3.25,887)
--(axis cs:3.25,1008)
--(axis cs:2.75,1008)
--(axis cs:2.75,887)
--cycle;
\path [draw=color1, fill=color1, semithick]
(axis cs:3.75,207)
--(axis cs:4.25,207)
--(axis cs:4.25,229)
--(axis cs:3.75,229)
--(axis cs:3.75,207)
--cycle;
\path [draw=color2, fill=color2, semithick]
(axis cs:4.75,62)
--(axis cs:5.25,62)
--(axis cs:5.25,73)
--(axis cs:4.75,73)
--(axis cs:4.75,62)
--cycle;
\path [draw=color1, fill=color1, semithick]
(axis cs:5.75,537)
--(axis cs:6.25,537)
--(axis cs:6.25,586)
--(axis cs:5.75,586)
--(axis cs:5.75,537)
--cycle;
\path [draw=color1, fill=color1, semithick]
(axis cs:6.75,339)
--(axis cs:7.25,339)
--(axis cs:7.25,376)
--(axis cs:6.75,376)
--(axis cs:6.75,339)
--cycle;
\path [draw=color1, fill=color1, semithick]
(axis cs:7.75,451)
--(axis cs:8.25,451)
--(axis cs:8.25,494)
--(axis cs:7.75,494)
--(axis cs:7.75,451)
--cycle;
\path [draw=color1, fill=color1, semithick]
(axis cs:8.75,153)
--(axis cs:9.25,153)
--(axis cs:9.25,174)
--(axis cs:8.75,174)
--(axis cs:8.75,153)
--cycle;
\path [draw=color1, fill=color1, semithick]
(axis cs:9.75,153)
--(axis cs:10.25,153)
--(axis cs:10.25,174)
--(axis cs:9.75,174)
--(axis cs:9.75,153)
--cycle;
\addplot [line width=1pt, white]
table {%
0.75 37
1.25 37
};
\addplot [line width=1pt, white]
table {%
1.75 671
2.25 671
};
\addplot [line width=1pt, white]
table {%
2.75 953
3.25 953
};
\addplot [line width=1pt, white]
table {%
3.75 218
4.25 218
};
\addplot [line width=1pt, white]
table {%
4.75 67
5.25 67
};
\addplot [line width=1pt, white]
table {%
5.75 564
6.25 564
};
\addplot [line width=1pt, white]
table {%
6.75 355
7.25 355
};
\addplot [line width=1pt, white]
table {%
7.75 471
8.25 471
};
\addplot [line width=1pt, white]
table {%
8.75 163
9.25 163
};
\addplot [line width=1pt, white]
table {%
9.75 163
10.25 163
};
\end{axis}

\end{tikzpicture}

%% file: tex/3_Computational_efficiency.tex
\begin{tikzpicture}

\definecolor{darkgoldenrod}{RGB}{184,134,11}
\definecolor{darkgray176}{RGB}{176,176,176}
\definecolor{firebrick}{RGB}{178,34,34}
\definecolor{steelblue31119180}{RGB}{31,119,180}
\definecolor{color0}{rgb}{0.90, 0.62, 0.00}
\definecolor{color1}{rgb}{0.34, 0.70, 0.91}
\definecolor{color2}{rgb}{0.00, 0.62, 0.45}
\definecolor{color3}{rgb}{0.94, 0.89, 0.27}
\definecolor{color4}{rgb}{0.00, 0.45, 0.69}
\definecolor{color5}{rgb}{0.83, 0.37, 0.00}

\begin{axis}[
width=\figurewidth,
height=\figureheight,
axis line style={white},
tick align=outside,
tick pos=left,
x grid style={darkgray176},
xmin=0.5, xmax=33.5,
xtick style={color=black},
xtick={1,2,3,4,5,6,7,8,9,10,11,12,13,14,15,16,17,18,19,20,21,22,23,24,25,26,27,28,29,30,31,32,33},
xticklabel style={rotate=45.0,anchor=east},
xticklabels={
  AKAZE\_AKAZE,
  BRISK\_BRIEF,
  BRISK\_BRISK,
  BRISK\_FREAK,
  BRISK\_SIFT,
  BRISK\_SURF,
  FAST\_BRIEF,
  FAST\_BRISK,
  FAST\_FREAK,
  FAST\_SIFT,
  FAST\_SURF,
  HARRIS\_BRIEF,
  HARRIS\_BRISK,
  HARRIS\_FREAK,
  HARRIS\_SIFT,
  HARRIS\_SURF,
  ORB\_ORB,
  SHITOMASI\_BRIEF,
  SHITOMASI\_BRISK,
  SHITOMASI\_FREAK,
  SHITOMASI\_SIFT,
  SHITOMASI\_SURF,
  SIFT\_BRIEF,
  SIFT\_BRISK,
  SIFT\_FREAK,
  SIFT\_SIFT,
  SIFT\_SURF,
  SURF\_BRIEF,
  SURF\_BRISK,
  SURF\_SIFT,
  SURF\_SURF,
  Superpoint,
  Superpoint\_cuda
},
y grid style={darkgray176},
ymin=-18.59986165, ymax=413.97268865,
ytick style={color=black},
ymajorgrids,
ymajorticks=true,
minor y tick num = 2,
minor y grid style={dashed},
yminorgrids,
]
\addplot [black]
table {%
1 11.70028425
1 9.643115
};
\addplot [black]
table {%
1 13.073789
1 15.096856
};
\addplot [black]
table {%
0.875 9.643115
1.125 9.643115
};
\addplot [black]
table {%
0.875 15.096856
1.125 15.096856
};
\addplot [black]
table {%
2 183.89566725
2 176.402697
};
\addplot [black]
table {%
2 191.88633275
2 202.792253
};
\addplot [black]
table {%
1.875 176.402697
2.125 176.402697
};
\addplot [black]
table {%
1.875 202.792253
2.125 202.792253
};
\addplot [black]
table {%
3 375.2055225
3 363.76456
};
\addplot [black]
table {%
3 382.8742725
3 394.3103
};
\addplot [black]
table {%
2.875 363.76456
3.125 363.76456
};
\addplot [black]
table {%
2.875 394.3103
3.125 394.3103
};
\addplot [black]
table {%
4 211.94901175
4 207.587539
};
\addplot [black]
table {%
4 214.92333475
4 219.310377
};
\addplot [black]
table {%
3.875 207.587539
4.125 207.587539
};
\addplot [black]
table {%
3.875 219.310377
4.125 219.310377
};
\addplot [black]
table {%
5 216.0403925
5 205.683
};
\addplot [black]
table {%
5 223.8317275
5 235.46849
};
\addplot [black]
table {%
4.875 205.683
5.125 205.683
};
\addplot [black]
table {%
4.875 235.46849
5.125 235.46849
};
\addplot [black]
table {%
6 197.15223
6 188.63047
};
\addplot [black]
table {%
6 202.8341525
6 211.06955
};
\addplot [black]
table {%
5.875 188.63047
6.125 188.63047
};
\addplot [black]
table {%
5.875 211.06955
6.125 211.06955
};
\addplot [black]
table {%
7 1.3900025
7 1.062527
};
\addplot [black]
table {%
7 1.653393
7 2.043197
};
\addplot [black]
table {%
6.875 1.062527
7.125 1.062527
};
\addplot [black]
table {%
6.875 2.043197
7.125 2.043197
};
\addplot [black]
table {%
8 200.89312
8 196.02887
};
\addplot [black]
table {%
8 204.655735
8 210.2818
};
\addplot [black]
table {%
7.875 196.02887
8.125 196.02887
};
\addplot [black]
table {%
7.875 210.2818
8.125 210.2818
};
\addplot [black]
table {%
9 31.95269
9 30.2887
};
\addplot [black]
table {%
9 33.1604925
9 34.94671
};
\addplot [black]
table {%
8.875 30.2887
9.125 30.2887
};
\addplot [black]
table {%
8.875 34.94671
9.125 34.94671
};
\addplot [black]
table {%
10 30.741925
10 25.9092
};
\addplot [black]
table {%
10 36.590175
10 44.9921
};
\addplot [black]
table {%
9.875 25.9092
10.125 25.9092
};
\addplot [black]
table {%
9.875 44.9921
10.125 44.9921
};
\addplot [black]
table {%
11 15.86009
11 12.26816
};
\addplot [black]
table {%
11 18.525365
11 22.4329
};
\addplot [black]
table {%
10.875 12.26816
11.125 12.26816
};
\addplot [black]
table {%
10.875 22.4329
11.125 22.4329
};
\addplot [black]
table {%
12 1.794938
12 1.602358
};
\addplot [black]
table {%
12 2.11381725
12 2.589345
};
\addplot [black]
table {%
11.875 1.602358
12.125 1.602358
};
\addplot [black]
table {%
11.875 2.589345
12.125 2.589345
};
\addplot [black]
table {%
13 188.65909575
13 185.02339
};
\addplot [black]
table {%
13 191.172125
13 194.911742
};
\addplot [black]
table {%
12.875 185.02339
13.125 185.02339
};
\addplot [black]
table {%
12.875 194.911742
13.125 194.911742
};
\addplot [black]
table {%
14 26.088098
14 23.180006
};
\addplot [black]
table {%
14 32.2198855
14 40.493329
};
\addplot [black]
table {%
13.875 23.180006
14.125 23.180006
};
\addplot [black]
table {%
13.875 40.493329
14.125 40.493329
};
\addplot [black]
table {%
15 5.030363
15 4.302946
};
\addplot [black]
table {%
15 6.041639
15 7.55087
};
\addplot [black]
table {%
14.875 4.302946
15.125 4.302946
};
\addplot [black]
table {%
14.875 7.55087
15.125 7.55087
};
\addplot [black]
table {%
16 3.88785775
16 3.288434
};
\addplot [black]
table {%
16 4.709887
16 5.942233
};
\addplot [black]
table {%
15.875 3.288434
16.125 3.288434
};
\addplot [black]
table {%
15.875 5.942233
16.125 5.942233
};
\addplot [black]
table {%
17 2.4651345
17 2.048268
};
\addplot [black]
table {%
17 2.90320175
17 3.556133
};
\addplot [black]
table {%
16.875 2.048268
17.125 2.048268
};
\addplot [black]
table {%
16.875 3.556133
17.125 3.556133
};
\addplot [black]
table {%
18 2.01009225
18 1.687362
};
\addplot [black]
table {%
18 2.30050975
18 2.726169
};
\addplot [black]
table {%
17.875 1.687362
18.125 1.687362
};
\addplot [black]
table {%
17.875 2.726169
18.125 2.726169
};
\addplot [black]
table {%
19 194.2868475
19 188.78382
};
\addplot [black]
table {%
19 198.180745
19 203.94356
};
\addplot [black]
table {%
18.875 188.78382
19.125 188.78382
};
\addplot [black]
table {%
18.875 203.94356
19.125 203.94356
};
\addplot [black]
table {%
20 30.21459525
20 28.654897
};
\addplot [black]
table {%
20 31.30356
20 32.931898
};
\addplot [black]
table {%
19.875 28.654897
20.125 28.654897
};
\addplot [black]
table {%
19.875 32.931898
20.125 32.931898
};
\addplot [black]
table {%
21 10.6162425
21 8.54346
};
\addplot [black]
table {%
21 12.21573
21 14.61074
};
\addplot [black]
table {%
20.875 8.54346
21.125 8.54346
};
\addplot [black]
table {%
20.875 14.61074
21.125 14.61074
};
\addplot [black]
table {%
22 8.235085
22 6.54651
};
\addplot [black]
table {%
22 9.9443425
22 12.50495
};
\addplot [black]
table {%
21.875 6.54651
22.125 6.54651
};
\addplot [black]
table {%
21.875 12.50495
22.125 12.50495
};
\addplot [black]
table {%
23 11.432842
23 9.508887
};
\addplot [black]
table {%
23 13.43078225
23 16.3848
};
\addplot [black]
table {%
22.875 9.508887
23.125 9.508887
};
\addplot [black]
table {%
22.875 16.3848
23.125 16.3848
};
\addplot [black]
table {%
24 212.215695
24 205.753981
};
\addplot [black]
table {%
24 217.1769125
24 224.556227
};
\addplot [black]
table {%
23.875 205.753981
24.125 205.753981
};
\addplot [black]
table {%
23.875 224.556227
24.125 224.556227
};
\addplot [black]
table {%
25 41.14965625
25 38.736559
};
\addplot [black]
table {%
25 43.99273225
25 48.2146
};
\addplot [black]
table {%
24.875 38.736559
25.125 38.736559
};
\addplot [black]
table {%
24.875 48.2146
25.125 48.2146
};
\addplot [black]
table {%
26 25.749255
26 22.20059
};
\addplot [black]
table {%
26 28.495655
26 32.46593
};
\addplot [black]
table {%
25.875 22.20059
26.125 22.20059
};
\addplot [black]
table {%
25.875 32.46593
26.125 32.46593
};
\addplot [black]
table {%
27 14.99509
27 12.127642
};
\addplot [black]
table {%
27 17.812025
27 21.99142
};
\addplot [black]
table {%
26.875 12.127642
27.125 12.127642
};
\addplot [black]
table {%
26.875 21.99142
27.125 21.99142
};
\addplot [black]
table {%
28 8.46559875
28 7.104306
};
\addplot [black]
table {%
28 9.92243375
28 11.984214
};
\addplot [black]
table {%
27.875 7.104306
28.125 7.104306
};
\addplot [black]
table {%
27.875 11.984214
28.125 11.984214
};
\addplot [black]
table {%
29 210.434251
29 206.530442
};
\addplot [black]
table {%
29 213.649874
29 218.469689
};
\addplot [black]
table {%
28.875 206.530442
29.125 206.530442
};
\addplot [black]
table {%
28.875 218.469689
29.125 218.469689
};
\addplot [black]
table {%
30 51.96651
30 43.76533
};
\addplot [black]
table {%
30 57.438665
30 65.27064
};
\addplot [black]
table {%
29.875 43.76533
30.125 43.76533
};
\addplot [black]
table {%
29.875 65.27064
30.125 65.27064
};
\addplot [black]
table {%
31 19.09462
31 15.77822
};
\addplot [black]
table {%
31 21.74155
31 25.65828
};
\addplot [black]
table {%
30.875 15.77822
31.125 15.77822
};
\addplot [black]
table {%
30.875 25.65828
31.125 25.65828
};
\addplot [black]
table {%
32 125.970542430878
32 101.925134658813
};
\addplot [black]
table {%
32 158.323049545288
32 205.810308456421
};
\addplot [black]
table {%
31.875 101.925134658813
32.125 101.925134658813
};
\addplot [black]
table {%
31.875 205.810308456421
32.125 205.810308456421
};
\addplot [black]
table {%
33 31.8306684494019
33 27.5800228118896
};
\addplot [black]
table {%
33 34.742534160614
33 39.0715599060059
};
\addplot [black]
table {%
32.875 27.5800228118896
33.125 27.5800228118896
};
\addplot [black]
table {%
32.875 39.0715599060059
33.125 39.0715599060059
};
\path [draw=color1, fill=color1, semithick]
(axis cs:0.75,11.70028425)
--(axis cs:1.25,11.70028425)
--(axis cs:1.25,13.073789)
--(axis cs:0.75,13.073789)
--(axis cs:0.75,11.70028425)
--cycle;
\path [draw=color1, fill=color1, semithick]
(axis cs:1.75,183.89566725)
--(axis cs:2.25,183.89566725)
--(axis cs:2.25,191.88633275)
--(axis cs:1.75,191.88633275)
--(axis cs:1.75,183.89566725)
--cycle;
\path [draw=color1, fill=color1, semithick]
(axis cs:2.75,375.2055225)
--(axis cs:3.25,375.2055225)
--(axis cs:3.25,382.8742725)
--(axis cs:2.75,382.8742725)
--(axis cs:2.75,375.2055225)
--cycle;
\path [draw=color1, fill=color1, semithick]
(axis cs:3.75,211.94901175)
--(axis cs:4.25,211.94901175)
--(axis cs:4.25,214.92333475)
--(axis cs:3.75,214.92333475)
--(axis cs:3.75,211.94901175)
--cycle;
\path [draw=color1, fill=color1, semithick]
(axis cs:4.75,216.0403925)
--(axis cs:5.25,216.0403925)
--(axis cs:5.25,223.8317275)
--(axis cs:4.75,223.8317275)
--(axis cs:4.75,216.0403925)
--cycle;
\path [draw=color1, fill=color1, semithick]
(axis cs:5.75,197.15223)
--(axis cs:6.25,197.15223)
--(axis cs:6.25,202.8341525)
--(axis cs:5.75,202.8341525)
--(axis cs:5.75,197.15223)
--cycle;
\path [draw=color1, fill=color1, semithick]
(axis cs:6.75,1.3900025)
--(axis cs:7.25,1.3900025)
--(axis cs:7.25,1.653393)
--(axis cs:6.75,1.653393)
--(axis cs:6.75,1.3900025)
--cycle;
\path [draw=color1, fill=color1, semithick]
(axis cs:7.75,200.89312)
--(axis cs:8.25,200.89312)
--(axis cs:8.25,204.655735)
--(axis cs:7.75,204.655735)
--(axis cs:7.75,200.89312)
--cycle;
\path [draw=color1, fill=color1, semithick]
(axis cs:8.75,31.95269)
--(axis cs:9.25,31.95269)
--(axis cs:9.25,33.1604925)
--(axis cs:8.75,33.1604925)
--(axis cs:8.75,31.95269)
--cycle;
\path [draw=color1, fill=color1, semithick]
(axis cs:9.75,30.741925)
--(axis cs:10.25,30.741925)
--(axis cs:10.25,36.590175)
--(axis cs:9.75,36.590175)
--(axis cs:9.75,30.741925)
--cycle;
\path [draw=color1, fill=color1, semithick]
(axis cs:10.75,15.86009)
--(axis cs:11.25,15.86009)
--(axis cs:11.25,18.525365)
--(axis cs:10.75,18.525365)
--(axis cs:10.75,15.86009)
--cycle;
\path [draw=color1, fill=color1, semithick]
(axis cs:11.75,1.794938)
--(axis cs:12.25,1.794938)
--(axis cs:12.25,2.11381725)
--(axis cs:11.75,2.11381725)
--(axis cs:11.75,1.794938)
--cycle;
\path [draw=color1, fill=color1, semithick]
(axis cs:12.75,188.65909575)
--(axis cs:13.25,188.65909575)
--(axis cs:13.25,191.172125)
--(axis cs:12.75,191.172125)
--(axis cs:12.75,188.65909575)
--cycle;
\path [draw=color1, fill=color1, semithick]
(axis cs:13.75,26.088098)
--(axis cs:14.25,26.088098)
--(axis cs:14.25,32.2198855)
--(axis cs:13.75,32.2198855)
--(axis cs:13.75,26.088098)
--cycle;
\path [draw=color1, fill=color1, semithick]
(axis cs:14.75,5.030363)
--(axis cs:15.25,5.030363)
--(axis cs:15.25,6.041639)
--(axis cs:14.75,6.041639)
--(axis cs:14.75,5.030363)
--cycle;
\path [draw=color1, fill=color1, semithick]
(axis cs:15.75,3.88785775)
--(axis cs:16.25,3.88785775)
--(axis cs:16.25,4.709887)
--(axis cs:15.75,4.709887)
--(axis cs:15.75,3.88785775)
--cycle;
\path [draw=color1, fill=color1, semithick]
(axis cs:16.75,2.4651345)
--(axis cs:17.25,2.4651345)
--(axis cs:17.25,2.90320175)
--(axis cs:16.75,2.90320175)
--(axis cs:16.75,2.4651345)
--cycle;
\path [draw=color1, fill=color1, semithick]
(axis cs:17.75,2.01009225)
--(axis cs:18.25,2.01009225)
--(axis cs:18.25,2.30050975)
--(axis cs:17.75,2.30050975)
--(axis cs:17.75,2.01009225)
--cycle;
\path [draw=color1, fill=color1, semithick]
(axis cs:18.75,194.2868475)
--(axis cs:19.25,194.2868475)
--(axis cs:19.25,198.180745)
--(axis cs:18.75,198.180745)
--(axis cs:18.75,194.2868475)
--cycle;
\path [draw=color1, fill=color1, semithick]
(axis cs:19.75,30.21459525)
--(axis cs:20.25,30.21459525)
--(axis cs:20.25,31.30356)
--(axis cs:19.75,31.30356)
--(axis cs:19.75,30.21459525)
--cycle;
\path [draw=color1, fill=color1, semithick]
(axis cs:20.75,10.6162425)
--(axis cs:21.25,10.6162425)
--(axis cs:21.25,12.21573)
--(axis cs:20.75,12.21573)
--(axis cs:20.75,10.6162425)
--cycle;
\path [draw=color1, fill=color1, semithick]
(axis cs:21.75,8.235085)
--(axis cs:22.25,8.235085)
--(axis cs:22.25,9.9443425)
--(axis cs:21.75,9.9443425)
--(axis cs:21.75,8.235085)
--cycle;
\path [draw=color1, fill=color1, semithick]
(axis cs:22.75,11.432842)
--(axis cs:23.25,11.432842)
--(axis cs:23.25,13.43078225)
--(axis cs:22.75,13.43078225)
--(axis cs:22.75,11.432842)
--cycle;
\path [draw=color1, fill=color1, semithick]
(axis cs:23.75,212.215695)
--(axis cs:24.25,212.215695)
--(axis cs:24.25,217.1769125)
--(axis cs:23.75,217.1769125)
--(axis cs:23.75,212.215695)
--cycle;
\path [draw=color1, fill=color1, semithick]
(axis cs:24.75,41.14965625)
--(axis cs:25.25,41.14965625)
--(axis cs:25.25,43.99273225)
--(axis cs:24.75,43.99273225)
--(axis cs:24.75,41.14965625)
--cycle;
\path [draw=color1, fill=color1, semithick]
(axis cs:25.75,25.749255)
--(axis cs:26.25,25.749255)
--(axis cs:26.25,28.495655)
--(axis cs:25.75,28.495655)
--(axis cs:25.75,25.749255)
--cycle;
\path [draw=color1, fill=color1, semithick]
(axis cs:26.75,14.99509)
--(axis cs:27.25,14.99509)
--(axis cs:27.25,17.812025)
--(axis cs:26.75,17.812025)
--(axis cs:26.75,14.99509)
--cycle;
\path [draw=color1, fill=color1, semithick]
(axis cs:27.75,8.46559875)
--(axis cs:28.25,8.46559875)
--(axis cs:28.25,9.92243375)
--(axis cs:27.75,9.92243375)
--(axis cs:27.75,8.46559875)
--cycle;
\path [draw=color1, fill=color1, semithick]
(axis cs:28.75,210.434251)
--(axis cs:29.25,210.434251)
--(axis cs:29.25,213.649874)
--(axis cs:28.75,213.649874)
--(axis cs:28.75,210.434251)
--cycle;
\path [draw=color1, fill=color1, semithick]
(axis cs:29.75,51.96651)
--(axis cs:30.25,51.96651)
--(axis cs:30.25,57.438665)
--(axis cs:29.75,57.438665)
--(axis cs:29.75,51.96651)
--cycle;
\path [draw=color1, fill=color1, semithick]
(axis cs:30.75,19.09462)
--(axis cs:31.25,19.09462)
--(axis cs:31.25,21.74155)
--(axis cs:30.75,21.74155)
--(axis cs:30.75,19.09462)
--cycle;
\path [draw=color1, fill=color1, semithick]
(axis cs:31.75,125.970542430878)
--(axis cs:32.25,125.970542430878)
--(axis cs:32.25,158.323049545288)
--(axis cs:31.75,158.323049545288)
--(axis cs:31.75,125.970542430878)
--cycle;
\path [draw=color1, fill=color1, semithick]
(axis cs:32.75,31.8306684494019)
--(axis cs:33.25,31.8306684494019)
--(axis cs:33.25,34.742534160614)
--(axis cs:32.75,34.742534160614)
--(axis cs:32.75,31.8306684494019)
--cycle;
\addplot [line width=1pt, white]
table {%
0.75 12.332437
1.25 12.332437
};
\addplot [line width=1pt, white]
table {%
1.75 189.7202455
2.25 189.7202455
};
\addplot [line width=1pt, white]
table {%
2.75 379.29683
3.25 379.29683
};
\addplot [line width=1pt, white]
table {%
3.75 213.3597915
4.25 213.3597915
};
\addplot [line width=1pt, white]
table {%
4.75 219.768745
5.25 219.768745
};
\addplot [line width=1pt, white]
table {%
5.75 199.833825
6.25 199.833825
};
\addplot [line width=1pt, white]
table {%
6.75 1.51902
7.25 1.51902
};
\addplot [line width=1pt, white]
table {%
7.75 202.58638
8.25 202.58638
};
\addplot [line width=1pt, white]
table {%
8.75 32.564405
9.25 32.564405
};
\addplot [line width=1pt, white]
table {%
9.75 33.72365
10.25 33.72365
};
\addplot [line width=1pt, white]
table {%
10.75 16.95556
11.25 16.95556
};
\addplot [line width=1pt, white]
table {%
11.75 1.936866
12.25 1.936866
};
\addplot [line width=1pt, white]
table {%
12.75 189.8176355
13.25 189.8176355
};
\addplot [line width=1pt, white]
table {%
13.75 30.3864415
14.25 30.3864415
};
\addplot [line width=1pt, white]
table {%
14.75 5.390781
15.25 5.390781
};
\addplot [line width=1pt, white]
table {%
15.75 4.183715
16.25 4.183715
};
\addplot [line width=1pt, white]
table {%
16.75 2.6619025
17.25 2.6619025
};
\addplot [line width=1pt, white]
table {%
17.75 2.154668
18.25 2.154668
};
\addplot [line width=1pt, white]
table {%
18.75 195.590395
19.25 195.590395
};
\addplot [line width=1pt, white]
table {%
19.75 30.721
20.25 30.721
};
\addplot [line width=1pt, white]
table {%
20.75 11.439775
21.25 11.439775
};
\addplot [line width=1pt, white]
table {%
21.75 8.921495
22.25 8.921495
};
\addplot [line width=1pt, white]
table {%
22.75 12.2912895
23.25 12.2912895
};
\addplot [line width=1pt, white]
table {%
23.75 214.3597155
24.25 214.3597155
};
\addplot [line width=1pt, white]
table {%
24.75 42.343524
25.25 42.343524
};
\addplot [line width=1pt, white]
table {%
25.75 26.947185
26.25 26.947185
};
\addplot [line width=1pt, white]
table {%
26.75 16.154165
27.25 16.154165
};
\addplot [line width=1pt, white]
table {%
27.75 9.072689
28.25 9.072689
};
\addplot [line width=1pt, white]
table {%
28.75 211.6475915
29.25 211.6475915
};
\addplot [line width=1pt, white]
table {%
29.75 54.84649
30.25 54.84649
};
\addplot [line width=1pt, white]
table {%
30.75 20.247235
31.25 20.247235
};
\addplot [line width=1pt, white]
table {%
31.75 145.36190032959
32.25 145.36190032959
};
\addplot [line width=1pt, white]
table {%
32.75 33.1500768661499
33.25 33.1500768661499
};
\end{axis}

\end{tikzpicture}

%% file: tex/2_1_Robustness_rotated.tex
\begin{tikzpicture}

\definecolor{darkgoldenrod}{RGB}{184,134,11}
\definecolor{darkgray176}{RGB}{176,176,176}
\definecolor{firebrick}{RGB}{178,34,34}
\definecolor{steelblue31119180}{RGB}{31,119,180}
\definecolor{color0}{rgb}{0.90, 0.62, 0.00}
\definecolor{color1}{rgb}{0.34, 0.70, 0.91}
\definecolor{color2}{rgb}{0.00, 0.62, 0.45}
\definecolor{color3}{rgb}{0.94, 0.89, 0.27}
\definecolor{color4}{rgb}{0.00, 0.45, 0.69}
\definecolor{color5}{rgb}{0.83, 0.37, 0.00}

\begin{axis}[
width=\figurewidth,
height=\figureheight,
axis line style={white},
tick align=outside,
tick pos=left,
x grid style={darkgray176},
xmin=0.5, xmax=10.5,
xtick style={color=black},
xtick={1,2,3,4,5,6,7,8,9,10},
xticklabel style={rotate=45.0,anchor=east},
xticklabels={AKAZE,BRISK,FAST,HARRIS,ORB,SHITOMASI,SIFT,SURF,Superpoint,Superpoint\_cuda},
y grid style={darkgray176},
ymin=0.146875, ymax=1.040625,
ytick style={color=black},
ymajorgrids,
ymajorticks=true,
minor y tick num = 2,
minor y grid style={dashed},
yminorgrids,
]
\addplot [black]
table {%
1 0.9375
1 0.878788
};
\addplot [black]
table {%
1 0.976744
1 1
};
\addplot [black]
table {%
0.875 0.878788
1.125 0.878788
};
\addplot [black]
table {%
0.875 1
1.125 1
};
\addplot [black]
table {%
2 0.8881195
2 0.854314
};
\addplot [black]
table {%
2 0.91075525
2 0.9424
};
\addplot [black]
table {%
1.875 0.854314
2.125 0.854314
};
\addplot [black]
table {%
1.875 0.9424
2.125 0.9424
};
\addplot [black]
table {%
3 0.82684475
3 0.794808
};
\addplot [black]
table {%
3 0.84900925
3 0.881499
};
\addplot [black]
table {%
2.875 0.794808
3.125 0.794808
};
\addplot [black]
table {%
2.875 0.881499
3.125 0.881499
};
\addplot [black]
table {%
4 0.86015275
4 0.788018
};
\addplot [black]
table {%
4 0.908297
4 0.961722
};
\addplot [black]
table {%
3.875 0.788018
4.125 0.788018
};
\addplot [black]
table {%
3.875 0.961722
4.125 0.961722
};
\addplot [black]
table {%
5 0.3519225
5 0.1875
};
\addplot [black]
table {%
5 0.461538
5 0.625
};
\addplot [black]
table {%
4.875 0.1875
5.125 0.1875
};
\addplot [black]
table {%
4.875 0.625
5.125 0.625
};
\addplot [black]
table {%
6 0.86457825
6 0.818021
};
\addplot [black]
table {%
6 0.89608125
6 0.936214
};
\addplot [black]
table {%
5.875 0.818021
6.125 0.818021
};
\addplot [black]
table {%
5.875 0.936214
6.125 0.936214
};
\addplot [black]
table {%
7 0.78485725
7 0.741294
};
\addplot [black]
table {%
7 0.81641025
7 0.862162
};
\addplot [black]
table {%
6.875 0.741294
7.125 0.741294
};
\addplot [black]
table {%
6.875 0.862162
7.125 0.862162
};
\addplot [black]
table {%
8 0.6563535
8 0.607219
};
\addplot [black]
table {%
8 0.689355
8 0.735149
};
\addplot [black]
table {%
7.875 0.607219
8.125 0.607219
};
\addplot [black]
table {%
7.875 0.735149
8.125 0.735149
};
\addplot [black]
table {%
9 0.779598703888335
9 0.720238095238095
};
\addplot [black]
table {%
9 0.819875776397516
9 0.879194630872483
};
\addplot [black]
table {%
8.875 0.720238095238095
9.125 0.720238095238095
};
\addplot [black]
table {%
8.875 0.879194630872483
9.125 0.879194630872483
};
\addplot [black]
table {%
10 0.779598703888335
10 0.720238095238095
};
\addplot [black]
table {%
10 0.819875776397516
10 0.879194630872483
};
\addplot [black]
table {%
9.875 0.720238095238095
10.125 0.720238095238095
};
\addplot [black]
table {%
9.875 0.879194630872483
10.125 0.879194630872483
};
\path [draw=color1, fill=color1, semithick]
(axis cs:0.75,0.9375)
--(axis cs:1.25,0.9375)
--(axis cs:1.25,0.976744)
--(axis cs:0.75,0.976744)
--(axis cs:0.75,0.9375)
--cycle;
\path [draw=color1, fill=color1, semithick]
(axis cs:1.75,0.8881195)
--(axis cs:2.25,0.8881195)
--(axis cs:2.25,0.91075525)
--(axis cs:1.75,0.91075525)
--(axis cs:1.75,0.8881195)
--cycle;
\path [draw=color1, fill=color1, semithick]
(axis cs:2.75,0.82684475)
--(axis cs:3.25,0.82684475)
--(axis cs:3.25,0.84900925)
--(axis cs:2.75,0.84900925)
--(axis cs:2.75,0.82684475)
--cycle;
\path [draw=color1, fill=color1, semithick]
(axis cs:3.75,0.86015275)
--(axis cs:4.25,0.86015275)
--(axis cs:4.25,0.908297)
--(axis cs:3.75,0.908297)
--(axis cs:3.75,0.86015275)
--cycle;
\path [draw=color1, fill=color1, semithick]
(axis cs:4.75,0.3519225)
--(axis cs:5.25,0.3519225)
--(axis cs:5.25,0.461538)
--(axis cs:4.75,0.461538)
--(axis cs:4.75,0.3519225)
--cycle;
\path [draw=color1, fill=color1, semithick]
(axis cs:5.75,0.86457825)
--(axis cs:6.25,0.86457825)
--(axis cs:6.25,0.89608125)
--(axis cs:5.75,0.89608125)
--(axis cs:5.75,0.86457825)
--cycle;
\path [draw=color1, fill=color1, semithick]
(axis cs:6.75,0.78485725)
--(axis cs:7.25,0.78485725)
--(axis cs:7.25,0.81641025)
--(axis cs:6.75,0.81641025)
--(axis cs:6.75,0.78485725)
--cycle;
\path [draw=color1, fill=color1, semithick]
(axis cs:7.75,0.6563535)
--(axis cs:8.25,0.6563535)
--(axis cs:8.25,0.689355)
--(axis cs:7.75,0.689355)
--(axis cs:7.75,0.6563535)
--cycle;
\path [draw=color1, fill=color1, semithick]
(axis cs:8.75,0.779598703888335)
--(axis cs:9.25,0.779598703888335)
--(axis cs:9.25,0.819875776397516)
--(axis cs:8.75,0.819875776397516)
--(axis cs:8.75,0.779598703888335)
--cycle;
\path [draw=color1, fill=color1, semithick]
(axis cs:9.75,0.779598703888335)
--(axis cs:10.25,0.779598703888335)
--(axis cs:10.25,0.819875776397516)
--(axis cs:9.75,0.819875776397516)
--(axis cs:9.75,0.779598703888335)
--cycle;
\addplot [line width=1pt, white]
table {%
0.75 0.964286
1.25 0.964286
};
\addplot [line width=1pt, white]
table {%
1.75 0.9
2.25 0.9
};
\addplot [line width=1pt, white]
table {%
2.75 0.837587
3.25 0.837587
};
\addplot [line width=1pt, white]
table {%
3.75 0.8865975
4.25 0.8865975
};
\addplot [line width=1pt, white]
table {%
4.75 0.403226
5.25 0.403226
};
\addplot [line width=1pt, white]
table {%
5.75 0.8830745
6.25 0.8830745
};
\addplot [line width=1pt, white]
table {%
6.75 0.8009655
7.25 0.8009655
};
\addplot [line width=1pt, white]
table {%
7.75 0.6730565
8.25 0.6730565
};
\addplot [line width=1pt, white]
table {%
8.75 0.798699928242793
9.25 0.798699928242793
};
\addplot [line width=1pt, white]
table {%
9.75 0.798699928242793
10.25 0.798699928242793
};
\end{axis}

\end{tikzpicture}

%% file: tex/2_2_Robustness_scaling.tex
\begin{tikzpicture}

\definecolor{darkgoldenrod}{RGB}{184,134,11}
\definecolor{darkgray176}{RGB}{176,176,176}
\definecolor{firebrick}{RGB}{178,34,34}
\definecolor{steelblue31119180}{RGB}{31,119,180}
\definecolor{color0}{rgb}{0.90, 0.62, 0.00}
\definecolor{color1}{rgb}{0.34, 0.70, 0.91}
\definecolor{color2}{rgb}{0.00, 0.62, 0.45}
\definecolor{color3}{rgb}{0.94, 0.89, 0.27}
\definecolor{color4}{rgb}{0.00, 0.45, 0.69}
\definecolor{color5}{rgb}{0.83, 0.37, 0.00}

\begin{axis}[
width=\figurewidth,
height=\figureheight,
axis line style={white},
tick align=outside,
tick pos=left,
x grid style={darkgray176},
xmin=0.5, xmax=10.5,
xtick style={color=black},
xtick={1,2,3,4,5,6,7,8,9,10},
xticklabel style={rotate=45.0,anchor=east},
xticklabels={AKAZE,BRISK,FAST,HARRIS,ORB,SHITOMASI,SIFT,SURF,Superpoint,Superpoint\_cuda},
y grid style={darkgray176},
ymin=-0.0433734939759036, ymax=0.910843373493976,
ytick style={color=black},
ymajorgrids,
ymajorticks=true,
minor y tick num = 2,
minor y grid style={dashed},
yminorgrids,
]
\addplot [black]
table {%
1 0.0196078
1 0
};
\addplot [black]
table {%
1 0.0487805
1 0.0909091
};
\addplot [black]
table {%
0.875 0
1.125 0
};
\addplot [black]
table {%
0.875 0.0909091
1.125 0.0909091
};
\addplot [black]
table {%
2 0.066150775
2 0.0388693
};
\addplot [black]
table {%
2 0.10075775
2 0.152542
};
\addplot [black]
table {%
1.875 0.0388693
2.125 0.0388693
};
\addplot [black]
table {%
1.875 0.152542
2.125 0.152542
};
\addplot [black]
table {%
3 0.049640425
3 0.0327198
};
\addplot [black]
table {%
3 0.08051775
3 0.122857
};
\addplot [black]
table {%
2.875 0.0327198
3.125 0.0327198
};
\addplot [black]
table {%
2.875 0.122857
3.125 0.122857
};
\addplot [black]
table {%
4 0.0414747
4 0.0148515
};
\addplot [black]
table {%
4 0.0678733
4 0.106061
};
\addplot [black]
table {%
3.875 0.0148515
4.125 0.0148515
};
\addplot [black]
table {%
3.875 0.106061
4.125 0.106061
};
\addplot [black]
table {%
5 0
5 0
};
\addplot [black]
table {%
5 0
5 0
};
\addplot [black]
table {%
4.875 0
5.125 0
};
\addplot [black]
table {%
4.875 0
5.125 0
};
\addplot [black]
table {%
6 0.15212875
6 0.102564
};
\addplot [black]
table {%
6 0.1861385
6 0.235993
};
\addplot [black]
table {%
5.875 0.102564
6.125 0.102564
};
\addplot [black]
table {%
5.875 0.235993
6.125 0.235993
};
\addplot [black]
table {%
7 0.103746
7 0.0512821
};
\addplot [black]
table {%
7 0.14213825
7 0.198511
};
\addplot [black]
table {%
6.875 0.0512821
7.125 0.0512821
};
\addplot [black]
table {%
6.875 0.198511
7.125 0.198511
};
\addplot [black]
table {%
8 0.15865325
8 0.112676
};
\addplot [black]
table {%
8 0.18981525
8 0.234862
};
\addplot [black]
table {%
7.875 0.112676
8.125 0.112676
};
\addplot [black]
table {%
7.875 0.234862
8.125 0.234862
};
\addplot [black]
table {%
9 0.757862046420399
9 0.69281045751634
};
\addplot [black]
table {%
9 0.802293717649447
9 0.867469879518072
};
\addplot [black]
table {%
8.875 0.69281045751634
9.125 0.69281045751634
};
\addplot [black]
table {%
8.875 0.867469879518072
9.125 0.867469879518072
};
\addplot [black]
table {%
10 0.757862046420399
10 0.69281045751634
};
\addplot [black]
table {%
10 0.802293717649447
10 0.867469879518072
};
\addplot [black]
table {%
9.875 0.69281045751634
10.125 0.69281045751634
};
\addplot [black]
table {%
9.875 0.867469879518072
10.125 0.867469879518072
};
\path [draw=color1, fill=color1, semithick]
(axis cs:0.75,0.0196078)
--(axis cs:1.25,0.0196078)
--(axis cs:1.25,0.0487805)
--(axis cs:0.75,0.0487805)
--(axis cs:0.75,0.0196078)
--cycle;
\path [draw=color1, fill=color1, semithick]
(axis cs:1.75,0.066150775)
--(axis cs:2.25,0.066150775)
--(axis cs:2.25,0.10075775)
--(axis cs:1.75,0.10075775)
--(axis cs:1.75,0.066150775)
--cycle;
\path [draw=color1, fill=color1, semithick]
(axis cs:2.75,0.049640425)
--(axis cs:3.25,0.049640425)
--(axis cs:3.25,0.08051775)
--(axis cs:2.75,0.08051775)
--(axis cs:2.75,0.049640425)
--cycle;
\path [draw=color1, fill=color1, semithick]
(axis cs:3.75,0.0414747)
--(axis cs:4.25,0.0414747)
--(axis cs:4.25,0.0678733)
--(axis cs:3.75,0.0678733)
--(axis cs:3.75,0.0414747)
--cycle;
\path [draw=color1, fill=color1, semithick]
(axis cs:4.75,0)
--(axis cs:5.25,0)
--(axis cs:5.25,0)
--(axis cs:4.75,0)
--(axis cs:4.75,0)
--cycle;
\path [draw=color1, fill=color1, semithick]
(axis cs:5.75,0.15212875)
--(axis cs:6.25,0.15212875)
--(axis cs:6.25,0.1861385)
--(axis cs:5.75,0.1861385)
--(axis cs:5.75,0.15212875)
--cycle;
\path [draw=color1, fill=color1, semithick]
(axis cs:6.75,0.103746)
--(axis cs:7.25,0.103746)
--(axis cs:7.25,0.14213825)
--(axis cs:6.75,0.14213825)
--(axis cs:6.75,0.103746)
--cycle;
\path [draw=color1, fill=color1, semithick]
(axis cs:7.75,0.15865325)
--(axis cs:8.25,0.15865325)
--(axis cs:8.25,0.18981525)
--(axis cs:7.75,0.18981525)
--(axis cs:7.75,0.15865325)
--cycle;
\path [draw=color1, fill=color1, semithick]
(axis cs:8.75,0.757862046420399)
--(axis cs:9.25,0.757862046420399)
--(axis cs:9.25,0.802293717649447)
--(axis cs:8.75,0.802293717649447)
--(axis cs:8.75,0.757862046420399)
--cycle;
\path [draw=color1, fill=color1, semithick]
(axis cs:9.75,0.757862046420399)
--(axis cs:10.25,0.757862046420399)
--(axis cs:10.25,0.802293717649447)
--(axis cs:9.75,0.802293717649447)
--(axis cs:9.75,0.757862046420399)
--cycle;
\addplot [line width=1pt, white]
table {%
0.75 0.030303
1.25 0.030303
};
\addplot [line width=1pt, white]
table {%
1.75 0.08112895
2.25 0.08112895
};
\addplot [line width=1pt, white]
table {%
2.75 0.06323015
3.25 0.06323015
};
\addplot [line width=1pt, white]
table {%
3.75 0.0526316
4.25 0.0526316
};
\addplot [line width=1pt, white]
table {%
4.75 0
5.25 0
};
\addplot [line width=1pt, white]
table {%
5.75 0.1641525
6.25 0.1641525
};
\addplot [line width=1pt, white]
table {%
6.75 0.123209
7.25 0.123209
};
\addplot [line width=1pt, white]
table {%
7.75 0.17316
8.25 0.17316
};
\addplot [line width=1pt, white]
table {%
8.75 0.782122905027933
9.25 0.782122905027933
};
\addplot [line width=1pt, white]
table {%
9.75 0.782122905027933
10.25 0.782122905027933
};
\end{axis}

\end{tikzpicture}

%% file: tex/2_3_Robustness_blurred.tex
\begin{tikzpicture}

\definecolor{darkgoldenrod}{RGB}{184,134,11}
\definecolor{darkgray176}{RGB}{176,176,176}
\definecolor{firebrick}{RGB}{178,34,34}
\definecolor{steelblue31119180}{RGB}{31,119,180}
\definecolor{color0}{rgb}{0.90, 0.62, 0.00}
\definecolor{color1}{rgb}{0.34, 0.70, 0.91}
\definecolor{color2}{rgb}{0.00, 0.62, 0.45}
\definecolor{color3}{rgb}{0.94, 0.89, 0.27}
\definecolor{color4}{rgb}{0.00, 0.45, 0.69}
\definecolor{color5}{rgb}{0.83, 0.37, 0.00}

\begin{axis}[
width=\figurewidth,
height=\figureheight,
axis line style={white},
tick align=outside,
tick pos=left,
x grid style={darkgray176},
xmin=0.5, xmax=10.5,
xtick style={color=black},
xtick={1,2,3,4,5,6,7,8,9,10},
xticklabel style={rotate=45.0,anchor=east},
xticklabels={AKAZE,BRISK,FAST,HARRIS,ORB,SHITOMASI,SIFT,SURF,Superpoint,Superpoint\_cuda},
y grid style={darkgray176},
ymin=0.11132465, ymax=0.92143835,
ytick style={color=black},
ymajorgrids,
ymajorticks=true,
minor y tick num = 2,
minor y grid style={dashed},
yminorgrids,
]
\addplot [black]
table {%
1 0.675676
1 0.551724
};
\addplot [black]
table {%
1 0.761905
1 0.884615
};
\addplot [black]
table {%
0.875 0.551724
1.125 0.551724
};
\addplot [black]
table {%
0.875 0.884615
1.125 0.884615
};
\addplot [black]
table {%
2 0.28745275
2 0.226446
};
\addplot [black]
table {%
2 0.3328165
2 0.393136
};
\addplot [black]
table {%
1.875 0.226446
2.125 0.226446
};
\addplot [black]
table {%
1.875 0.393136
2.125 0.393136
};
\addplot [black]
table {%
3 0.34492175
3 0.285592
};
\addplot [black]
table {%
3 0.38551075
3 0.438007
};
\addplot [black]
table {%
2.875 0.285592
3.125 0.285592
};
\addplot [black]
table {%
2.875 0.438007
3.125 0.438007
};
\addplot [black]
table {%
4 0.68397175
4 0.603604
};
\addplot [black]
table {%
4 0.739583
4 0.822727
};
\addplot [black]
table {%
3.875 0.603604
4.125 0.603604
};
\addplot [black]
table {%
3.875 0.822727
4.125 0.822727
};
\addplot [black]
table {%
5 0.28390925
5 0.148148
};
\addplot [black]
table {%
5 0.382353
5 0.513514
};
\addplot [black]
table {%
4.875 0.148148
5.125 0.148148
};
\addplot [black]
table {%
4.875 0.513514
5.125 0.513514
};
\addplot [black]
table {%
6 0.68643325
6 0.612688
};
\addplot [black]
table {%
6 0.73642275
6 0.80754
};
\addplot [black]
table {%
5.875 0.612688
6.125 0.612688
};
\addplot [black]
table {%
5.875 0.80754
6.125 0.80754
};
\addplot [black]
table {%
7 0.584699
7 0.519886
};
\addplot [black]
table {%
7 0.62972675
7 0.694915
};
\addplot [black]
table {%
6.875 0.519886
7.125 0.519886
};
\addplot [black]
table {%
6.875 0.694915
7.125 0.694915
};
\addplot [black]
table {%
8 0.63908375
8 0.584711
};
\addplot [black]
table {%
8 0.67648775
8 0.732
};
\addplot [black]
table {%
7.875 0.584711
8.125 0.584711
};
\addplot [black]
table {%
7.875 0.732
8.125 0.732
};
\addplot [black]
table {%
9 0.614933604626258
9 0.54421768707483
};
\addplot [black]
table {%
9 0.662790697674419
9 0.733766233766234
};
\addplot [black]
table {%
8.875 0.54421768707483
9.125 0.54421768707483
};
\addplot [black]
table {%
8.875 0.733766233766234
9.125 0.733766233766234
};
\addplot [black]
table {%
10 0.614933604626258
10 0.54421768707483
};
\addplot [black]
table {%
10 0.662790697674419
10 0.733766233766234
};
\addplot [black]
table {%
9.875 0.54421768707483
10.125 0.54421768707483
};
\addplot [black]
table {%
9.875 0.733766233766234
10.125 0.733766233766234
};
\path [draw=color1, fill=color1, semithick]
(axis cs:0.75,0.675676)
--(axis cs:1.25,0.675676)
--(axis cs:1.25,0.761905)
--(axis cs:0.75,0.761905)
--(axis cs:0.75,0.675676)
--cycle;
\path [draw=color1, fill=color1, semithick]
(axis cs:1.75,0.28745275)
--(axis cs:2.25,0.28745275)
--(axis cs:2.25,0.3328165)
--(axis cs:1.75,0.3328165)
--(axis cs:1.75,0.28745275)
--cycle;
\path [draw=color1, fill=color1, semithick]
(axis cs:2.75,0.34492175)
--(axis cs:3.25,0.34492175)
--(axis cs:3.25,0.38551075)
--(axis cs:2.75,0.38551075)
--(axis cs:2.75,0.34492175)
--cycle;
\path [draw=color1, fill=color1, semithick]
(axis cs:3.75,0.68397175)
--(axis cs:4.25,0.68397175)
--(axis cs:4.25,0.739583)
--(axis cs:3.75,0.739583)
--(axis cs:3.75,0.68397175)
--cycle;
\path [draw=color1, fill=color1, semithick]
(axis cs:4.75,0.28390925)
--(axis cs:5.25,0.28390925)
--(axis cs:5.25,0.382353)
--(axis cs:4.75,0.382353)
--(axis cs:4.75,0.28390925)
--cycle;
\path [draw=color1, fill=color1, semithick]
(axis cs:5.75,0.68643325)
--(axis cs:6.25,0.68643325)
--(axis cs:6.25,0.73642275)
--(axis cs:5.75,0.73642275)
--(axis cs:5.75,0.68643325)
--cycle;
\path [draw=color1, fill=color1, semithick]
(axis cs:6.75,0.584699)
--(axis cs:7.25,0.584699)
--(axis cs:7.25,0.62972675)
--(axis cs:6.75,0.62972675)
--(axis cs:6.75,0.584699)
--cycle;
\path [draw=color1, fill=color1, semithick]
(axis cs:7.75,0.63908375)
--(axis cs:8.25,0.63908375)
--(axis cs:8.25,0.67648775)
--(axis cs:7.75,0.67648775)
--(axis cs:7.75,0.63908375)
--cycle;
\path [draw=color1, fill=color1, semithick]
(axis cs:8.75,0.614933604626258)
--(axis cs:9.25,0.614933604626258)
--(axis cs:9.25,0.662790697674419)
--(axis cs:8.75,0.662790697674419)
--(axis cs:8.75,0.614933604626258)
--cycle;
\path [draw=color1, fill=color1, semithick]
(axis cs:9.75,0.614933604626258)
--(axis cs:10.25,0.614933604626258)
--(axis cs:10.25,0.662790697674419)
--(axis cs:9.75,0.662790697674419)
--(axis cs:9.75,0.614933604626258)
--cycle;
\addplot [line width=1pt, white]
table {%
0.75 0.71875
1.25 0.71875
};
\addplot [line width=1pt, white]
table {%
1.75 0.311021
2.25 0.311021
};
\addplot [line width=1pt, white]
table {%
2.75 0.3638805
3.25 0.3638805
};
\addplot [line width=1pt, white]
table {%
3.75 0.715663
4.25 0.715663
};
\addplot [line width=1pt, white]
table {%
4.75 0.333333
5.25 0.333333
};
\addplot [line width=1pt, white]
table {%
5.75 0.7135235
6.25 0.7135235
};
\addplot [line width=1pt, white]
table {%
6.75 0.6070475
7.25 0.6070475
};
\addplot [line width=1pt, white]
table {%
7.75 0.657476
8.25 0.657476
};
\addplot [line width=1pt, white]
table {%
8.75 0.638426044603948
9.25 0.638426044603948
};
\addplot [line width=1pt, white]
table {%
9.75 0.638426044603948
10.25 0.638426044603948
};
\end{axis}

\end{tikzpicture}

%% file: tex/4_Match_Ratio.tex
\begin{tikzpicture}

\definecolor{darkgoldenrod}{RGB}{184,134,11}
\definecolor{darkgray176}{RGB}{176,176,176}
\definecolor{firebrick}{RGB}{178,34,34}
\definecolor{steelblue31119180}{RGB}{31,119,180}

\definecolor{color0}{rgb}{0.90, 0.62, 0.00}
\definecolor{color1}{rgb}{0.34, 0.70, 0.91}
\definecolor{color2}{rgb}{0.00, 0.62, 0.45}
\definecolor{color3}{rgb}{0.94, 0.89, 0.27}
\definecolor{color4}{rgb}{0.00, 0.45, 0.69}
\definecolor{color5}{rgb}{0.83, 0.37, 0.00}

\begin{axis}[
width=\figurewidth,
height=\figureheight,
axis line style={white},
tick align=outside,
tick pos=left,
x grid style={darkgray176},
xmin=0.5, xmax=33.5,
xtick style={color=black},
xtick={1,2,3,4,5,6,7,8,9,10,11,12,13,14,15,16,17,18,19,20,21,22,23,24,25,26,27,28,29,30,31,32,33},
xticklabel style={rotate=45.0,anchor=east},
xticklabels={
  AKAZE\_AKAZE,
  BRISK\_BRIEF,
  BRISK\_BRISK,
  BRISK\_FREAK,
  BRISK\_SIFT,
  BRISK\_SURF,
  FAST\_BRIEF,
  FAST\_BRISK,
  FAST\_FREAK,
  FAST\_SIFT,
  FAST\_SURF,
  HARRIS\_BRIEF,
  HARRIS\_BRISK,
  HARRIS\_FREAK,
  HARRIS\_SIFT,
  HARRIS\_SURF,
  ORB\_ORB,
  SHITOMASI\_BRIEF,
  SHITOMASI\_BRISK,
  SHITOMASI\_FREAK,
  SHITOMASI\_SIFT,
  SHITOMASI\_SURF,
  SIFT\_BRIEF,
  SIFT\_BRISK,
  SIFT\_FREAK,
  SIFT\_SIFT,
  SIFT\_SURF,
  SURF\_BRIEF,
  SURF\_BRISK,
  SURF\_SIFT,
  SURF\_SURF,
  Superpoint,
  Superpoint\_cuda
},
y grid style={darkgray176},
ymin=0.26914855, ymax=1.03480245,
ytick style={color=black},
ymajorgrids,
ymajorticks=true,
minor y tick num = 2,
minor y grid style={dashed},
yminorgrids,
]
\addplot [black]
table {%
1 0.789474
1 0.633333
};
\addplot [black]
table {%
1 0.894737
1 1
};
\addplot [black]
table {%
0.875 0.633333
1.125 0.633333
};
\addplot [black]
table {%
0.875 1
1.125 1
};
\addplot [black]
table {%
2 0.647329
2 0.545852
};
\addplot [black]
table {%
2 0.71690125
2 0.821256
};
\addplot [black]
table {%
1.875 0.545852
2.125 0.545852
};
\addplot [black]
table {%
1.875 0.821256
2.125 0.821256
};
\addplot [black]
table {%
3 0.59242625
3 0.506211
};
\addplot [black]
table {%
3 0.66207525
3 0.765957
};
\addplot [black]
table {%
2.875 0.506211
3.125 0.506211
};
\addplot [black]
table {%
2.875 0.765957
3.125 0.765957
};
\addplot [black]
table {%
4 0.5
4 0.347826
};
\addplot [black]
table {%
4 0.603774
4 0.759259
};
\addplot [black]
table {%
3.875 0.347826
4.125 0.347826
};
\addplot [black]
table {%
3.875 0.759259
4.125 0.759259
};
\addplot [black]
table {%
5 0.58088575
5 0.492586
};
\addplot [black]
table {%
5 0.63983925
5 0.727407
};
\addplot [black]
table {%
4.875 0.492586
5.125 0.492586
};
\addplot [black]
table {%
4.875 0.727407
5.125 0.727407
};
\addplot [black]
table {%
6 0.53068475
6 0.456597
};
\addplot [black]
table {%
6 0.586998
6 0.671111
};
\addplot [black]
table {%
5.875 0.456597
6.125 0.456597
};
\addplot [black]
table {%
5.875 0.671111
6.125 0.671111
};
\addplot [black]
table {%
7 0.723245
7 0.618182
};
\addplot [black]
table {%
7 0.79641925
7 0.894231
};
\addplot [black]
table {%
6.875 0.618182
7.125 0.618182
};
\addplot [black]
table {%
6.875 0.894231
7.125 0.894231
};
\addplot [black]
table {%
8 0.64180025
8 0.537748
};
\addplot [black]
table {%
8 0.71816175
8 0.825342
};
\addplot [black]
table {%
7.875 0.537748
8.125 0.537748
};
\addplot [black]
table {%
7.875 0.825342
8.125 0.825342
};
\addplot [black]
table {%
9 0.664502
9 0.559211
};
\addplot [black]
table {%
9 0.735906
9 0.842572
};
\addplot [black]
table {%
8.875 0.559211
9.125 0.559211
};
\addplot [black]
table {%
8.875 0.842572
9.125 0.842572
};
\addplot [black]
table {%
10 0.72112975
10 0.645198
};
\addplot [black]
table {%
10 0.772062
10 0.847992
};
\addplot [black]
table {%
9.875 0.645198
10.125 0.645198
};
\addplot [black]
table {%
9.875 0.847992
10.125 0.847992
};
\addplot [black]
table {%
11 0.53421975
11 0.446645
};
\addplot [black]
table {%
11 0.599813
11 0.698015
};
\addplot [black]
table {%
10.875 0.446645
11.125 0.446645
};
\addplot [black]
table {%
10.875 0.698015
11.125 0.698015
};
\addplot [black]
table {%
12 0.781818
12 0.642857
};
\addplot [black]
table {%
12 0.875
12 1
};
\addplot [black]
table {%
11.875 0.642857
12.125 0.642857
};
\addplot [black]
table {%
11.875 1
12.125 1
};
\addplot [black]
table {%
13 0.70693725
13 0.569307
};
\addplot [black]
table {%
13 0.801047
13 0.918129
};
\addplot [black]
table {%
12.875 0.569307
13.125 0.569307
};
\addplot [black]
table {%
12.875 0.918129
13.125 0.918129
};
\addplot [black]
table {%
14 0.715447
14 0.568807
};
\addplot [black]
table {%
14 0.81740825
14 0.942149
};
\addplot [black]
table {%
13.875 0.568807
14.125 0.568807
};
\addplot [black]
table {%
13.875 0.942149
14.125 0.942149
};
\addplot [black]
table {%
15 0.826286
15 0.736607
};
\addplot [black]
table {%
15 0.886149
15 0.956938
};
\addplot [black]
table {%
14.875 0.736607
15.125 0.736607
};
\addplot [black]
table {%
14.875 0.956938
15.125 0.956938
};
\addplot [black]
table {%
16 0.522727
16 0.408696
};
\addplot [black]
table {%
16 0.61754575
16 0.759336
};
\addplot [black]
table {%
15.875 0.408696
16.125 0.408696
};
\addplot [black]
table {%
15.875 0.759336
16.125 0.759336
};
\addplot [black]
table {%
17 0.676471
17 0.532258
};
\addplot [black]
table {%
17 0.775862
17 0.923077
};
\addplot [black]
table {%
16.875 0.532258
17.125 0.532258
};
\addplot [black]
table {%
16.875 0.923077
17.125 0.923077
};
\addplot [black]
table {%
18 0.728395
18 0.615385
};
\addplot [black]
table {%
18 0.804348
18 0.916667
};
\addplot [black]
table {%
17.875 0.615385
18.125 0.615385
};
\addplot [black]
table {%
17.875 0.916667
18.125 0.916667
};
\addplot [black]
table {%
19 0.5592495
19 0.448357
};
\addplot [black]
table {%
19 0.63613525
19 0.748908
};
\addplot [black]
table {%
18.875 0.448357
19.125 0.448357
};
\addplot [black]
table {%
18.875 0.748908
19.125 0.748908
};
\addplot [black]
table {%
20 0.641509
20 0.543568
};
\addplot [black]
table {%
20 0.719048
20 0.833333
};
\addplot [black]
table {%
19.875 0.543568
20.125 0.543568
};
\addplot [black]
table {%
19.875 0.833333
20.125 0.833333
};
\addplot [black]
table {%
21 0.74421675
21 0.669054
};
\addplot [black]
table {%
21 0.79619925
21 0.866551
};
\addplot [black]
table {%
20.875 0.669054
21.125 0.669054
};
\addplot [black]
table {%
20.875 0.866551
21.125 0.866551
};
\addplot [black]
table {%
22 0.436396
22 0.372449
};
\addplot [black]
table {%
22 0.48979225
22 0.569728
};
\addplot [black]
table {%
21.875 0.372449
22.125 0.372449
};
\addplot [black]
table {%
21.875 0.569728
22.125 0.569728
};
\addplot [black]
table {%
23 0.5420435
23 0.426471
};
\addplot [black]
table {%
23 0.62378275
23 0.746032
};
\addplot [black]
table {%
22.875 0.426471
23.125 0.426471
};
\addplot [black]
table {%
22.875 0.746032
23.125 0.746032
};
\addplot [black]
table {%
24 0.5
24 0.41048
};
\addplot [black]
table {%
24 0.57518475
24 0.682759
};
\addplot [black]
table {%
23.875 0.41048
24.125 0.41048
};
\addplot [black]
table {%
23.875 0.682759
24.125 0.682759
};
\addplot [black]
table {%
25 0.508646
25 0.392593
};
\addplot [black]
table {%
25 0.590278
25 0.707792
};
\addplot [black]
table {%
24.875 0.392593
25.125 0.392593
};
\addplot [black]
table {%
24.875 0.707792
25.125 0.707792
};
\addplot [black]
table {%
26 0.570225
26 0.481081
};
\addplot [black]
table {%
26 0.63713625
26 0.733496
};
\addplot [black]
table {%
25.875 0.481081
26.125 0.481081
};
\addplot [black]
table {%
25.875 0.733496
26.125 0.733496
};
\addplot [black]
table {%
27 0.3753805
27 0.303951
};
\addplot [black]
table {%
27 0.42671675
27 0.50289
};
\addplot [black]
table {%
26.875 0.303951
27.125 0.303951
};
\addplot [black]
table {%
26.875 0.50289
27.125 0.50289
};
\addplot [black]
table {%
28 0.63548975
28 0.5
};
\addplot [black]
table {%
28 0.737374
28 0.87931
};
\addplot [black]
table {%
27.875 0.5
28.125 0.5
};
\addplot [black]
table {%
27.875 0.87931
28.125 0.87931
};
\addplot [black]
table {%
29 0.57518475
29 0.418605
};
\addplot [black]
table {%
29 0.688964
29 0.859155
};
\addplot [black]
table {%
28.875 0.418605
29.125 0.418605
};
\addplot [black]
table {%
28.875 0.859155
29.125 0.859155
};
\addplot [black]
table {%
30 0.5941535
30 0.507901
};
\addplot [black]
table {%
30 0.6659665
30 0.773512
};
\addplot [black]
table {%
29.875 0.507901
30.125 0.507901
};
\addplot [black]
table {%
29.875 0.773512
30.125 0.773512
};
\addplot [black]
table {%
31 0.57667325
31 0.487472
};
\addplot [black]
table {%
31 0.6543235
31 0.76
};
\addplot [black]
table {%
30.875 0.487472
31.125 0.487472
};
\addplot [black]
table {%
30.875 0.76
31.125 0.76
};
\addplot [black]
table {%
32 0.832369942196532
32 0.764397905759162
};
\addplot [black]
table {%
32 0.881656804733728
32 0.954545454545455
};
\addplot [black]
table {%
31.875 0.764397905759162
32.125 0.764397905759162
};
\addplot [black]
table {%
31.875 0.954545454545455
32.125 0.954545454545455
};
\addplot [black]
table {%
33 0.832369942196532
33 0.764397905759162
};
\addplot [black]
table {%
33 0.881656804733728
33 0.954545454545455
};
\addplot [black]
table {%
32.875 0.764397905759162
33.125 0.764397905759162
};
\addplot [black]
table {%
32.875 0.954545454545455
33.125 0.954545454545455
};
\path [draw=color1, fill=color1, semithick]
(axis cs:0.75,0.789474)
--(axis cs:1.25,0.789474)
--(axis cs:1.25,0.894737)
--(axis cs:0.75,0.894737)
--(axis cs:0.75,0.789474)
--cycle;
\path [draw=color1, fill=color1, semithick]
(axis cs:1.75,0.647329)
--(axis cs:2.25,0.647329)
--(axis cs:2.25,0.71690125)
--(axis cs:1.75,0.71690125)
--(axis cs:1.75,0.647329)
--cycle;
\path [draw=color1, fill=color1, semithick]
(axis cs:2.75,0.59242625)
--(axis cs:3.25,0.59242625)
--(axis cs:3.25,0.66207525)
--(axis cs:2.75,0.66207525)
--(axis cs:2.75,0.59242625)
--cycle;
\path [draw=color1, fill=color1, semithick]
(axis cs:3.75,0.5)
--(axis cs:4.25,0.5)
--(axis cs:4.25,0.603774)
--(axis cs:3.75,0.603774)
--(axis cs:3.75,0.5)
--cycle;
\path [draw=color1, fill=color1, semithick]
(axis cs:4.75,0.58088575)
--(axis cs:5.25,0.58088575)
--(axis cs:5.25,0.63983925)
--(axis cs:4.75,0.63983925)
--(axis cs:4.75,0.58088575)
--cycle;
\path [draw=color1, fill=color1, semithick]
(axis cs:5.75,0.53068475)
--(axis cs:6.25,0.53068475)
--(axis cs:6.25,0.586998)
--(axis cs:5.75,0.586998)
--(axis cs:5.75,0.53068475)
--cycle;
\path [draw=color1, fill=color1, semithick]
(axis cs:6.75,0.723245)
--(axis cs:7.25,0.723245)
--(axis cs:7.25,0.79641925)
--(axis cs:6.75,0.79641925)
--(axis cs:6.75,0.723245)
--cycle;
\path [draw=color1, fill=color1, semithick]
(axis cs:7.75,0.64180025)
--(axis cs:8.25,0.64180025)
--(axis cs:8.25,0.71816175)
--(axis cs:7.75,0.71816175)
--(axis cs:7.75,0.64180025)
--cycle;
\path [draw=color1, fill=color1, semithick]
(axis cs:8.75,0.664502)
--(axis cs:9.25,0.664502)
--(axis cs:9.25,0.735906)
--(axis cs:8.75,0.735906)
--(axis cs:8.75,0.664502)
--cycle;
\path [draw=color1, fill=color1, semithick]
(axis cs:9.75,0.72112975)
--(axis cs:10.25,0.72112975)
--(axis cs:10.25,0.772062)
--(axis cs:9.75,0.772062)
--(axis cs:9.75,0.72112975)
--cycle;
\path [draw=color1, fill=color1, semithick]
(axis cs:10.75,0.53421975)
--(axis cs:11.25,0.53421975)
--(axis cs:11.25,0.599813)
--(axis cs:10.75,0.599813)
--(axis cs:10.75,0.53421975)
--cycle;
\path [draw=color1, fill=color1, semithick]
(axis cs:11.75,0.781818)
--(axis cs:12.25,0.781818)
--(axis cs:12.25,0.875)
--(axis cs:11.75,0.875)
--(axis cs:11.75,0.781818)
--cycle;
\path [draw=color1, fill=color1, semithick]
(axis cs:12.75,0.70693725)
--(axis cs:13.25,0.70693725)
--(axis cs:13.25,0.801047)
--(axis cs:12.75,0.801047)
--(axis cs:12.75,0.70693725)
--cycle;
\path [draw=color1, fill=color1, semithick]
(axis cs:13.75,0.715447)
--(axis cs:14.25,0.715447)
--(axis cs:14.25,0.81740825)
--(axis cs:13.75,0.81740825)
--(axis cs:13.75,0.715447)
--cycle;
\path [draw=color1, fill=color1, semithick]
(axis cs:14.75,0.826286)
--(axis cs:15.25,0.826286)
--(axis cs:15.25,0.886149)
--(axis cs:14.75,0.886149)
--(axis cs:14.75,0.826286)
--cycle;
\path [draw=color1, fill=color1, semithick]
(axis cs:15.75,0.522727)
--(axis cs:16.25,0.522727)
--(axis cs:16.25,0.61754575)
--(axis cs:15.75,0.61754575)
--(axis cs:15.75,0.522727)
--cycle;
\path [draw=color1, fill=color1, semithick]
(axis cs:16.75,0.676471)
--(axis cs:17.25,0.676471)
--(axis cs:17.25,0.775862)
--(axis cs:16.75,0.775862)
--(axis cs:16.75,0.676471)
--cycle;
\path [draw=color1, fill=color1, semithick]
(axis cs:17.75,0.728395)
--(axis cs:18.25,0.728395)
--(axis cs:18.25,0.804348)
--(axis cs:17.75,0.804348)
--(axis cs:17.75,0.728395)
--cycle;
\path [draw=color1, fill=color1, semithick]
(axis cs:18.75,0.5592495)
--(axis cs:19.25,0.5592495)
--(axis cs:19.25,0.63613525)
--(axis cs:18.75,0.63613525)
--(axis cs:18.75,0.5592495)
--cycle;
\path [draw=color1, fill=color1, semithick]
(axis cs:19.75,0.641509)
--(axis cs:20.25,0.641509)
--(axis cs:20.25,0.719048)
--(axis cs:19.75,0.719048)
--(axis cs:19.75,0.641509)
--cycle;
\path [draw=color1, fill=color1, semithick]
(axis cs:20.75,0.74421675)
--(axis cs:21.25,0.74421675)
--(axis cs:21.25,0.79619925)
--(axis cs:20.75,0.79619925)
--(axis cs:20.75,0.74421675)
--cycle;
\path [draw=color1, fill=color1, semithick]
(axis cs:21.75,0.436396)
--(axis cs:22.25,0.436396)
--(axis cs:22.25,0.48979225)
--(axis cs:21.75,0.48979225)
--(axis cs:21.75,0.436396)
--cycle;
\path [draw=color1, fill=color1, semithick]
(axis cs:22.75,0.5420435)
--(axis cs:23.25,0.5420435)
--(axis cs:23.25,0.62378275)
--(axis cs:22.75,0.62378275)
--(axis cs:22.75,0.5420435)
--cycle;
\path [draw=color1, fill=color1, semithick]
(axis cs:23.75,0.5)
--(axis cs:24.25,0.5)
--(axis cs:24.25,0.57518475)
--(axis cs:23.75,0.57518475)
--(axis cs:23.75,0.5)
--cycle;
\path [draw=color1, fill=color1, semithick]
(axis cs:24.75,0.508646)
--(axis cs:25.25,0.508646)
--(axis cs:25.25,0.590278)
--(axis cs:24.75,0.590278)
--(axis cs:24.75,0.508646)
--cycle;
\path [draw=color1, fill=color1, semithick]
(axis cs:25.75,0.570225)
--(axis cs:26.25,0.570225)
--(axis cs:26.25,0.63713625)
--(axis cs:25.75,0.63713625)
--(axis cs:25.75,0.570225)
--cycle;
\path [draw=color1, fill=color1, semithick]
(axis cs:26.75,0.3753805)
--(axis cs:27.25,0.3753805)
--(axis cs:27.25,0.42671675)
--(axis cs:26.75,0.42671675)
--(axis cs:26.75,0.3753805)
--cycle;
\path [draw=color1, fill=color1, semithick]
(axis cs:27.75,0.63548975)
--(axis cs:28.25,0.63548975)
--(axis cs:28.25,0.737374)
--(axis cs:27.75,0.737374)
--(axis cs:27.75,0.63548975)
--cycle;
\path [draw=color1, fill=color1, semithick]
(axis cs:28.75,0.57518475)
--(axis cs:29.25,0.57518475)
--(axis cs:29.25,0.688964)
--(axis cs:28.75,0.688964)
--(axis cs:28.75,0.57518475)
--cycle;
\path [draw=color1, fill=color1, semithick]
(axis cs:29.75,0.5941535)
--(axis cs:30.25,0.5941535)
--(axis cs:30.25,0.6659665)
--(axis cs:29.75,0.6659665)
--(axis cs:29.75,0.5941535)
--cycle;
\path [draw=color1, fill=color1, semithick]
(axis cs:30.75,0.57667325)
--(axis cs:31.25,0.57667325)
--(axis cs:31.25,0.6543235)
--(axis cs:30.75,0.6543235)
--(axis cs:30.75,0.57667325)
--cycle;
\path [draw=color1, fill=color1, semithick]
(axis cs:31.75,0.832369942196532)
--(axis cs:32.25,0.832369942196532)
--(axis cs:32.25,0.881656804733728)
--(axis cs:31.75,0.881656804733728)
--(axis cs:31.75,0.832369942196532)
--cycle;
\path [draw=color1, fill=color1, semithick]
(axis cs:32.75,0.832369942196532)
--(axis cs:33.25,0.832369942196532)
--(axis cs:33.25,0.881656804733728)
--(axis cs:32.75,0.881656804733728)
--(axis cs:32.75,0.832369942196532)
--cycle;
\addplot [line width=1pt, white]
table {%
0.75 0.846154
1.25 0.846154
};
\addplot [line width=1pt, white]
table {%
1.75 0.6787885
2.25 0.6787885
};
\addplot [line width=1pt, white]
table {%
2.75 0.6210685
3.25 0.6210685
};
\addplot [line width=1pt, white]
table {%
3.75 0.546875
4.25 0.546875
};
\addplot [line width=1pt, white]
table {%
4.75 0.6052415
5.25 0.6052415
};
\addplot [line width=1pt, white]
table {%
5.75 0.553285
6.25 0.553285
};
\addplot [line width=1pt, white]
table {%
6.75 0.7584575
7.25 0.7584575
};
\addplot [line width=1pt, white]
table {%
7.75 0.673174
8.25 0.673174
};
\addplot [line width=1pt, white]
table {%
8.75 0.6965275
9.25 0.6965275
};
\addplot [line width=1pt, white]
table {%
9.75 0.74253
10.25 0.74253
};
\addplot [line width=1pt, white]
table {%
10.75 0.557957
11.25 0.557957
};
\addplot [line width=1pt, white]
table {%
11.75 0.825
12.25 0.825
};
\addplot [line width=1pt, white]
table {%
12.75 0.747979
13.25 0.747979
};
\addplot [line width=1pt, white]
table {%
13.75 0.758929
14.25 0.758929
};
\addplot [line width=1pt, white]
table {%
14.75 0.8564455
15.25 0.8564455
};
\addplot [line width=1pt, white]
table {%
15.75 0.559406
16.25 0.559406
};
\addplot [line width=1pt, white]
table {%
16.75 0.722222
17.25 0.722222
};
\addplot [line width=1pt, white]
table {%
17.75 0.768357
18.25 0.768357
};
\addplot [line width=1pt, white]
table {%
18.75 0.591745
19.25 0.591745
};
\addplot [line width=1pt, white]
table {%
19.75 0.6771065
20.25 0.6771065
};
\addplot [line width=1pt, white]
table {%
20.75 0.7710125
21.25 0.7710125
};
\addplot [line width=1pt, white]
table {%
21.75 0.4558945
22.25 0.4558945
};
\addplot [line width=1pt, white]
table {%
22.75 0.583333
23.25 0.583333
};
\addplot [line width=1pt, white]
table {%
23.75 0.5275665
24.25 0.5275665
};
\addplot [line width=1pt, white]
table {%
24.75 0.546053
25.25 0.546053
};
\addplot [line width=1pt, white]
table {%
25.75 0.597963
26.25 0.597963
};
\addplot [line width=1pt, white]
table {%
26.75 0.395664
27.25 0.395664
};
\addplot [line width=1pt, white]
table {%
27.75 0.680412
28.25 0.680412
};
\addplot [line width=1pt, white]
table {%
28.75 0.623709
29.25 0.623709
};
\addplot [line width=1pt, white]
table {%
29.75 0.621613
30.25 0.621613
};
\addplot [line width=1pt, white]
table {%
30.75 0.607215
31.25 0.607215
};
\addplot [line width=1pt, white]
table {%
31.75 0.85792349726776
32.25 0.85792349726776
};
\addplot [line width=1pt, white]
table {%
32.75 0.85792349726776
33.25 0.85792349726776
};
\end{axis}

\end{tikzpicture}

%% file: tex/5_Match_Score.tex
\begin{tikzpicture}

\definecolor{darkgoldenrod}{RGB}{184,134,11}
\definecolor{darkgray176}{RGB}{176,176,176}
\definecolor{firebrick}{RGB}{178,34,34}
\definecolor{steelblue31119180}{RGB}{31,119,180}
\definecolor{color0}{rgb}{0.90, 0.62, 0.00}
\definecolor{color1}{rgb}{0.34, 0.70, 0.91}
\definecolor{color2}{rgb}{0.00, 0.62, 0.45}
\definecolor{color3}{rgb}{0.94, 0.89, 0.27}
\definecolor{color4}{rgb}{0.00, 0.45, 0.69}
\definecolor{color5}{rgb}{0.83, 0.37, 0.00}

\begin{axis}[
width=\figurewidth,
height=\figureheight,
axis line style={white},
tick align=outside,
tick pos=left,
x grid style={darkgray176},
xmin=0.5, xmax=33.5,
xtick style={color=black},
xtick={1,2,3,4,5,6,7,8,9,10,11,12,13,14,15,16,17,18,19,20,21,22,23,24,25,26,27,28,29,30,31,32,33},
xticklabel style={rotate=45.0,anchor=east},
xticklabels={
  AKAZE\_AKAZE,
  BRISK\_BRIEF,
  BRISK\_BRISK,
  BRISK\_FREAK,
  BRISK\_SIFT,
  BRISK\_SURF,
  FAST\_BRIEF,
  FAST\_BRISK,
  FAST\_FREAK,
  FAST\_SIFT,
  FAST\_SURF,
  HARRIS\_BRIEF,
  HARRIS\_BRISK,
  HARRIS\_FREAK,
  HARRIS\_SIFT,
  HARRIS\_SURF,
  ORB\_ORB,
  SHITOMASI\_BRIEF,
  SHITOMASI\_BRISK,
  SHITOMASI\_FREAK,
  SHITOMASI\_SIFT,
  SHITOMASI\_SURF,
  SIFT\_BRIEF,
  SIFT\_BRISK,
  SIFT\_FREAK,
  SIFT\_SIFT,
  SIFT\_SURF,
  SURF\_BRIEF,
  SURF\_BRISK,
  SURF\_SIFT,
  SURF\_SURF,
  Superpoint,
  Superpoint\_cuda
},
y grid style={darkgray176},
ymin=0.2227753, ymax=1.0370107,
ytick style={color=black},
ymajorgrids,
ymajorticks=true,
minor y tick num = 2,
minor y grid style={dashed},
yminorgrids,
]
\addplot [black]
table {%
1 0.638889
1 0.451613
};
\addplot [black]
table {%
1 0.852941
1 1
};
\addplot [black]
table {%
0.875 0.451613
1.125 0.451613
};
\addplot [black]
table {%
0.875 1
1.125 1
};
\addplot [black]
table {%
2 0.601247
2 0.40566
};
\addplot [black]
table {%
2 0.81173125
2 1
};
\addplot [black]
table {%
1.875 0.40566
2.125 0.40566
};
\addplot [black]
table {%
1.875 1
2.125 1
};
\addplot [black]
table {%
3 0.6480945
3 0.436823
};
\addplot [black]
table {%
3 0.846329
3 0.990975
};
\addplot [black]
table {%
2.875 0.436823
3.125 0.436823
};
\addplot [black]
table {%
2.875 0.990975
3.125 0.990975
};
\addplot [black]
table {%
4 0.53449625
4 0.342857
};
\addplot [black]
table {%
4 0.710526
4 0.972973
};
\addplot [black]
table {%
3.875 0.342857
4.125 0.342857
};
\addplot [black]
table {%
3.875 0.972973
4.125 0.972973
};
\addplot [black]
table {%
5 0.6034795
5 0.426829
};
\addplot [black]
table {%
5 0.79200075
5 0.978221
};
\addplot [black]
table {%
4.875 0.426829
5.125 0.426829
};
\addplot [black]
table {%
4.875 0.978221
5.125 0.978221
};
\addplot [black]
table {%
6 0.60641975
6 0.407666
};
\addplot [black]
table {%
6 0.79293575
6 0.973392
};
\addplot [black]
table {%
5.875 0.407666
6.125 0.407666
};
\addplot [black]
table {%
5.875 0.973392
6.125 0.973392
};
\addplot [black]
table {%
7 0.80575375
7 0.580882
};
\addplot [black]
table {%
7 0.97561
7 1
};
\addplot [black]
table {%
6.875 0.580882
7.125 0.580882
};
\addplot [black]
table {%
6.875 1
7.125 1
};
\addplot [black]
table {%
8 0.7600975
8 0.587234
};
\addplot [black]
table {%
8 0.94881925
8 0.997238
};
\addplot [black]
table {%
7.875 0.587234
8.125 0.587234
};
\addplot [black]
table {%
7.875 0.997238
8.125 0.997238
};
\addplot [black]
table {%
9 0.7876005
9 0.536481
};
\addplot [black]
table {%
9 0.959979
9 1
};
\addplot [black]
table {%
8.875 0.536481
9.125 0.536481
};
\addplot [black]
table {%
8.875 1
9.125 1
};
\addplot [black]
table {%
10 0.765463
10 0.518581
};
\addplot [black]
table {%
10 0.9557485
10 0.997596
};
\addplot [black]
table {%
9.875 0.518581
10.125 0.518581
};
\addplot [black]
table {%
9.875 0.997596
10.125 0.997596
};
\addplot [black]
table {%
11 0.67491025
11 0.450839
};
\addplot [black]
table {%
11 0.85851875
11 0.971649
};
\addplot [black]
table {%
10.875 0.450839
11.125 0.450839
};
\addplot [black]
table {%
10.875 0.971649
11.125 0.971649
};
\addplot [black]
table {%
12 0.772727
12 0.545455
};
\addplot [black]
table {%
12 0.944444
12 1
};
\addplot [black]
table {%
11.875 0.545455
12.125 0.545455
};
\addplot [black]
table {%
11.875 1
12.125 1
};
\addplot [black]
table {%
13 0.7893845
13 0.572581
};
\addplot [black]
table {%
13 0.96418875
13 1
};
\addplot [black]
table {%
12.875 0.572581
13.125 0.572581
};
\addplot [black]
table {%
12.875 1
13.125 1
};
\addplot [black]
table {%
14 0.792998
14 0.579545
};
\addplot [black]
table {%
14 0.96208475
14 1
};
\addplot [black]
table {%
13.875 0.579545
14.125 0.579545
};
\addplot [black]
table {%
13.875 1
14.125 1
};
\addplot [black]
table {%
15 0.7469605
15 0.510204
};
\addplot [black]
table {%
15 0.945537
15 1
};
\addplot [black]
table {%
14.875 0.510204
15.125 0.510204
};
\addplot [black]
table {%
14.875 1
15.125 1
};
\addplot [black]
table {%
16 0.626742
16 0.401869
};
\addplot [black]
table {%
16 0.8287925
16 0.978378
};
\addplot [black]
table {%
15.875 0.401869
16.125 0.401869
};
\addplot [black]
table {%
15.875 0.978378
16.125 0.978378
};
\addplot [black]
table {%
17 0.80927075
17 0.580645
};
\addplot [black]
table {%
17 0.963636
17 1
};
\addplot [black]
table {%
16.875 0.580645
17.125 0.580645
};
\addplot [black]
table {%
16.875 1
17.125 1
};
\addplot [black]
table {%
18 0.716418
18 0.438596
};
\addplot [black]
table {%
18 0.90411
18 1
};
\addplot [black]
table {%
17.875 0.438596
18.125 0.438596
};
\addplot [black]
table {%
17.875 1
18.125 1
};
\addplot [black]
table {%
19 0.705652
19 0.521505
};
\addplot [black]
table {%
19 0.885827
19 0.977444
};
\addplot [black]
table {%
18.875 0.521505
19.125 0.521505
};
\addplot [black]
table {%
18.875 0.977444
19.125 0.977444
};
\addplot [black]
table {%
20 0.72
20 0.507812
};
\addplot [black]
table {%
20 0.90795
20 0.99435
};
\addplot [black]
table {%
19.875 0.507812
20.125 0.507812
};
\addplot [black]
table {%
19.875 0.99435
20.125 0.99435
};
\addplot [black]
table {%
21 0.640216
21 0.45045
};
\addplot [black]
table {%
21 0.85746875
21 0.965092
};
\addplot [black]
table {%
20.875 0.45045
21.125 0.45045
};
\addplot [black]
table {%
20.875 0.965092
21.125 0.965092
};
\addplot [black]
table {%
22 0.45067625
22 0.259786
};
\addplot [black]
table {%
22 0.6280145
22 0.883582
};
\addplot [black]
table {%
21.875 0.259786
22.125 0.259786
};
\addplot [black]
table {%
21.875 0.883582
22.125 0.883582
};
\addplot [black]
table {%
23 0.6
23 0.405405
};
\addplot [black]
table {%
23 0.8
23 1
};
\addplot [black]
table {%
22.875 0.405405
23.125 0.405405
};
\addplot [black]
table {%
22.875 1
23.125 1
};
\addplot [black]
table {%
24 0.666667
24 0.439655
};
\addplot [black]
table {%
24 0.8335985
24 0.988235
};
\addplot [black]
table {%
23.875 0.439655
24.125 0.439655
};
\addplot [black]
table {%
23.875 0.988235
24.125 0.988235
};
\addplot [black]
table {%
25 0.65902425
25 0.446154
};
\addplot [black]
table {%
25 0.8438515
25 0.990909
};
\addplot [black]
table {%
24.875 0.446154
25.125 0.446154
};
\addplot [black]
table {%
24.875 0.990909
25.125 0.990909
};
\addplot [black]
table {%
26 0.6525175
26 0.446328
};
\addplot [black]
table {%
26 0.83734075
26 0.970149
};
\addplot [black]
table {%
25.875 0.446328
26.125 0.446328
};
\addplot [black]
table {%
25.875 0.970149
26.125 0.970149
};
\addplot [black]
table {%
27 0.46197525
27 0.266055
};
\addplot [black]
table {%
27 0.63891625
27 0.838384
};
\addplot [black]
table {%
26.875 0.266055
27.125 0.266055
};
\addplot [black]
table {%
26.875 0.838384
27.125 0.838384
};
\addplot [black]
table {%
28 0.615385
28 0.425532
};
\addplot [black]
table {%
28 0.83103175
28 1
};
\addplot [black]
table {%
27.875 0.425532
28.125 0.425532
};
\addplot [black]
table {%
27.875 1
28.125 1
};
\addplot [black]
table {%
29 0.64
29 0.419355
};
\addplot [black]
table {%
29 0.847458
29 1
};
\addplot [black]
table {%
28.875 0.419355
29.125 0.419355
};
\addplot [black]
table {%
28.875 1
29.125 1
};
\addplot [black]
table {%
30 0.5946985
30 0.412214
};
\addplot [black]
table {%
30 0.81447225
30 0.976501
};
\addplot [black]
table {%
29.875 0.412214
30.125 0.412214
};
\addplot [black]
table {%
29.875 0.976501
30.125 0.976501
};
\addplot [black]
table {%
31 0.642857
31 0.436441
};
\addplot [black]
table {%
31 0.85026725
31 0.979487
};
\addplot [black]
table {%
30.875 0.436441
31.125 0.436441
};
\addplot [black]
table {%
30.875 0.979487
31.125 0.979487
};
\addplot [black]
table {%
32 0.733629191321499
32 0.512820512820513
};
\addplot [black]
table {%
32 0.928250641573995
32 1
};
\addplot [black]
table {%
31.875 0.512820512820513
32.125 0.512820512820513
};
\addplot [black]
table {%
31.875 1
32.125 1
};
\addplot [black]
table {%
33 0.733219178082192
33 0.512820512820513
};
\addplot [black]
table {%
33 0.928250641573995
33 1
};
\addplot [black]
table {%
32.875 0.512820512820513
33.125 0.512820512820513
};
\addplot [black]
table {%
32.875 1
33.125 1
};
\path [draw=color1, fill=color1, semithick]
(axis cs:0.75,0.638889)
--(axis cs:1.25,0.638889)
--(axis cs:1.25,0.852941)
--(axis cs:0.75,0.852941)
--(axis cs:0.75,0.638889)
--cycle;
\path [draw=color1, fill=color1, semithick]
(axis cs:1.75,0.601247)
--(axis cs:2.25,0.601247)
--(axis cs:2.25,0.81173125)
--(axis cs:1.75,0.81173125)
--(axis cs:1.75,0.601247)
--cycle;
\path [draw=color1, fill=color1, semithick]
(axis cs:2.75,0.6480945)
--(axis cs:3.25,0.6480945)
--(axis cs:3.25,0.846329)
--(axis cs:2.75,0.846329)
--(axis cs:2.75,0.6480945)
--cycle;
\path [draw=color1, fill=color1, semithick]
(axis cs:3.75,0.53449625)
--(axis cs:4.25,0.53449625)
--(axis cs:4.25,0.710526)
--(axis cs:3.75,0.710526)
--(axis cs:3.75,0.53449625)
--cycle;
\path [draw=color1, fill=color1, semithick]
(axis cs:4.75,0.6034795)
--(axis cs:5.25,0.6034795)
--(axis cs:5.25,0.79200075)
--(axis cs:4.75,0.79200075)
--(axis cs:4.75,0.6034795)
--cycle;
\path [draw=color1, fill=color1, semithick]
(axis cs:5.75,0.60641975)
--(axis cs:6.25,0.60641975)
--(axis cs:6.25,0.79293575)
--(axis cs:5.75,0.79293575)
--(axis cs:5.75,0.60641975)
--cycle;
\path [draw=color1, fill=color1, semithick]
(axis cs:6.75,0.80575375)
--(axis cs:7.25,0.80575375)
--(axis cs:7.25,0.97561)
--(axis cs:6.75,0.97561)
--(axis cs:6.75,0.80575375)
--cycle;
\path [draw=color1, fill=color1, semithick]
(axis cs:7.75,0.7600975)
--(axis cs:8.25,0.7600975)
--(axis cs:8.25,0.94881925)
--(axis cs:7.75,0.94881925)
--(axis cs:7.75,0.7600975)
--cycle;
\path [draw=color1, fill=color1, semithick]
(axis cs:8.75,0.7876005)
--(axis cs:9.25,0.7876005)
--(axis cs:9.25,0.959979)
--(axis cs:8.75,0.959979)
--(axis cs:8.75,0.7876005)
--cycle;
\path [draw=color1, fill=color1, semithick]
(axis cs:9.75,0.765463)
--(axis cs:10.25,0.765463)
--(axis cs:10.25,0.9557485)
--(axis cs:9.75,0.9557485)
--(axis cs:9.75,0.765463)
--cycle;
\path [draw=color1, fill=color1, semithick]
(axis cs:10.75,0.67491025)
--(axis cs:11.25,0.67491025)
--(axis cs:11.25,0.85851875)
--(axis cs:10.75,0.85851875)
--(axis cs:10.75,0.67491025)
--cycle;
\path [draw=color1, fill=color1, semithick]
(axis cs:11.75,0.772727)
--(axis cs:12.25,0.772727)
--(axis cs:12.25,0.944444)
--(axis cs:11.75,0.944444)
--(axis cs:11.75,0.772727)
--cycle;
\path [draw=color1, fill=color1, semithick]
(axis cs:12.75,0.7893845)
--(axis cs:13.25,0.7893845)
--(axis cs:13.25,0.96418875)
--(axis cs:12.75,0.96418875)
--(axis cs:12.75,0.7893845)
--cycle;
\path [draw=color1, fill=color1, semithick]
(axis cs:13.75,0.792998)
--(axis cs:14.25,0.792998)
--(axis cs:14.25,0.96208475)
--(axis cs:13.75,0.96208475)
--(axis cs:13.75,0.792998)
--cycle;
\path [draw=color1, fill=color1, semithick]
(axis cs:14.75,0.7469605)
--(axis cs:15.25,0.7469605)
--(axis cs:15.25,0.945537)
--(axis cs:14.75,0.945537)
--(axis cs:14.75,0.7469605)
--cycle;
\path [draw=color1, fill=color1, semithick]
(axis cs:15.75,0.626742)
--(axis cs:16.25,0.626742)
--(axis cs:16.25,0.8287925)
--(axis cs:15.75,0.8287925)
--(axis cs:15.75,0.626742)
--cycle;
\path [draw=color1, fill=color1, semithick]
(axis cs:16.75,0.80927075)
--(axis cs:17.25,0.80927075)
--(axis cs:17.25,0.963636)
--(axis cs:16.75,0.963636)
--(axis cs:16.75,0.80927075)
--cycle;
\path [draw=color1, fill=color1, semithick]
(axis cs:17.75,0.716418)
--(axis cs:18.25,0.716418)
--(axis cs:18.25,0.90411)
--(axis cs:17.75,0.90411)
--(axis cs:17.75,0.716418)
--cycle;
\path [draw=color1, fill=color1, semithick]
(axis cs:18.75,0.705652)
--(axis cs:19.25,0.705652)
--(axis cs:19.25,0.885827)
--(axis cs:18.75,0.885827)
--(axis cs:18.75,0.705652)
--cycle;
\path [draw=color1, fill=color1, semithick]
(axis cs:19.75,0.72)
--(axis cs:20.25,0.72)
--(axis cs:20.25,0.90795)
--(axis cs:19.75,0.90795)
--(axis cs:19.75,0.72)
--cycle;
\path [draw=color1, fill=color1, semithick]
(axis cs:20.75,0.640216)
--(axis cs:21.25,0.640216)
--(axis cs:21.25,0.85746875)
--(axis cs:20.75,0.85746875)
--(axis cs:20.75,0.640216)
--cycle;
\path [draw=color1, fill=color1, semithick]
(axis cs:21.75,0.45067625)
--(axis cs:22.25,0.45067625)
--(axis cs:22.25,0.6280145)
--(axis cs:21.75,0.6280145)
--(axis cs:21.75,0.45067625)
--cycle;
\path [draw=color1, fill=color1, semithick]
(axis cs:22.75,0.6)
--(axis cs:23.25,0.6)
--(axis cs:23.25,0.8)
--(axis cs:22.75,0.8)
--(axis cs:22.75,0.6)
--cycle;
\path [draw=color1, fill=color1, semithick]
(axis cs:23.75,0.666667)
--(axis cs:24.25,0.666667)
--(axis cs:24.25,0.8335985)
--(axis cs:23.75,0.8335985)
--(axis cs:23.75,0.666667)
--cycle;
\path [draw=color1, fill=color1, semithick]
(axis cs:24.75,0.65902425)
--(axis cs:25.25,0.65902425)
--(axis cs:25.25,0.8438515)
--(axis cs:24.75,0.8438515)
--(axis cs:24.75,0.65902425)
--cycle;
\path [draw=color1, fill=color1, semithick]
(axis cs:25.75,0.6525175)
--(axis cs:26.25,0.6525175)
--(axis cs:26.25,0.83734075)
--(axis cs:25.75,0.83734075)
--(axis cs:25.75,0.6525175)
--cycle;
\path [draw=color1, fill=color1, semithick]
(axis cs:26.75,0.46197525)
--(axis cs:27.25,0.46197525)
--(axis cs:27.25,0.63891625)
--(axis cs:26.75,0.63891625)
--(axis cs:26.75,0.46197525)
--cycle;
\path [draw=color1, fill=color1, semithick]
(axis cs:27.75,0.615385)
--(axis cs:28.25,0.615385)
--(axis cs:28.25,0.83103175)
--(axis cs:27.75,0.83103175)
--(axis cs:27.75,0.615385)
--cycle;
\path [draw=color1, fill=color1, semithick]
(axis cs:28.75,0.64)
--(axis cs:29.25,0.64)
--(axis cs:29.25,0.847458)
--(axis cs:28.75,0.847458)
--(axis cs:28.75,0.64)
--cycle;
\path [draw=color1, fill=color1, semithick]
(axis cs:29.75,0.5946985)
--(axis cs:30.25,0.5946985)
--(axis cs:30.25,0.81447225)
--(axis cs:29.75,0.81447225)
--(axis cs:29.75,0.5946985)
--cycle;
\path [draw=color1, fill=color1, semithick]
(axis cs:30.75,0.642857)
--(axis cs:31.25,0.642857)
--(axis cs:31.25,0.85026725)
--(axis cs:30.75,0.85026725)
--(axis cs:30.75,0.642857)
--cycle;
\path [draw=color1, fill=color1, semithick]
(axis cs:31.75,0.733629191321499)
--(axis cs:32.25,0.733629191321499)
--(axis cs:32.25,0.928250641573995)
--(axis cs:31.75,0.928250641573995)
--(axis cs:31.75,0.733629191321499)
--cycle;
\path [draw=color1, fill=color1, semithick]
(axis cs:32.75,0.733219178082192)
--(axis cs:33.25,0.733219178082192)
--(axis cs:33.25,0.928250641573995)
--(axis cs:32.75,0.928250641573995)
--(axis cs:32.75,0.733219178082192)
--cycle;
\addplot [line width=1pt, white]
table {%
0.75 0.735294
1.25 0.735294
};
\addplot [line width=1pt, white]
table {%
1.75 0.692754
2.25 0.692754
};
\addplot [line width=1pt, white]
table {%
2.75 0.7351545
3.25 0.7351545
};
\addplot [line width=1pt, white]
table {%
3.75 0.612903
4.25 0.612903
};
\addplot [line width=1pt, white]
table {%
4.75 0.6831725
5.25 0.6831725
};
\addplot [line width=1pt, white]
table {%
5.75 0.68289
6.25 0.68289
};
\addplot [line width=1pt, white]
table {%
6.75 0.902727
7.25 0.902727
};
\addplot [line width=1pt, white]
table {%
7.75 0.864603
8.25 0.864603
};
\addplot [line width=1pt, white]
table {%
8.75 0.880198
9.25 0.880198
};
\addplot [line width=1pt, white]
table {%
9.75 0.8756355
10.25 0.8756355
};
\addplot [line width=1pt, white]
table {%
10.75 0.7648895
11.25 0.7648895
};
\addplot [line width=1pt, white]
table {%
11.75 0.866667
12.25 0.866667
};
\addplot [line width=1pt, white]
table {%
12.75 0.875969
13.25 0.875969
};
\addplot [line width=1pt, white]
table {%
13.75 0.878049
14.25 0.878049
};
\addplot [line width=1pt, white]
table {%
14.75 0.844757
15.25 0.844757
};
\addplot [line width=1pt, white]
table {%
15.75 0.713043
16.25 0.713043
};
\addplot [line width=1pt, white]
table {%
16.75 0.893617
17.25 0.893617
};
\addplot [line width=1pt, white]
table {%
17.75 0.809524
18.25 0.809524
};
\addplot [line width=1pt, white]
table {%
18.75 0.792915
19.25 0.792915
};
\addplot [line width=1pt, white]
table {%
19.75 0.811998
20.25 0.811998
};
\addplot [line width=1pt, white]
table {%
20.75 0.7431415
21.25 0.7431415
};
\addplot [line width=1pt, white]
table {%
21.75 0.519472
22.25 0.519472
};
\addplot [line width=1pt, white]
table {%
22.75 0.685714
23.25 0.685714
};
\addplot [line width=1pt, white]
table {%
23.75 0.74812
24.25 0.74812
};
\addplot [line width=1pt, white]
table {%
24.75 0.75
25.25 0.75
};
\addplot [line width=1pt, white]
table {%
25.75 0.736021
26.25 0.736021
};
\addplot [line width=1pt, white]
table {%
26.75 0.530026
27.25 0.530026
};
\addplot [line width=1pt, white]
table {%
27.75 0.714286
28.25 0.714286
};
\addplot [line width=1pt, white]
table {%
28.75 0.7403705
29.25 0.7403705
};
\addplot [line width=1pt, white]
table {%
29.75 0.6998405
30.25 0.6998405
};
\addplot [line width=1pt, white]
table {%
30.75 0.739866
31.25 0.739866
};
\addplot [line width=1pt, white]
table {%
31.75 0.824908759124088
32.25 0.824908759124088
};
\addplot [line width=1pt, white]
table {%
32.75 0.824908759124088
33.25 0.824908759124088
};
\end{axis}

\end{tikzpicture}

%% file: tex/6_Distinctiveness.tex
\begin{tikzpicture}

\definecolor{darkgoldenrod}{RGB}{184,134,11}
\definecolor{darkgray176}{RGB}{176,176,176}
\definecolor{firebrick}{RGB}{178,34,34}
\definecolor{steelblue31119180}{RGB}{31,119,180}
\definecolor{color0}{rgb}{0.90, 0.62, 0.00}
\definecolor{color1}{rgb}{0.34, 0.70, 0.91}
\definecolor{color2}{rgb}{0.00, 0.62, 0.45}
\definecolor{color3}{rgb}{0.94, 0.89, 0.27}
\definecolor{color4}{rgb}{0.00, 0.45, 0.69}
\definecolor{color5}{rgb}{0.83, 0.37, 0.00}

\begin{axis}[
width=\figurewidth,
height=\figureheight,
axis line style={white},
tick align=outside,
tick pos=left,
x grid style={darkgray176},
xmin=0.5, xmax=33.5,
xtick style={color=black},
xtick={1,2,3,4,5,6,7,8,9,10,11,12,13,14,15,16,17,18,19,20,21,22,23,24,25,26,27,28,29,30,31,32,33},
xticklabel style={rotate=45.0,anchor=east},
xticklabels={
  AKAZE\_AKAZE,
  BRISK\_BRIEF,
  BRISK\_BRISK,
  BRISK\_FREAK,
  BRISK\_SIFT,
  BRISK\_SURF,
  FAST\_BRIEF,
  FAST\_BRISK,
  FAST\_FREAK,
  FAST\_SIFT,
  FAST\_SURF,
  HARRIS\_BRIEF,
  HARRIS\_BRISK,
  HARRIS\_FREAK,
  HARRIS\_SIFT,
  HARRIS\_SURF,
  ORB\_ORB,
  SHITOMASI\_BRIEF,
  SHITOMASI\_BRISK,
  SHITOMASI\_FREAK,
  SHITOMASI\_SIFT,
  SHITOMASI\_SURF,
  SIFT\_BRIEF,
  SIFT\_BRISK,
  SIFT\_FREAK,
  SIFT\_SIFT,
  SIFT\_SURF,
  SURF\_BRIEF,
  SURF\_BRISK,
  SURF\_SIFT,
  SURF\_SURF,
  Superpoint,
  Superpoint\_cuda
},
y grid style={darkgray176},
ymin=0.1309318, ymax=1.0413842,
ytick style={color=black},
ymajorgrids,
ymajorticks=true,
minor y tick num = 2,
minor y grid style={dashed},
yminorgrids,
]
\addplot [black]
table {%
1 0.8
1 0.642857
};
\addplot [black]
table {%
1 0.90625
1 1
};
\addplot [black]
table {%
0.875 0.642857
1.125 0.642857
};
\addplot [black]
table {%
0.875 1
1.125 1
};
\addplot [black]
table {%
2 0.714815
2 0.588983
};
\addplot [black]
table {%
2 0.79884275
2 0.899543
};
\addplot [black]
table {%
1.875 0.588983
2.125 0.588983
};
\addplot [black]
table {%
1.875 0.899543
2.125 0.899543
};
\addplot [black]
table {%
3 0.5179265
3 0.386054
};
\addplot [black]
table {%
3 0.61574875
3 0.762178
};
\addplot [black]
table {%
2.875 0.386054
3.125 0.386054
};
\addplot [black]
table {%
2.875 0.762178
3.125 0.762178
};
\addplot [black]
table {%
4 0.461538
4 0.28125
};
\addplot [black]
table {%
4 0.603448
4 0.80597
};
\addplot [black]
table {%
3.875 0.28125
4.125 0.28125
};
\addplot [black]
table {%
3.875 0.80597
4.125 0.80597
};
\addplot [black]
table {%
5 0.60827825
5 0.488226
};
\addplot [black]
table {%
5 0.690647
5 0.81402
};
\addplot [black]
table {%
4.875 0.488226
5.125 0.488226
};
\addplot [black]
table {%
4.875 0.81402
5.125 0.81402
};
\addplot [black]
table {%
6 0.4237515
6 0.312899
};
\addplot [black]
table {%
6 0.51980325
6 0.662411
};
\addplot [black]
table {%
5.875 0.312899
6.125 0.312899
};
\addplot [black]
table {%
5.875 0.662411
6.125 0.662411
};
\addplot [black]
table {%
7 0.79495225
7 0.680556
};
\addplot [black]
table {%
7 0.87161875
7 0.955
};
\addplot [black]
table {%
6.875 0.680556
7.125 0.680556
};
\addplot [black]
table {%
6.875 0.955
7.125 0.955
};
\addplot [black]
table {%
8 0.55779775
8 0.419118
};
\addplot [black]
table {%
8 0.6664675
8 0.810872
};
\addplot [black]
table {%
7.875 0.419118
8.125 0.419118
};
\addplot [black]
table {%
7.875 0.810872
8.125 0.810872
};
\addplot [black]
table {%
9 0.61852875
9 0.472868
};
\addplot [black]
table {%
9 0.7208185
9 0.870748
};
\addplot [black]
table {%
8.875 0.472868
9.125 0.472868
};
\addplot [black]
table {%
8.875 0.870748
9.125 0.870748
};
\addplot [black]
table {%
10 0.75955325
10 0.635955
};
\addplot [black]
table {%
10 0.8430525
10 0.919694
};
\addplot [black]
table {%
9.875 0.635955
10.125 0.635955
};
\addplot [black]
table {%
9.875 0.919694
10.125 0.919694
};
\addplot [black]
table {%
11 0.41479025
11 0.312439
};
\addplot [black]
table {%
11 0.510311
11 0.653455
};
\addplot [black]
table {%
10.875 0.312439
11.125 0.312439
};
\addplot [black]
table {%
10.875 0.653455
11.125 0.653455
};
\addplot [black]
table {%
12 0.807018
12 0.666667
};
\addplot [black]
table {%
12 0.901961
12 1
};
\addplot [black]
table {%
11.875 0.666667
12.125 0.666667
};
\addplot [black]
table {%
11.875 1
12.125 1
};
\addplot [black]
table {%
13 0.6260505
13 0.47486
};
\addplot [black]
table {%
13 0.75
13 0.906433
};
\addplot [black]
table {%
12.875 0.47486
13.125 0.47486
};
\addplot [black]
table {%
12.875 0.906433
13.125 0.906433
};
\addplot [black]
table {%
14 0.656122
14 0.5
};
\addplot [black]
table {%
14 0.785047
14 0.954545
};
\addplot [black]
table {%
13.875 0.5
14.125 0.5
};
\addplot [black]
table {%
13.875 0.954545
14.125 0.954545
};
\addplot [black]
table {%
15 0.825532
15 0.72807
};
\addplot [black]
table {%
15 0.893023
15 0.956098
};
\addplot [black]
table {%
14.875 0.72807
15.125 0.72807
};
\addplot [black]
table {%
14.875 0.956098
15.125 0.956098
};
\addplot [black]
table {%
16 0.4176605
16 0.261682
};
\addplot [black]
table {%
16 0.53558725
16 0.712389
};
\addplot [black]
table {%
15.875 0.261682
16.125 0.261682
};
\addplot [black]
table {%
15.875 0.712389
16.125 0.712389
};
\addplot [black]
table {%
17 0.7
17 0.526316
};
\addplot [black]
table {%
17 0.81638725
17 0.984848
};
\addplot [black]
table {%
16.875 0.526316
17.125 0.526316
};
\addplot [black]
table {%
16.875 0.984848
17.125 0.984848
};
\addplot [black]
table {%
18 0.760417
18 0.649351
};
\addplot [black]
table {%
18 0.837209
18 0.951807
};
\addplot [black]
table {%
17.875 0.649351
18.125 0.649351
};
\addplot [black]
table {%
17.875 0.951807
18.125 0.951807
};
\addplot [black]
table {%
19 0.44370125
19 0.316667
};
\addplot [black]
table {%
19 0.548135
19 0.693512
};
\addplot [black]
table {%
18.875 0.316667
19.125 0.316667
};
\addplot [black]
table {%
18.875 0.693512
19.125 0.693512
};
\addplot [black]
table {%
20 0.562786
20 0.433862
};
\addplot [black]
table {%
20 0.662411
20 0.807339
};
\addplot [black]
table {%
19.875 0.433862
20.125 0.433862
};
\addplot [black]
table {%
19.875 0.807339
20.125 0.807339
};
\addplot [black]
table {%
21 0.720786
21 0.625
};
\addplot [black]
table {%
21 0.78476
21 0.864769
};
\addplot [black]
table {%
20.875 0.625
21.125 0.625
};
\addplot [black]
table {%
20.875 0.864769
21.125 0.864769
};
\addplot [black]
table {%
22 0.27867325
22 0.209854
};
\addplot [black]
table {%
22 0.3561085
22 0.472172
};
\addplot [black]
table {%
21.875 0.209854
22.125 0.209854
};
\addplot [black]
table {%
21.875 0.472172
22.125 0.472172
};
\addplot [black]
table {%
23 0.52459
23 0.389831
};
\addplot [black]
table {%
23 0.617647
23 0.753846
};
\addplot [black]
table {%
22.875 0.389831
23.125 0.389831
};
\addplot [black]
table {%
22.875 0.753846
23.125 0.753846
};
\addplot [black]
table {%
24 0.397887
24 0.278761
};
\addplot [black]
table {%
24 0.48609375
24 0.609562
};
\addplot [black]
table {%
23.875 0.278761
24.125 0.278761
};
\addplot [black]
table {%
23.875 0.609562
24.125 0.609562
};
\addplot [black]
table {%
25 0.43899025
25 0.298851
};
\addplot [black]
table {%
25 0.53549275
25 0.670588
};
\addplot [black]
table {%
24.875 0.298851
25.125 0.298851
};
\addplot [black]
table {%
24.875 0.670588
25.125 0.670588
};
\addplot [black]
table {%
26 0.510989
26 0.401869
};
\addplot [black]
table {%
26 0.5985645
26 0.724675
};
\addplot [black]
table {%
25.875 0.401869
26.125 0.401869
};
\addplot [black]
table {%
25.875 0.724675
26.125 0.724675
};
\addplot [black]
table {%
27 0.235638
27 0.172316
};
\addplot [black]
table {%
27 0.2984895
27 0.39255
};
\addplot [black]
table {%
26.875 0.172316
27.125 0.172316
};
\addplot [black]
table {%
26.875 0.39255
27.125 0.39255
};
\addplot [black]
table {%
28 0.70188325
28 0.545455
};
\addplot [black]
table {%
28 0.80627225
28 0.944954
};
\addplot [black]
table {%
27.875 0.545455
28.125 0.545455
};
\addplot [black]
table {%
27.875 0.944954
28.125 0.944954
};
\addplot [black]
table {%
29 0.557707
29 0.37037
};
\addplot [black]
table {%
29 0.68482025
29 0.863636
};
\addplot [black]
table {%
28.875 0.37037
29.125 0.37037
};
\addplot [black]
table {%
28.875 0.863636
29.125 0.863636
};
\addplot [black]
table {%
30 0.663597
30 0.545035
};
\addplot [black]
table {%
30 0.74388375
30 0.851685
};
\addplot [black]
table {%
29.875 0.545035
30.125 0.545035
};
\addplot [black]
table {%
29.875 0.851685
30.125 0.851685
};
\addplot [black]
table {%
31 0.574793
31 0.435597
};
\addplot [black]
table {%
31 0.6744095
31 0.814596
};
\addplot [black]
table {%
30.875 0.435597
31.125 0.435597
};
\addplot [black]
table {%
30.875 0.814596
31.125 0.814596
};
\addplot [black]
table {%
32 0.87218992248062
32 0.806896551724138
};
\addplot [black]
table {%
32 0.917197452229299
32 0.982142857142857
};
\addplot [black]
table {%
31.875 0.806896551724138
32.125 0.806896551724138
};
\addplot [black]
table {%
31.875 0.982142857142857
32.125 0.982142857142857
};
\addplot [black]
table {%
33 0.87218992248062
33 0.806896551724138
};
\addplot [black]
table {%
33 0.917197452229299
33 0.982142857142857
};
\addplot [black]
table {%
32.875 0.806896551724138
33.125 0.806896551724138
};
\addplot [black]
table {%
32.875 0.982142857142857
33.125 0.982142857142857
};
\path [draw=color1, fill=color1, semithick]
(axis cs:0.75,0.8)
--(axis cs:1.25,0.8)
--(axis cs:1.25,0.90625)
--(axis cs:0.75,0.90625)
--(axis cs:0.75,0.8)
--cycle;
\path [draw=color1, fill=color1, semithick]
(axis cs:1.75,0.714815)
--(axis cs:2.25,0.714815)
--(axis cs:2.25,0.79884275)
--(axis cs:1.75,0.79884275)
--(axis cs:1.75,0.714815)
--cycle;
\path [draw=color1, fill=color1, semithick]
(axis cs:2.75,0.5179265)
--(axis cs:3.25,0.5179265)
--(axis cs:3.25,0.61574875)
--(axis cs:2.75,0.61574875)
--(axis cs:2.75,0.5179265)
--cycle;
\path [draw=color1, fill=color1, semithick]
(axis cs:3.75,0.461538)
--(axis cs:4.25,0.461538)
--(axis cs:4.25,0.603448)
--(axis cs:3.75,0.603448)
--(axis cs:3.75,0.461538)
--cycle;
\path [draw=color1, fill=color1, semithick]
(axis cs:4.75,0.60827825)
--(axis cs:5.25,0.60827825)
--(axis cs:5.25,0.690647)
--(axis cs:4.75,0.690647)
--(axis cs:4.75,0.60827825)
--cycle;
\path [draw=color1, fill=color1, semithick]
(axis cs:5.75,0.4237515)
--(axis cs:6.25,0.4237515)
--(axis cs:6.25,0.51980325)
--(axis cs:5.75,0.51980325)
--(axis cs:5.75,0.4237515)
--cycle;
\path [draw=color1, fill=color1, semithick]
(axis cs:6.75,0.79495225)
--(axis cs:7.25,0.79495225)
--(axis cs:7.25,0.87161875)
--(axis cs:6.75,0.87161875)
--(axis cs:6.75,0.79495225)
--cycle;
\path [draw=color1, fill=color1, semithick]
(axis cs:7.75,0.55779775)
--(axis cs:8.25,0.55779775)
--(axis cs:8.25,0.6664675)
--(axis cs:7.75,0.6664675)
--(axis cs:7.75,0.55779775)
--cycle;
\path [draw=color1, fill=color1, semithick]
(axis cs:8.75,0.61852875)
--(axis cs:9.25,0.61852875)
--(axis cs:9.25,0.7208185)
--(axis cs:8.75,0.7208185)
--(axis cs:8.75,0.61852875)
--cycle;
\path [draw=color1, fill=color1, semithick]
(axis cs:9.75,0.75955325)
--(axis cs:10.25,0.75955325)
--(axis cs:10.25,0.8430525)
--(axis cs:9.75,0.8430525)
--(axis cs:9.75,0.75955325)
--cycle;
\path [draw=color1, fill=color1, semithick]
(axis cs:10.75,0.41479025)
--(axis cs:11.25,0.41479025)
--(axis cs:11.25,0.510311)
--(axis cs:10.75,0.510311)
--(axis cs:10.75,0.41479025)
--cycle;
\path [draw=color1, fill=color1, semithick]
(axis cs:11.75,0.807018)
--(axis cs:12.25,0.807018)
--(axis cs:12.25,0.901961)
--(axis cs:11.75,0.901961)
--(axis cs:11.75,0.807018)
--cycle;
\path [draw=color1, fill=color1, semithick]
(axis cs:12.75,0.6260505)
--(axis cs:13.25,0.6260505)
--(axis cs:13.25,0.75)
--(axis cs:12.75,0.75)
--(axis cs:12.75,0.6260505)
--cycle;
\path [draw=color1, fill=color1, semithick]
(axis cs:13.75,0.656122)
--(axis cs:14.25,0.656122)
--(axis cs:14.25,0.785047)
--(axis cs:13.75,0.785047)
--(axis cs:13.75,0.656122)
--cycle;
\path [draw=color1, fill=color1, semithick]
(axis cs:14.75,0.825532)
--(axis cs:15.25,0.825532)
--(axis cs:15.25,0.893023)
--(axis cs:14.75,0.893023)
--(axis cs:14.75,0.825532)
--cycle;
\path [draw=color1, fill=color1, semithick]
(axis cs:15.75,0.4176605)
--(axis cs:16.25,0.4176605)
--(axis cs:16.25,0.53558725)
--(axis cs:15.75,0.53558725)
--(axis cs:15.75,0.4176605)
--cycle;
\path [draw=color1, fill=color1, semithick]
(axis cs:16.75,0.7)
--(axis cs:17.25,0.7)
--(axis cs:17.25,0.81638725)
--(axis cs:16.75,0.81638725)
--(axis cs:16.75,0.7)
--cycle;
\path [draw=color1, fill=color1, semithick]
(axis cs:17.75,0.760417)
--(axis cs:18.25,0.760417)
--(axis cs:18.25,0.837209)
--(axis cs:17.75,0.837209)
--(axis cs:17.75,0.760417)
--cycle;
\path [draw=color1, fill=color1, semithick]
(axis cs:18.75,0.44370125)
--(axis cs:19.25,0.44370125)
--(axis cs:19.25,0.548135)
--(axis cs:18.75,0.548135)
--(axis cs:18.75,0.44370125)
--cycle;
\path [draw=color1, fill=color1, semithick]
(axis cs:19.75,0.562786)
--(axis cs:20.25,0.562786)
--(axis cs:20.25,0.662411)
--(axis cs:19.75,0.662411)
--(axis cs:19.75,0.562786)
--cycle;
\path [draw=color1, fill=color1, semithick]
(axis cs:20.75,0.720786)
--(axis cs:21.25,0.720786)
--(axis cs:21.25,0.78476)
--(axis cs:20.75,0.78476)
--(axis cs:20.75,0.720786)
--cycle;
\path [draw=color1, fill=color1, semithick]
(axis cs:21.75,0.27867325)
--(axis cs:22.25,0.27867325)
--(axis cs:22.25,0.3561085)
--(axis cs:21.75,0.3561085)
--(axis cs:21.75,0.27867325)
--cycle;
\path [draw=color1, fill=color1, semithick]
(axis cs:22.75,0.52459)
--(axis cs:23.25,0.52459)
--(axis cs:23.25,0.617647)
--(axis cs:22.75,0.617647)
--(axis cs:22.75,0.52459)
--cycle;
\path [draw=color1, fill=color1, semithick]
(axis cs:23.75,0.397887)
--(axis cs:24.25,0.397887)
--(axis cs:24.25,0.48609375)
--(axis cs:23.75,0.48609375)
--(axis cs:23.75,0.397887)
--cycle;
\path [draw=color1, fill=color1, semithick]
(axis cs:24.75,0.43899025)
--(axis cs:25.25,0.43899025)
--(axis cs:25.25,0.53549275)
--(axis cs:24.75,0.53549275)
--(axis cs:24.75,0.43899025)
--cycle;
\path [draw=color1, fill=color1, semithick]
(axis cs:25.75,0.510989)
--(axis cs:26.25,0.510989)
--(axis cs:26.25,0.5985645)
--(axis cs:25.75,0.5985645)
--(axis cs:25.75,0.510989)
--cycle;
\path [draw=color1, fill=color1, semithick]
(axis cs:26.75,0.235638)
--(axis cs:27.25,0.235638)
--(axis cs:27.25,0.2984895)
--(axis cs:26.75,0.2984895)
--(axis cs:26.75,0.235638)
--cycle;
\path [draw=color1, fill=color1, semithick]
(axis cs:27.75,0.70188325)
--(axis cs:28.25,0.70188325)
--(axis cs:28.25,0.80627225)
--(axis cs:27.75,0.80627225)
--(axis cs:27.75,0.70188325)
--cycle;
\path [draw=color1, fill=color1, semithick]
(axis cs:28.75,0.557707)
--(axis cs:29.25,0.557707)
--(axis cs:29.25,0.68482025)
--(axis cs:28.75,0.68482025)
--(axis cs:28.75,0.557707)
--cycle;
\path [draw=color1, fill=color1, semithick]
(axis cs:29.75,0.663597)
--(axis cs:30.25,0.663597)
--(axis cs:30.25,0.74388375)
--(axis cs:29.75,0.74388375)
--(axis cs:29.75,0.663597)
--cycle;
\path [draw=color1, fill=color1, semithick]
(axis cs:30.75,0.574793)
--(axis cs:31.25,0.574793)
--(axis cs:31.25,0.6744095)
--(axis cs:30.75,0.6744095)
--(axis cs:30.75,0.574793)
--cycle;
\path [draw=color1, fill=color1, semithick]
(axis cs:31.75,0.87218992248062)
--(axis cs:32.25,0.87218992248062)
--(axis cs:32.25,0.917197452229299)
--(axis cs:31.75,0.917197452229299)
--(axis cs:31.75,0.87218992248062)
--cycle;
\path [draw=color1, fill=color1, semithick]
(axis cs:32.75,0.87218992248062)
--(axis cs:33.25,0.87218992248062)
--(axis cs:33.25,0.917197452229299)
--(axis cs:32.75,0.917197452229299)
--(axis cs:32.75,0.87218992248062)
--cycle;
\addplot [line width=1pt, white]
table {%
0.75 0.853659
1.25 0.853659
};
\addplot [line width=1pt, white]
table {%
1.75 0.75901
2.25 0.75901
};
\addplot [line width=1pt, white]
table {%
2.75 0.559641
3.25 0.559641
};
\addplot [line width=1pt, white]
table {%
3.75 0.527273
4.25 0.527273
};
\addplot [line width=1pt, white]
table {%
4.75 0.64574
5.25 0.64574
};
\addplot [line width=1pt, white]
table {%
5.75 0.4627695
6.25 0.4627695
};
\addplot [line width=1pt, white]
table {%
6.75 0.8325685
7.25 0.8325685
};
\addplot [line width=1pt, white]
table {%
7.75 0.601375
8.25 0.601375
};
\addplot [line width=1pt, white]
table {%
8.75 0.661135
9.25 0.661135
};
\addplot [line width=1pt, white]
table {%
9.75 0.7989855
10.25 0.7989855
};
\addplot [line width=1pt, white]
table {%
10.75 0.450357
11.25 0.450357
};
\addplot [line width=1pt, white]
table {%
11.75 0.857143
12.25 0.857143
};
\addplot [line width=1pt, white]
table {%
12.75 0.674559
13.25 0.674559
};
\addplot [line width=1pt, white]
table {%
13.75 0.709402
14.25 0.709402
};
\addplot [line width=1pt, white]
table {%
14.75 0.860456
15.25 0.860456
};
\addplot [line width=1pt, white]
table {%
15.75 0.461917
16.25 0.461917
};
\addplot [line width=1pt, white]
table {%
16.75 0.75641
17.25 0.75641
};
\addplot [line width=1pt, white]
table {%
17.75 0.8
18.25 0.8
};
\addplot [line width=1pt, white]
table {%
18.75 0.485244
19.25 0.485244
};
\addplot [line width=1pt, white]
table {%
19.75 0.607955
20.25 0.607955
};
\addplot [line width=1pt, white]
table {%
20.75 0.7520235
21.25 0.7520235
};
\addplot [line width=1pt, white]
table {%
21.75 0.309091
22.25 0.309091
};
\addplot [line width=1pt, white]
table {%
22.75 0.571429
23.25 0.571429
};
\addplot [line width=1pt, white]
table {%
23.75 0.4307495
24.25 0.4307495
};
\addplot [line width=1pt, white]
table {%
24.75 0.4809435
25.25 0.4809435
};
\addplot [line width=1pt, white]
table {%
25.75 0.5473525
26.25 0.5473525
};
\addplot [line width=1pt, white]
table {%
26.75 0.261187
27.25 0.261187
};
\addplot [line width=1pt, white]
table {%
27.75 0.75
28.25 0.75
};
\addplot [line width=1pt, white]
table {%
28.75 0.615385
29.25 0.615385
};
\addplot [line width=1pt, white]
table {%
29.75 0.69697
30.25 0.69697
};
\addplot [line width=1pt, white]
table {%
30.75 0.611682
31.25 0.611682
};
\addplot [line width=1pt, white]
table {%
31.75 0.895044676352227
32.25 0.895044676352227
};
\addplot [line width=1pt, white]
table {%
32.75 0.895044676352227
33.25 0.895044676352227
};
\end{axis}

\end{tikzpicture}

%% file: tbs/kissicp-neighbor-evaluate.tex
\begin{table}[H]
\centering
\caption{Performance evaluation of LO (KISS-ICP) with raw point cloud and our downsampled point cloud, `\textit{Sig}' and `\textit{Rng}' represent the size of neighboring point areas for the signal and range images, respectively, denoted as \textit{Sig\_Rng}. }
\label{tab:kissicp-size-evaluate}
\resizebox{0.48\textwidth}{!}{%
\begin{tabular}{cccccc}
\toprule 
\multirow{2}{*}{\textbf{\shortstack[c]{Neighbor Size\\ (\textit{Sig}\_\textit{Rng})}}}
& \multicolumn{1}{c}{\textbf{Open road}}               & \multicolumn{1}{c}{\textbf{Forest}}             & \multicolumn{1}{c}{\textbf{Lab space (hard)}}        & \multicolumn{1}{c}{\textbf{ Lab space (easy)}}          & \multicolumn{1}{c}{\textbf{Hall (large)}}        \\
& \multicolumn{5}{c}{\textit{(Translation error (mean/rmse)(m), rotation error(deg))} } \\
\midrule
\textbf{\textit{4\_4}}       & N / A                & (0.079/0.090, 6.58) & (0.052/0.062, \textbf{1.44})              & (0.027/0.031, 0.99) & (1.111/1.274, 3.37)  \\
\textbf{\textit{4\_5}}       & N / A                & (0.086/0.096, 7.22) & (0.043/0.051, 1.51)                       & (0.031/0.035, 1.05) & (0.724/0.819, 2.95)  \\
\textbf{\textit{4\_7}}       & (\textbf{0.817/0.952}, 2.33) & (0.082/0.102, 7.78) & (0.039/0.046, 1.46)                & (0.028/0.033, 0.98) & (\textbf{0.583/0.660}, 2.88)  \\
\textbf{\textit{5\_4}}       & (1.724/2.038, 2.10)  & (0.085/0.100, 6.81) & (0.059/0.070, 1.71)                       & (0.025/0.028, 0.98) & (1.065/1.242, 2.73)  \\
\textbf{\textit{5\_5}}       & (2.176/2.410, \textbf{1.76}) & (0.108/0.203, 6.96) & (0.037/0.043, 1.35)               & (0.028/0.032, 0.97) & (0.707/0.801, \textbf{2.66})  \\
\textbf{\textit{5\_7}}       & (1.298/1.443, 2.71)  & (0.076/0.084, \textbf{6.11}) & (0.064/0.075, 1.54)              & (0.025/0.028, 0.94) & (0.676/0.746, 3.67)  \\
\textbf{\textit{7\_4}}       & (1.696/1.888, 2.31)  & (0.082/0.094, 6.98) & (0.074/0.085, 1.64)                       & (0.027/0.032, 0.99) & (0.806/0.917, 3.68)  \\
\textbf{\textit{7\_5}}       & (1.784/2.006, 2.30)  & (0.080/0.102, 7.72) & (\textbf{0.033/0.047}, 1.59)               & (0.025/0.028, 0.97) & (0.698/0.803, 3.11)  \\
Raw PC                       & N / A                & (\textbf{0.057/0.073}, 8.91) & N / A                             & (\textbf{0.020/0.022}, \textbf{0.62} & N / A              ) \\
\bottomrule
\end{tabular}%
}
\end{table}

%% file: tbs/kiss-points.tex
\begin{table}[h]
\centering
\caption{The number of points left after downsampling with varied neighbor size, `\textit{Sig}' and `\textit{Rng}' represent the size of neighboring point areas for the signal and range images, respectively, denoted as \textit{Sig\_Rng}. }
\label{tab:kissicp-points}
\resizebox{0.48\textwidth}{!}{%
\begin{tabular}{cccccc}
\toprule 
\multirow{2}{*}{\textbf{\shortstack[c]{Neighbor Size\\ (\textit{Sig}\_\textit{Rng})}}}
& \multicolumn{1}{c}{\textbf{Open road}}               & \multicolumn{1}{c}{\textbf{Forest}}             & \multicolumn{1}{c}{\textbf{Lab space (hard)}}        & \multicolumn{1}{c}{\textbf{ Lab space (easy)}}          & \multicolumn{1}{c}{\textbf{Hall (large)}}        \\
& \multicolumn{5}{c}{\textit{Number of points(pts)} } \\
\midrule
\textbf{\textit{4\_4}}  &   2650   & 7787     & 6435                  & 6360         & 4742   \\
\textbf{\textit{4\_5}}  &   3206   & 7812     & 6416                  & 6302         & 4792   \\ 
\textbf{\textit{4\_7}}  &    4784   & 11447    & 9518                  & 9392         & 7094   \\
\textbf{\textit{5\_4}}  &    3182   & 7843     & 6409                  & 6333         & 4776   \\
\textbf{\textit{5\_5}}  &   3183   & 7568     & 6446                  & 6292         & 4783   \\
\textbf{\textit{5\_7}}  &   4763   & 11519    & 9513                  & 9386         & 7066   \\
\textbf{\textit{7\_4}}  &    4760   & 11631    & 9445                  & 9356         & 7070   \\
\textbf{\textit{7\_5}}  &   4756   & 11627    & 9469                  & 9378         & 7078   \\
Raw PC                  &    131072 & 131072   & 131072                & 131072       & 131072 \\
\bottomrule
\end{tabular}%
}
\end{table}

%% file: tbs/kiss-icp-perf.tex
\begin{table}[H]
\centering
\caption{Evaluation of LO based on conventional and DL keypoint detectors with KISS-ICP}
\label{tab:kiss-icp-perf}
\resizebox{0.48\textwidth}{!}{%
\begin{tabular}{l|ccc|ccc}
\hline
 \multirow{3}{*}{\backslashbox{Evaluation\\\strut Indicators}{\strut Approaches}}  & \multicolumn{6}{|c}{\textbf{KISS-ICP}}  \\ \cline{2-7} 
                    & \multicolumn{3}{|c|}{\textbf{Outdoor}} &  \multicolumn{3}{c}{\textbf{Indoor}}  \\\cline{2-7} 
                    & \multicolumn{1}{|c|}{\textbf{AKAZE}} & \multicolumn{1}{c|}{\textbf{Superpoint}} & \multicolumn{1}{c|}{\textbf{Raw PC}} & \multicolumn{1}{c|}{\textbf{AKAZE}} & \multicolumn{1}{c|}{\textbf{Superpoint}} & \multicolumn{1}{c}{\textbf{Raw PC}}  \\ 
\hline
Translation Error \textit{(m)}        & 0.096/0.107         & \textbf{0.082/0.102}    & N / A    & 0.092/0.099   & \textbf{0.037/0.043}  &  N / A           \\
Rotation error \textit{(deg)}         & \textbf{4.27}       & 7.78                    & N / A      & \textbf{1.22}        & 1.71            & N / A       \\
CPU \textit{(\%)}                     & \textbf{263.16}     & 457.59                  & 544.61   & \textbf{82.57}         & 425.93       & 572.50           \\
Mem \textit{(MB)}                     & 247.62     & 308.67                  & \textbf{165.73}   & 198.27       & 232.85       & \textbf{84.37}        \\
Avg Pts                             & \textbf{4849}           & 11447          & 131072         & \textbf{1176}          & 6446          & 131072            \\
Odom Rate \textit{(Hz)}     & \textbf{10.0}           & 4.0                     &2.95        & \textbf{10.0}          & 7.6          & 2.82           \\ 
\bottomrule
\end{tabular}%
}
\end{table}

%% file: tbs/kiss-ndt-evaluate.tex
\begin{table}[H]
\centering
\caption{Evaluation of LO based on conventional and DL keypoint detectors with  NDT}
\label{tab:kiss-ndt-evaluate}
\resizebox{0.48\textwidth}{!}{%
\begin{tabular}{l|ccc|ccc}
\hline
 \multirow{3}{*}{\backslashbox{Evaluation\\\strut Indicators}{\strut Approaches}}  &   \multicolumn{6}{c}{\textbf{NDT}} \\ \cline{2-7}
                     & \multicolumn{3}{c|}{\textbf{Outdoor}} &  \multicolumn{3}{c}{\textbf{Indoor}} \\\cline{2-7}
                      & \multicolumn{1}{c|}{\textbf{AKAZE}} & \multicolumn{1}{c}{\textbf{Superpoint}} & \multicolumn{1}{c|}{\textbf{Raw PC}}& \multicolumn{1}{c|}{\textbf{AKAZE}} & \multicolumn{1}{c}{\textbf{Superpoint}} & \multicolumn{1}{c}{\textbf{Raw PC}} \\
\hline
Translation Error \textit{(m)}      & 0.115/0.126           & \textbf{0.090/0.098} & N / A & 0.102/0.114  & \textbf{0.054/0.071} & N / A \\
Rotation error \textit{(deg)}       &  \textbf{4.84}        & 5.66                  & N / A & \textbf{1.15}        & 1.31       &N / A \\
CPU \textit{(\%)}                   &  \textbf{100.39}      & 325.69                &  571.12  & \textbf{82.20}      & 338.54   &  581.30  \\
Mem \textit{(MB)}                   & \textbf{285.06}       & 548.16                & 705.43  & \textbf{253.95}     & 290.34    & 645.21  \\
Avg Pts                             &\textbf{4849}          & 11447                 & 131072 &\textbf{1176} &   6446     &   131072 \\
Odom Rate \textit{(Hz)}             & \textbf{10.0}         & 4.6                   & 3.34  & \textbf{10}           & 8.21  &  1.20     \\ 
\bottomrule
\end{tabular}%
}
\end{table}

%% file: sections/05-conclusion.tex
\section{Conclusion and future work}
\label{sec:conclusion-future}
To mitigate computational overhead while ensuring the retention of a sufficient number of dependable key points for point cloud registration in LO, this study introduces a novel approach that incorporates LiDAR-generated images. A comprehensive analysis of keypoint detection and descriptors, originally designed for conventional images, is conducted on the LiDAR-generated image. This not only informs subsequent sections of this paper but also sets the stage for future research endeavors aimed at enhancing the robustness and resilience of LO and SLAM technology. Building upon the insights gleaned from this analysis, we propose a methodology for down-sampling the raw point cloud while preserving the integrity of salient points. Our experiments demonstrate that our proposed approach exhibits comparable performance to utilizing the complete raw point cloud and, notably, surpasses it in scenarios where the full raw point cloud proves ineffective, such as in cases of drift. Additionally, our approach exhibits commendable robustness in the face of rotational transformations. The computation overhead of our approach is lower than the LO utilizing raw point cloud but with a higher odometry publishing rate.

In future work, there is potential to seamlessly integrate the current LiDAR-generated image keypoint extraction process into the broader SLAM pipeline. For instance, one avenue of exploration could involve amalgamating features extracted from LiDAR-generated images with those derived from point cloud data, facilitating the development of a lightweight SLAM system complemented by additional sensors, such as an IMU.

